\documentclass[preprint,12pt]{elsarticle}

\usepackage{amsmath,amsfonts}
\usepackage{booktabs}
\usepackage{makecell}
\usepackage{algorithmic}
\usepackage{algorithm}
\usepackage{array}
\usepackage[caption=false,font=normalsize,labelfont=sf,textfont=sf]{subfig}
\usepackage{textcomp}
\usepackage{stfloats}
\usepackage{url}
\usepackage{verbatim}
\usepackage{graphicx}
\usepackage[colorlinks,linkcolor=magenta]{hyperref}
\usepackage{multirow} 
\usepackage{subfig}

\journal{Nuclear Physics B}

\begin{document}

\begin{frontmatter}



\title{Fourier-RWKV: A Multi-State Perception Network for Efficient Image Dehazing}

\author[1]{Lirong Zheng}
\ead{zhenglirong2021@email.szu.edu.cn}
\author[1]{Yanshan Li\protect\corref{cor1}}
\ead{lys@szu.edu.cn}
\cortext[cor1]{Corresponding author}

\author[1]{Rui Yu}
\ead{yurui2020@email.szu.edu.cn}
\author[2]{Kaihao Zhang}
\ead{super.khzhang@gmail.com}

\affiliation[1]{organization={Institute of Intelligent Information Processing, Guangdong Provincial Key Laboratory of Intelligent Information Processing, Shenzhen Key Laboratory of Modern Communications and Information Processing, Shenzhen University},
        city={Shenzhen},
        postcode={518060},
        country={China}}
        
\affiliation[2]{organization={College of Engineering and Computer Science, Australian National University},
        city={Canberra},
        postcode={2601},
        state={ACT},
        country={Australia}}

\begin{abstract}
Image dehazing is crucial for reliable visual perception, yet it remains highly challenging under real-world non-uniform haze conditions. Although Transformers excel at capturing global context, their quadratic computational complexity hinders real-time deployment. To address this, we propose Fourier Receptance Weighted Key Value (Fourier-RWKV), a novel dehazing framework based on a Multi-State Perception paradigm. The model achieves comprehensive haze degradation modeling with linear complexity by synergistically integrating three distinct perceptual states:
(1) Spatial-form Perception, realized through the Deformable Quad-directional Token Shift (DQ-Shift) operation, which dynamically adjusts receptive fields to accommodate local haze variations; (2) Frequency-domain Perception, implemented within the Fourier Mix block, which extends the core WKV attention mechanism of RWKV from the spatial domain to the Fourier domain, preserving the long-range dependencies essential for global haze estimation while mitigating spatial attenuation; (3) Semantic-relation Perception, facilitated by the Semantic Bridge Module (SBM), which utilizes Dynamic Semantic Kernel Fusion (DSK-Fusion) to precisely align encoder-decoder features and suppress artifacts.
Extensive experiments on multiple benchmarks demonstrate that Fourier-RWKV delivers state-of-the-art performance across diverse haze scenarios while significantly reducing computational overhead, establishing a favorable trade-off between restoration quality and practical efficiency. 
Code is available at: \hyperlink{code}{https://github.com/Dilizlr/Fourier-RWKV}.
\end{abstract}




\end{frontmatter}


\section{Introduction}
\label{sec: intro}
Robust visual perception systems play a critical role in applications such as autonomous driving~\cite{LU2025111652, GUO2025111174} and person re-identification~\cite{ZHANG2025111702, ZHU2025111669}, with their performance largely dependent on the quality of the input images. Haze under adverse weather conditions severely impairs images, resulting in color distortion, blurred textures, and reduced contrast. Furthermore, spatially non-uniform haze increases the complexity of restoration due to localized variations in density. Therefore, effective dehazing algorithms must balance local adaptability with global consistency to restore image fidelity.

\begin{figure}[tbp]
  \centering
  \includegraphics[width=0.95\columnwidth]{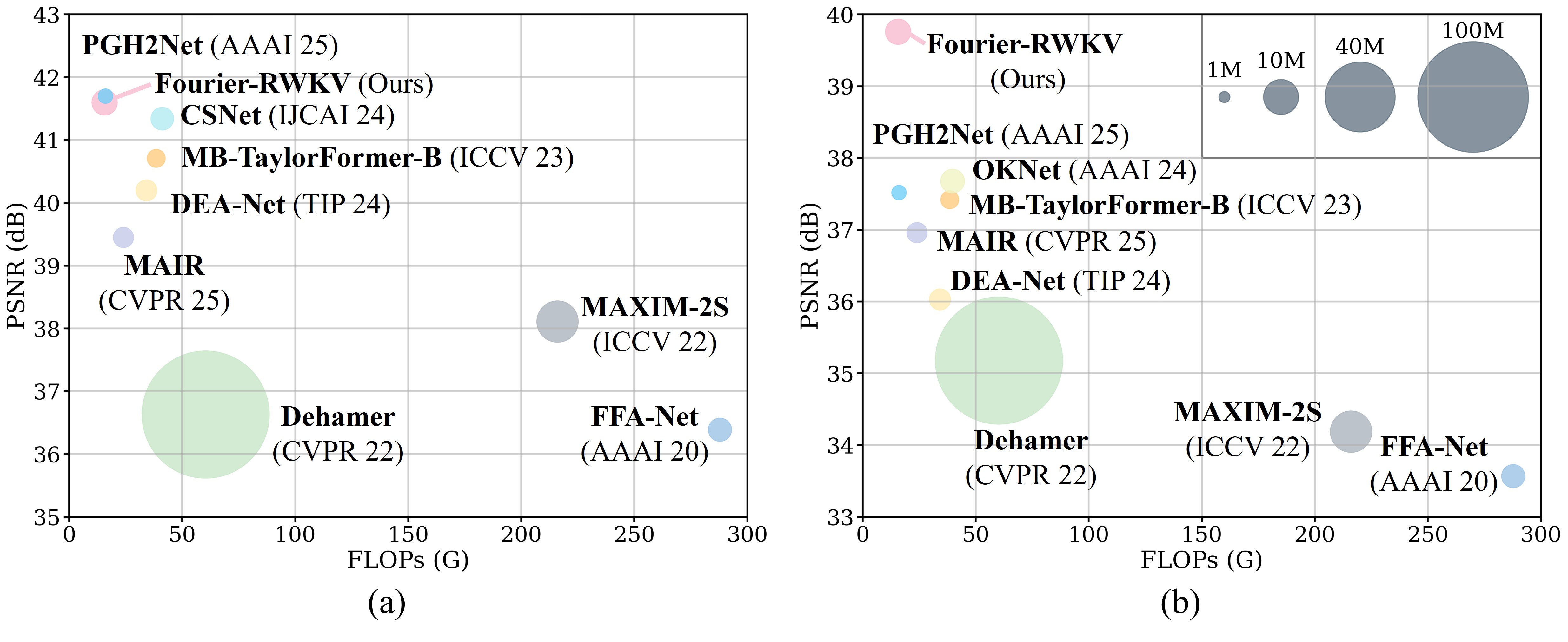}
  \caption{Improvement of our Fourier-RWKV over the SOTA approaches. The bubble size represents the number of model parameters. All models are tested on (a) SOTS-Indoor and (b) SOTS-Outdoor datasets.}
  \label{fig: sota}
\end{figure}

Early image dehazing methods~\cite{5567108, 7128396, 7984895, 8101508} rely on handcrafted priors to estimate transmission maps, but often fail to generalize in real-world settings due to their simplified assumptions. With the rise of deep learning, convolutional neural networks (CNNs)~\cite{2016DehazeNet, 9919385, WANG2024109956} have significantly enhanced dehazing performance by automatically learning haze-related features from large-scale data. However, the limited receptive fields of CNNs hinder the modeling of long-range dependencies, which are crucial for accurate global haze estimation, especially in complex scenes. By leveraging the self-attention mechanism, Transformers~\cite{10.1145/3664647.3681314, 10076399, Yang2023, wang2024gridformer} exhibit superior capability in capturing global context, driving remarkable progress in image dehazing. Nevertheless, their quadratic computational complexity incurs high computational costs on high-resolution images, restricting their real-time applicability.

To overcome the efficiency bottleneck, researchers have begun exploring attention architectures with linear complexity, such as State Space Models (SSMs)~\cite{DBLPconficlrGuGR22, DBLPconfnipsGuJGSDRR21, DBLPjournalscorrabs-2312-00752} and the RWKV model~\cite{DBLPconfemnlpPengAAAABCCCDDG23, DBLPjournalscorrabs-2404-05892}. Among them, Vision-RWKV has demonstrated promise for efficient long-range modeling in vision tasks. However, its direct application to image dehazing encounters several key architectural limitations: (1) The spatial operations in Vision-RWKV~\cite{DBLPconficlrDuanWCZLLQ0DW25}, such as the fixed Q-Shift, lack the adaptability required to capture irregular haze distributions; (2) The sequential state modeling in the spatial domain, driven by the WKV attention mechanism, suffers from long-range information decay, undermining both accuracy and consistency in global haze estimation; (3) When integrated into the U-Net encoder–decoder framework, the inherent semantic gap between encoding and decoding stages causes feature misalignment and artifacts—an issue that remains unresolved in the original Vision-RWKV.

In this paper, we propose Fourier-RWKV, a novel linear-complexity framework based on the Multi-State Perception paradigm. It is designed to address the architectural limitations of Vision-RWKV and advance the performance of efficient image restoration models. By unifying three complementary perceptual states—spatially deformable perception, frequency-domain global modeling, and semantic-guided feature fusion—Fourier-RWKV provides a comprehensive solution to diverse haze degradation patterns.

\begin{figure}[t]
    \centering
    \includegraphics[width=0.8\columnwidth]{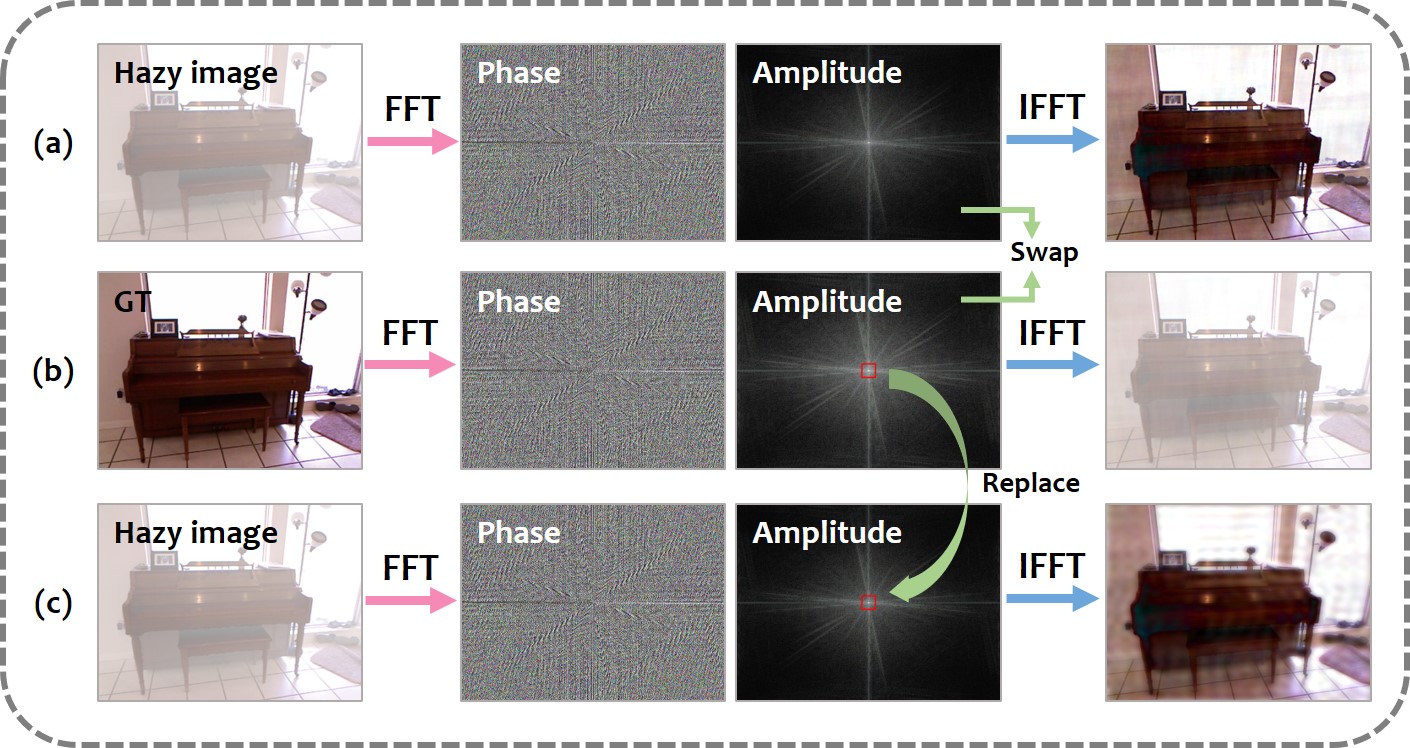}
    \caption{Spectral Properties of Haze. The exchange of amplitude spectrum (a, b) reveals that haze information is primarily encoded in the amplitude spectrum, while the phase spectrum retains most structural integrity. Replacing the DC component (b, c) improves global image contrast, indicating that haze degradation is concentrated in low-frequency regions and demonstrating the global effect of frequency-domain local operations in the spatial domain.}
    \label{fig: motiv}
\end{figure}

First, we introduce the deformable quad-directional token shift (DQ-Shift) operation, which performs dynamic receptive field adjustment to adapt to local haze variations, overcoming the rigidity of fixed spatial operations. Second, we present the Fourier Mix block, which transforms the WKV attention mechanism to the Fourier domain to facilitate efficient long-range modeling. This design is motivated by two key insights (see Figure~\ref{fig: motiv}): (1) The Fourier representation allows for effective separation of haze interference (primarily in amplitude) from structural content (preserved in phase); (2) The global statistical properties of the Fourier domain naturally establish long-range dependencies, mitigating the information decay in spatial-domain modeling. Furthermore, we propose the Semantic Bridge Module (SBM) to address the inherent semantic gap within the encoder-decoder structure. SBM employs the Dynamic Semantic Kernel Fusion (DSK-Fusion) mechanism to align features between corresponding stages, ensuring semantic consistency throughout the dehazing process.

The integration of these key components enables Fourier-RWKV to realize high-fidelity modeling of non-uniform haze with linear complexity, striking an effective balance between local detail recovery and global physical consistency. The main contributions are summarized as follows:
\begin{itemize}
    \item We introduce Fourier-RWKV, the first multi-state perception dehazing network built upon a linear-complexity RWKV architecture, establishing a new paradigm for efficient and high-quality image restoration.
    \item We propose the DQ-Shift operation, which enables adaptive spatial perception of irregular haze through dynamic receptive field adjustment.
    \item We develop the Fourier Mix block, which extends the core WKV attention mechanism to the Fourier domain to inherently capture global dependencies, effectively mitigating spatial attenuation.
    \item We design SBM, which leverages DSK-Fusion to align encoder-decoder features, ensuring semantic consistency and reducing artifact generation.
    \item Extensive experiments demonstrate that our method surpasses state-of-the-art techniques across multiple benchmarks, yielding superior restoration quality with significantly lower computational cost.
\end{itemize}

\section{Related Work}
\label{sec: related}

\subsection{Prior-based and CNN-based image dehazing}
\label{subsec: prior_cnn_dehaze}
Single-image dehazing is a highly ill-posed problem. Early works simplify this challenge by introducing handcrafted priors~\cite{5567108, 7128396, 7984895, 8101508, Berman_2016_CVPR, 2651362} to constrain the solution space. While effective in simpler scenarios, their rigid assumptions struggle to capture the complexity and variability of real-world haze, leading to poor performance in challenging environments.

The introduction of CNNs marks a paradigm shift from fixed priors to data-driven feature learning, substantially enhancing adaptability across diverse conditions. Early models, such as DehazeNet~\cite{2016DehazeNet}, AODNet~\cite{8237773}, and DCPDN~\cite{8578435}, estimate transmission maps and atmospheric light based on the physical scattering model~\cite{1977Optics, 2002Vision, 2016Haze}. Later methods, including GridDehazeNet~\cite{Liu_2019_ICCV}, FFANet~\cite{qin2020ffa}, and DEA-Net~\cite{10411857}, adopt end-to-end learning strategies to directly restore haze-free images. This transition reduces reliance on physical priors, improving output fidelity. However, the limited receptive fields of convolutions hinder global context modeling, making it challenging to balance global haze estimation and fine detail recovery under non-uniform haze distributions.

\subsection{Transformer-based image dehazing}
\label{subsec: transformer_dehaze}
Transformer-based models~\cite{9879191, wu2024adaptive, 10678109, wen2025all} have advanced the state of the art in image dehazing by effectively modeling long-range dependencies through the self-attention mechanism, greatly improving restoration quality. However, their quadratic computational complexity relative to input resolution limits their applicability in high-resolution and real-time scenarios, prompting a surge of efficiency-oriented designs.

Swin Transformer~\cite{DBLPconficcvLiuL00W0LG21} reduces computational cost by limiting self-attention to non-overlapping local windows. DehazeFormer~\cite{10076399} and Trinity-Net~\cite{10149032} extend this approach to image dehazing, achieving competitive performance with lower overhead. SwinIR~\cite{DBLPconficcvwLiangCSZGT21} further introduces a shifting mechanism for cross-window interactions, mitigating artifact generation. Alternatively, MB-TaylorFormer~\cite{10378631, 10962317} approximates self-attention via Taylor expansion, and Restormer~\cite{Zamir_2022_CVPR} replaces spatial attention with channel attention, both strategies effectively alleviating the computational burden. Despite these developments, most of these methods remain confined to the spatial domain or rely on approximations of self-attention, inevitably sacrificing global modeling accuracy for efficiency. This trade-off has motivated the exploration of more fundamental architectural innovations. 

\subsection{Frequency domain learning}
\label{subsec: frequency_learning}
Frequency-domain learning transforms images into the frequency domain to exploit the global statistics of the spectrum, enabling global dependency modeling with linear complexity and overcoming the efficiency bottleneck inherent in spatial-domain approaches. Moreover, the decoupled and structured nature of frequency-domain representations simplifies the learning of semantically coherent structures from highly entangled local pixels, as different frequency bands naturally correspond to distinct semantic components (e.g., low frequencies for contours and high frequencies for textures). These advantages have rendered frequency-domain learning attractive for fundamental vision tasks, including image deblurring~\cite{10204001}, low-light enhancement~\cite{DBLPconficlrLiGZLZFL23}, and super-resolution~\cite{DBLPconfmmFangZWZCH22}.

The spectral separability of haze degradation (see Figure~\ref{fig: motiv}) further offers a powerful physical prior for dehazing. Leveraging this property, recent frequency-domain dehazing methods~\cite{10678201, CUI2025111074, DBLPjournalscorrabs-2507-11035, gao2023frequency} perform targeted manipulation of frequency components to effectively disentangle haze from underlying image structures. These methods deliver improved dehazing performance with markedly reduced overhead.

\subsection{Receptance Weighted Key Value}
\label{subsec: rwkv_background}
The pursuit of efficient architectures has also driven the rise of linear attention models as viable alternatives to Transformers. Among these, the Receptance Weighted Key Value (RWKV) model~\cite{DBLPconfemnlpPengAAAABCCCDDG23, DBLPjournalscorrabs-2404-05892}, originally designed for NLP, offers notable advantages. Its core WKV mechanism facilitates long-range modeling with linear complexity, while token-shifting enhances local perception. Vision-RWKV~\cite{DBLPconficlrDuanWCZLLQ0DW25} extends this framework to vision tasks by reformulating WKV attention for two-dimensional, non-causal data. It introduces a bidirectional variant (Bi-WKV) to capture global dependencies, while the quad-directional token shift (Q-Shift) aggregates multi-directional local context. Several variants have been developed for specific vision tasks, including URWKV~\cite{DBLPconfcvpr0028NLXLC25} for low-light enhancement, CRWKV~\cite{DBLPconfijcaiChenCJXZW25} for image denoising, Restore-RWKV~\cite{DBLPjournalscorrabs-2407-11087} for medical image restoration, and BSBP-RWKV~\cite{DBLPconfmmZhouC24} for medical image segmentation.

Even with these advances, adapting the linear-complexity architecture of RWKV to image dehazing poses distinct challenges in handling irregular haze distributions, preventing spatial attenuation, and maintaining semantic coherence. This motivates the proposed Fourier-RWKV, which integrates frequency-domain learning with the RWKV architecture to tackle these specific issues.

\section{Method}
\label{sec: meth}

\subsection{Overall Architecture}
\label{subsec: over}
\begin{figure}[th]
  \centering
  \includegraphics[width=0.95\columnwidth]{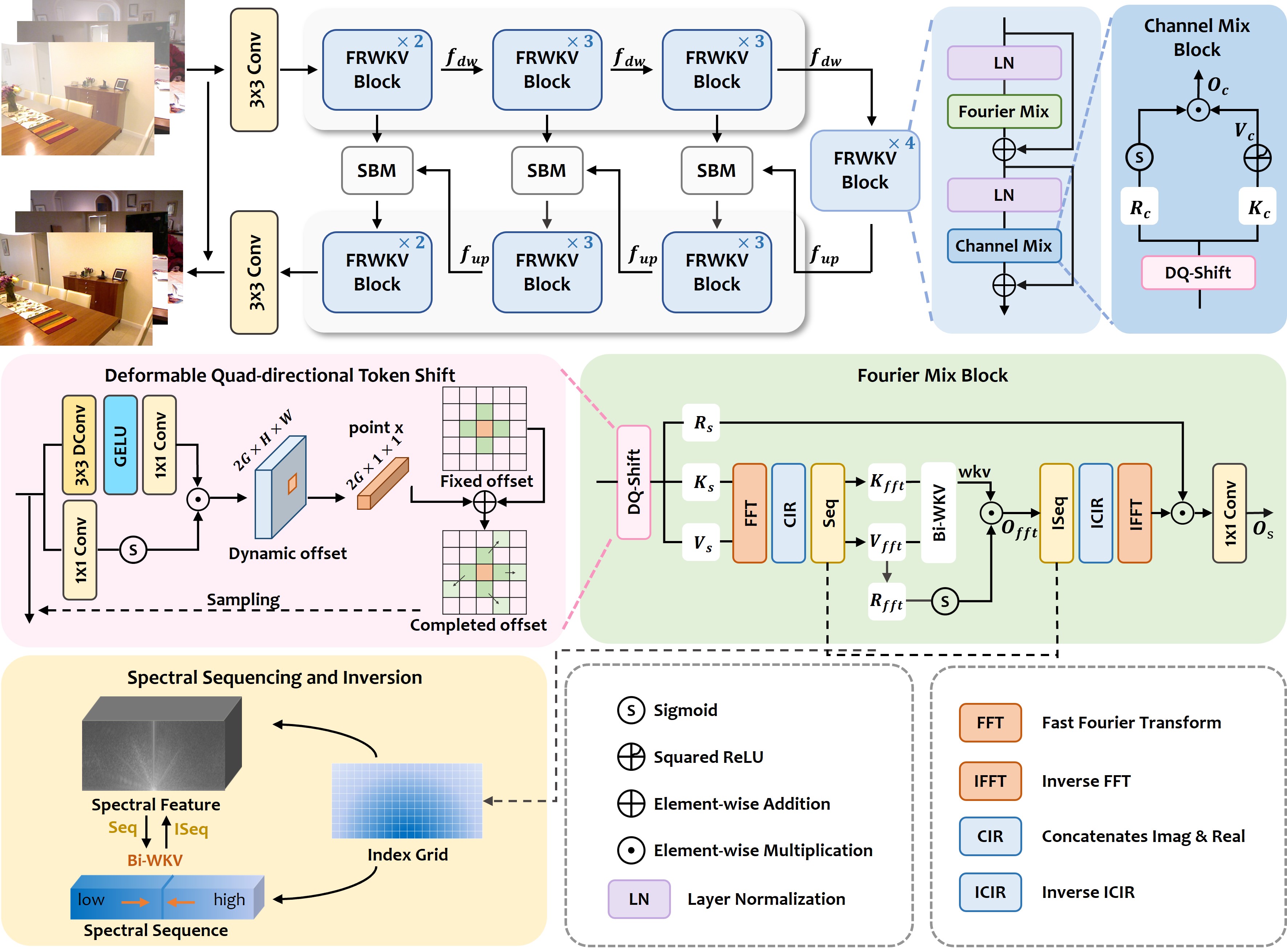}
  \caption{The architecture of the Fourier-RWKV. It adopts a classic encoder-decoder framework with four levels, each stacking $N_i$ FRWKV blocks. Each block consists of a Fourier Mix block and a Channel Mix block, both equipped with a deformable quad-directional token shift (DQ-Shift). Fourier Mix integrates spectral sequencing and inversion (Seq and ISeq), converting spectral features into sequences for the Bi-WKV mechanism. The semantic bridge module (SBM) in the skip connections ensures encoder-decoder feature alignment.}
  \label{fig: net}
\end{figure}

As illustrated in Figure~\ref{fig: net}, Fourier-RWKV employs a symmetric encoder-decoder architecture. Given an input hazy image $I \in \mathbb{R}^{3 \times H \times W}$, the network initially extracts shallow features using a $3 \times 3$ convolution. These features are processed through an encoder-decoder structure with four resolution levels, each containing multiple FRWKV blocks for feature transformation and reconstruction across scales. The output feature $I_{recon}$ is further refined via another $3 \times 3$ convolution to produce a residual image $I_r \in \mathbb{R}^{3\times H \times W}$. The final dehazed result is then obtained as $I_c = I + I_r$.

The FRWKV block serves as the core component of our architecture, augmenting the original RWKV framework with three key innovations. Each block consists of two complementary units: a Fourier Mix block and a Channel Mix block, both incorporating the DQ-Shift operation. The former captures global dependencies through synergistic processes in both the spatial and frequency domains, while preserving local structural integrity. The latter refines feature representations via inter-channel interactions. To optimize information flow between the encoder and decoder, we embed SBM in the skip connections. This module explicitly models the feature correlations between the encoder and decoder at the same scale, ensuring semantic alignment and selective feature fusion for effective information transfer. Moreover, scale transformations are implemented using stride convolutions and pixel shuffling operations to adjust the feature channels and spatial dimensions.

\subsection{FRWKV Block: The Building Block}
In the proposed Fourier-RWKV, features at each resolution level are processed through multiple stacked FRWKV blocks. The FRWKV pipeline can be described as:
\begin{align}
    &X'_n = X_{n-1} + \mathrm{FourierMix}(\mathrm{LN}(X_{n-1})), \\
    &X_n = X'_n + \mathrm{ChannelMix}(\mathrm{LN}(X'_n)),
\end{align}
where $\mathrm{LN}$ denotes layer normalization, $X'_n$ and $X_n$ represent the outputs of the Fourier Mix and Channel Mix blocks, respectively. Notably, both core blocks incorporate DQ-Shift to enhance adaptive spatial awareness for non-uniform haze. By consolidating operations across spatial, frequency, and channel dimensions, the FRWKV block delivers comprehensive haze degradation modeling while maintaining linear complexity.

\subsubsection{Deformable Quad-directional Token Shift}
\label{subsubsec: dqshift}
The original Q-Shift is designed to aggregate local context, but its fixed shift pattern lacks adaptability to irregular haze distributions. To address this issue, we propose the DQ-Shift operation, which retains the fixed shift prior while introducing an input-dependent dynamic offset mechanism. This allows for adaptive receptive field adjustment according to local haze density and structural variations.

We first define four fixed offsets to form the foundation of spatial perception, consistent with the original Q-Shift:
\begin{align}
    \{b_k\}_{k=0}^3=\{[-1,0],[1,0],[0,-1],[0,1]\},
\end{align}
where $b_k$ denotes the fixed offset in the $k$-th direction (up, down, left, right). 

To incorporate spatial adaptability, a lightweight gated CNN is introduced to predict dynamic offsets. Given an input feature $X \in \mathbb{R}^{C \times H \times W}$, we apply a $3\times 3$ depth-wise convolution, followed by GELU activation and a $1\times 1$ convolution to generate offset directions $D$. Concurrently, a separate $1\times 1$ convolution with Sigmoid activation ($\sigma$) is used to produce gating weights $A$. The final offsets $\Delta P$ are computed via element-wise multiplication:
\begin{align}
    &D = \mathrm{Conv}_{1\times 1}\big(\mathrm{GELU}(\mathrm{DConv}_{3\times 3}(X))\big), \\
    &A = \sigma(\mathrm{Conv}_{1\times 1}(X)), \\
    &\Delta P = D \odot A,
\end{align}
where $D, A, \Delta P \in \mathbb{R}^{2G \times H \times W}$, and $G=4$ indicates the number of offset groups corresponding to the four shift directions.

The final sampling coordinates of each position are obtained by combining the fixed and dynamic offsets:
\begin{align}
    P_k(h,w)=[h,w]+b_k+\Delta P_k(h,w),\;k\in\{0,1,2,3\},
\end{align}
where $\Delta P_k(h,w)\in\mathbb{R}^2$ represents the dynamic offset in the $k$-th direction, and $b_k$ provides the fixed offset, contributing to directional spatial perception.

For parameter efficiency, we adopt the same group sampling strategy as in Q-Shift. Thus, the complete DQ-Shift operation is formulated as:
\begin{align}
    &\mathrm{DQ}\text{-}\mathrm{Shift}(X) = X + (1 + \mu_{(*)})X', \\
    &X'[h,w] = \mathrm{Concat}\big( \mathcal{S}(X_0, P_0(h,w)), \mathcal{S}(X_1, P_1(h,w)), \notag \\
    &\qquad\qquad\qquad\qquad \mathcal{S}(X_2, P_2(h,w)), \mathcal{S}(X_3, P_3(h,w)) \big).
\end{align}
where $\mathcal{S}$ denotes the differentiable bilinear sampling operation, and $X_0$ to $X_3$ represent channel groups split evenly from $X$ along the channel dimension. $\mu_{(*)}$ is a learnable parameter vector for branch $(*) \in \{R, K, V\}$, modulating the effect of shifted features during subsequent computations.

\subsubsection{Fourier Mix Block with DQ-Shift}
\label{subsubsec: fftmix}
The WKV attention mechanism in RWKV achieves $O(N)$ complexity via recurrent computation with exponential decay, significantly improving over the $O(N^2)$ complexity of standard Transformers. However, this efficiency comes at the cost of long-range information decay in spatial modeling, which limits accuracy in global haze estimation. To remedy this, we transform the WKV mechanism to the Fourier domain, leveraging its global receptive field to mitigate information decay while exploiting the spectral properties of haze degradation as a physical prior for more consistent haze estimation.

We introduce the Fourier Mix block to replace its spatial counterpart. This module enhances spatial perception via DQ-Shift, followed by three parallel $1 \times 1$ convolutions to produce receptance $R_s$, key $K_s$, and value $V_s$. The Fast Fourier Transform (FFT, $\mathcal{F}$) is then applied to $K_s$ and $V_s$ to map them into the Fourier domain:
\begin{align}
&R_s, K_s, V_s = \mathrm{Conv}_{1\times1}(\mathrm{DQ\text{-}Shift}(X)) \\
&K_{fft} = \mathrm{CIR}(\mathcal{F}(K_s)), \\
&V_{fft} = \mathrm{CIR}(\mathcal{F}(V_s)),
\end{align}
where $\mathrm{CIR}(\cdot)$ refers to channel-wise concatenation of real and imaginary parts, yielding $K_{fft}, V_{fft} \in \mathbb{R}^{2C \times H \times \lceil \frac{W}{2} \rceil}$. Here, $\lceil \cdot \rceil$ denotes the round-up operator.

To align with the sequence modeling structure of Bi-WKV and preserve frequency point relationships, we unfold $K_{fft}$ and $V_{fft}$ using a distance-based sorting strategy. An index grid $M\in\mathbb{R}^{H\times \lceil \frac{W}{2} \rceil}$ is constructed to store the distance of each frequency point from the spectral origin, enabling the sorting of spectral features from low to high frequencies:
\begin{align}
F_{fft} &= \mathrm{Seq}(F_{fft}, M), \; F_{fft} \in \{K_{fft}, V_{fft}\}
\end{align}
where $\mathrm{Seq}(\cdot)$ refers to the operation that flattens the features into a sequence ordered by $M$, resulting in $K_{fft}, V_{fft} \in \mathbb{R}^{2C \times T}$ with sequence length $T = H \times \lceil \frac{W}{2} \rceil$.

In the RWKV framework, receptance serves as a gating mechanism to modulate historical information retention. Directly transforming $R_s$ into the Fourier domain for gating may compromise its frequency selectivity and spatial structural awareness. Conversely, applying the gate solely in the spatial domain would limit the representational capacity in the Fourier domain. To resolve this, we propose a dual-domain gating mechanism: $R_s$ acts as the spatial-domain gate to retain spatial sensitivity, while a linear layer generates the Fourier-domain gate $R_{fft}$ from $V_{fft}$ to regulate long-range dependencies. The complete Fourier-domain RWKV process is formulated as:
\begin{align}
&wkv = \mathrm{Bi\text{-}WKV}(K_{fft}, V_{fft}), \\
&O_{fft} = \sigma(R_{fft}) \odot wkv, \\
&O_{fft} = \mathcal{F}^{-1}(\mathrm{ICIR}(\mathrm{ISeq}(O_{fft}, M))), \\
&O_s = \mathrm{Conv}_{1\times1}(R_s \odot O_{fft}),
\end{align}
where $wkv$ denotes the attention weights from Bi-WKV, $\mathrm{ISeq}(\cdot)$ restores the original frequency grid structure using the index grid $M$, $\mathrm{ICIR}(\cdot)$ reverses channel-wise concatenation to reconstruct complex frequency representations, and $\mathcal{F}^{-1}$ represents the inverse FFT. The final output $O_s$ is obtained by fusing the spatial gate with the Fourier-modulated feature.

\subsubsection{Channel Mix Block with DQ-Shift}
\label{subsubsec: chamix}
The Channel Mix block complements the Fourier Mix block by focusing on cross-channel feature refinement. Building upon the original Vision-RWKV architecture, we replace the Q-Shift with the proposed DQ-Shift to enhance spatial adaptability.

Given an input feature $X\in\mathbb{R}^{C\times H\times W}$, we first apply DQ-Shift to obtain spatially enhanced features. Two parallel $1 \times 1$ convolutions then project these features into receptance $R_c$ and key $K_c$:
\begin{align}
R_c, K_c = \mathrm{Conv}_{1\times1}(\mathrm{DQ\text{-}Shift}(X)),
\end{align}

Subsequently, the features are flattened into a sequence and processed through the channel mixing transformation:
\begin{align}
O_c = (\sigma(R_c) \odot \mathrm{SquaredReLU}(K_c) W_{V_c}) W_{O_c},
\end{align}
where $W_{V_c}$ and $W_{O_c}$ are the weight matrices of the two projection layers, $\mathrm{SquaredReLU}$ represents the squared ReLU activation, and $O_c$ is the output feature after channel mixing.

\subsection{Semantic Bridge Module}
\label{subsec: sbm}
The traditional U-Net framework alleviates gradient vanishing and information loss by employing skip connections to transfer encoder features to the decoder, but it often overlooks the semantic discrepancies between features at different stages. These discrepancies may lead to noise propagation, redundant information transmission, and reconstruction artifacts.

\begin{figure}[t]
    \centering
    \includegraphics[width=0.7\columnwidth]{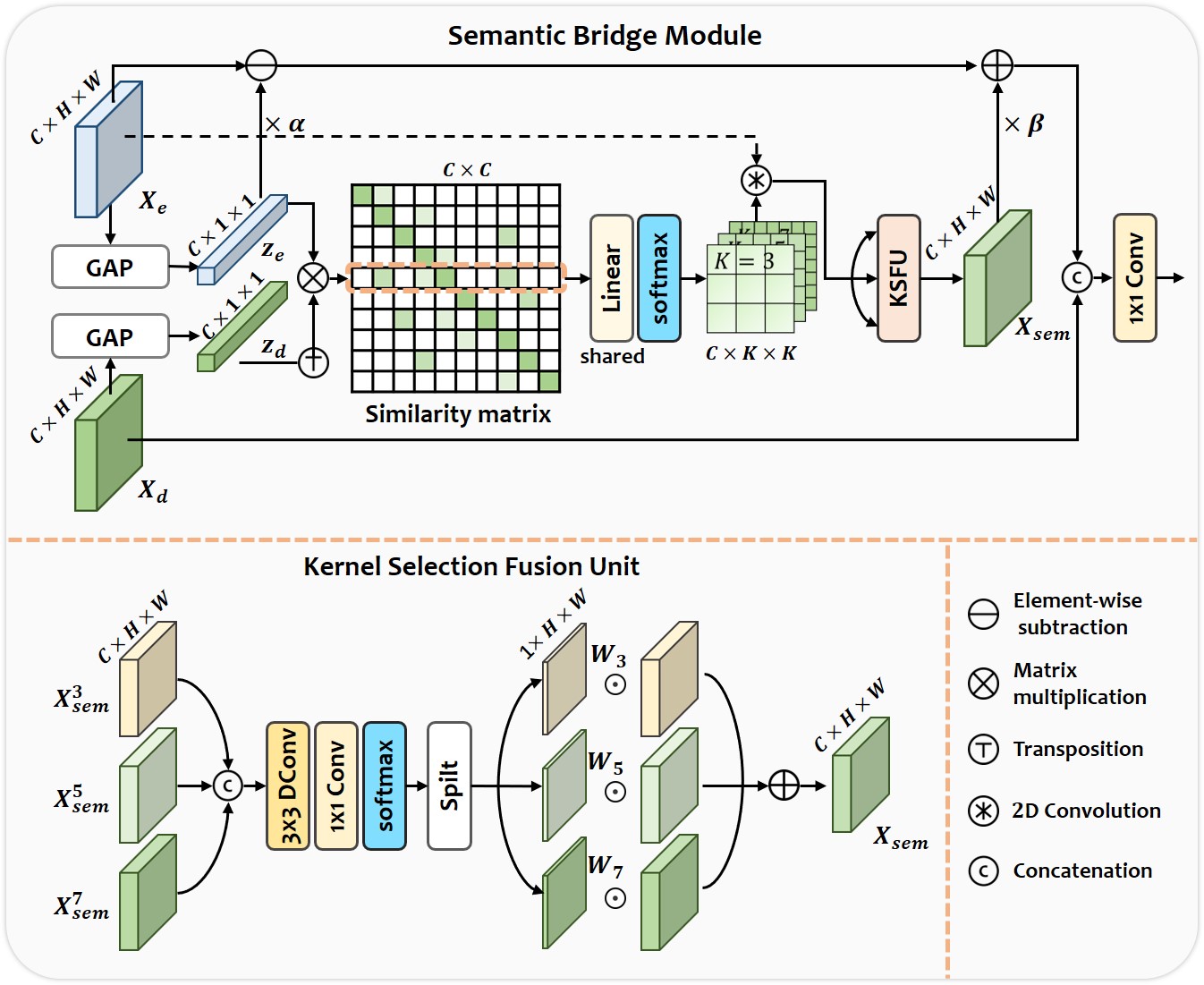}
    \caption{Illustration of the semantic bridge module (SBM). It dynamically generates multi-scale convolutional kernels based on the semantic relationships between encoder and decoder features. The extracted features are fused by Kernel Selection Fusion Unit (KSFU) as semantic priors, which are then used to refine the encoder features through DC component replacement. Subsequently, the refined features are concatenated with the corresponding decoder features to achieve cross-stage semantic alignment.}
    \label{fig: SBM}
\end{figure}

To tackle this issue, we propose the semantic bridge module (SBM), a lightweight fusion module that semantically aligns corresponding encoder and decoder features. The SBM structure is illustrated in Figure~\ref{fig: SBM}. At its core lies the dynamic semantic kernel fusion (DSK-Fusion) mechanism, which generates channel-adaptive convolution kernels by explicitly modeling inter-feature semantic correlations.

First, global average pooling (GAP) is applied to the encoder feature $X_e$ and the corresponding decoder feature $X_d$ (both in $\mathbb{R}^{C\times H \times W}$) to extract their channel-wise semantic descriptors $\mathbf{z}_e, \mathbf{z}_d\in\mathbb{R}^C$, which capture global semantic and structural information. The semantic similarity matrix $\mathbf{Sim}\in \mathbb{R}^{C \times C}$ is then computed as:
\begin{align}
\mathbf{Sim} = \mathbf{z}_e \mathbf{z}_d^\top,
\end{align}
where each element $\mathbf{Sim}_{i,j}$ indicates the semantic correlation between the $i$-th channel of the encoder and the $j$-th channel of the decoder. Given that haze degradation is concentrated in low-frequency components (Figure~\ref{fig: motiv}), the similarity matrix also reflects inter-channel correlations in degradation patterns.

Next, a channel-shared linear layer converts the similarity matrix into multi-scale dynamic convolution kernels:
\begin{align}
F^k= \mathrm{Softmax}(\mathrm{Linear}(\mathbf{Sim})),
\end{align}
where $F^k\in\mathbb{R}^{C \times (k \times k)}$ represents the dynamic kernel of size $k \times k$, with $k = \{3, 5, 7\}$ capturing multi-scale context. These dynamic kernels are applied to the encoder features for multi-scale semantic enhancement, expressed as:
\begin{align}
X_{sem}^k(c,i,j) = \sum_{m,n} X_e(c,i+m,j+n)\cdot F^k(c,m,n),
\end{align}
where $m,n \in [-r,r]$, and $r = \lfloor \frac{k}{2} \rfloor$, with $X_{sem}^k \in \mathbb{R}^{C \times H \times W}$ being the semantic-enhanced feature at scale $k$. Here, $\lfloor \cdot \rfloor$ denotes the round-down operator.

We also introduce a Kernel Selection Fusion Unit (KSFU) to adaptively fuse the multi-scale features via a spatial attention mechanism, which is defined as:
\begin{align}
&X_{con} = \mathrm{Concat}(X_{sem}^3, X_{sem}^5, X_{sem}^7), \\
& W = \mathrm{Softmax}(\mathrm{Conv}_{1\times1}(\mathrm{DConv}_{3\times3}(X_{con}))), \\
& X_{sem} = \sum_{k} W_k \odot X_{sem}^k,
\end{align}
where $W \in \mathbb{R}^{3 \times H \times W}$ contains the spatial attention maps for the three scales ($W_k$ for scale $k$), and $X_{sem} \in \mathbb{R}^{C \times H \times W}$ is the fused semantic-enhanced feature.

Finally, semantic replacement is performed by substituting the DC component of the encoder features with the fused semantic-enhanced feature, resulting in the aligned encoder feature. It is then fused with the decoder feature via a $1 \times 1$ convolution to produce the semantically consistent output:
\begin{align}
&\hat{X}_e = X_e - \alpha \cdot \mathrm{GAP}(X_e) + \beta \cdot X_{sem}, \\
&X_{fuse} = \mathrm{Conv}_{1\times1}(\mathrm{Concat}(\hat{X}_e, X_d)),
\end{align}
where $\alpha, \beta \in \mathbb{R}^C$ are learnable scaling parameters that control the relative weighting of the original and semantic-enhanced features, and $X_{fuse}$ is the final output.

\subsection{Loss Function}
\label{subsec: loss}
To achieve high-quality image restoration that balances local texture details with global structural coherence, we define the optimization objective as a dual-domain loss function. This function combines $L_1$ penalties in both the spatial and frequency domains, enforcing supervision at the pixel and component levels. The total loss is expressed as:
\begin{gather}
\mathcal{L} = \|I_c - I_g\|_1 + \lambda\|\mathcal{F}(I_c)-\mathcal{F}(I_g)\|_1,
\end{gather}
where $I_c$ and $I_g$ denote the dehazed result and ground-truth image, respectively. The term $\lambda$ is a weighting factor that balances the contributions of the two domains, set to 0.15.

\begin{table}[tbp]
  \centering
  \caption{Quantitative comparisons with state-of-the-art dehazing methods on the synthetic and real-world datasets.}
  \resizebox{\textwidth}{!}{
  \renewcommand\arraystretch{1.2}
  \begin{tabular}{c|c|cc|cc|cc|cc|cc}
  \toprule
  \multicolumn{1}{c|}{\multirow{1}[4]{*}{Method}} & \multicolumn{1}{c|}{\multirow{1}[4]{*}{Venue}} & \multicolumn{2}{c|}{SOTS-Indoor} & \multicolumn{2}{c|}{SOTS-Outdoor} &\multicolumn{2}{c|}{Dense-Haze} &\multicolumn{2}{c|}{NH-HAZE} &\multicolumn{1}{c}{Params} &\multicolumn{1}{c}{FLOPs}\\
  & & \multicolumn{1}{c}{PSNR$\uparrow$} & \multicolumn{1}{c|}{SSIM$\uparrow$} & \multicolumn{1}{c}{PSNR$\uparrow$} & \multicolumn{1}{c|}{SSIM$\uparrow$} & \multicolumn{1}{c}{PSNR$\uparrow$} & \multicolumn{1}{c|}{SSIM$\uparrow$} & \multicolumn{1}{c}{PSNR$\uparrow$} & \multicolumn{1}{c|}{SSIM$\uparrow$} & \multicolumn{1}{c}{(M)} & \multicolumn{1}{c}{(G)}\\
  \midrule
  GridDehazeNet~\cite{Liu_2019_ICCV}  &ICCV'19 &32.16 &0.984 &30.86 &0.982 &13.31 &0.37 &13.80 &0.54 &0.956 &21.49  \\ 
  MSBDN~\cite{dong2020multi}   &CVPR'20 &33.67 &0.985 &33.48 &0.982 &15.37 &0.49  &19.23 &0.71 &31.35 &41.54 \\ 
  FFA-Net~\cite{qin2020ffa} &AAAI'20 &36.39 &0.989 &33.57 &0.984  &14.39 &0.45 &19.87 &0.69 &4.456 &287.8 \\ 
  PMNet~\cite{ye2022perceiving}  &ECCV'22 &38.41 &0.990 &34.74 &0.985 &16.79 &0.51 &20.42 &0.73 &18.90 &81.13 \\ 
  MAXIM-2S~\cite{tu2022maxim} &ICCV'22 &38.11 &0.991 &34.19 &0.985 &-  &- &- &- &14.10 &216.0 \\
  Dehamer~\cite{9879191} &CVPR'22 &36.63 &0.988 &35.18 &0.986 &16.62 &0.56 &\underline{20.66} &0.68 &132.50 &60.3 \\
  Fourmer~\cite{zhou2023fourmer} &ICML'23 &37.32 &0.990 &- &- &15.95 &0.49 &- &- &1.29 &20.6 \\
  DehazeFormer~\cite{10076399} & TIP'23 &40.05 &\underline{0.996} &34.29 &0.983 &- &- &19.11 &0.66 &4.634 &48.64 \\
  FocalNet~\cite{10377428} &ICCV'23 &40.82 &\underline{0.996} &\underline{37.71} &\underline{0.995} &17.07 &0.63 &20.43 &0.79 &3.74 &30.63 \\
  MB-TaylorFormer-B~\cite{10378631} &ICCV'23 &40.71 &0.992 &37.42 &0.989 &16.66 &0.56 &20.43 &0.69 &2.68 &38.5 \\
  DEA-Net~\cite{10411857} &TIP'24 &40.20 &0.993 &36.03 &0.989 &- &- &- &- &3.65 &34.19 \\
  OKNet~\cite{cui2024omni} &AAAI'24 &40.79 &\underline{0.996} &37.68 &\underline{0.995} &16.92 &0.64 &20.48 &\underline{0.80} &4.72 &39.71 \\
  CSNet~\cite{cui2024hybrid} &IJCAI'24 &41.34 &\underline{0.996} &- &- &\underline{17.33} &\underline{0.65} &20.43 &\underline{0.80} &4.27 &41.19 \\
  MAIR~\cite{Li_2025_CVPR} &CVPR'25 &39.45 &\textbf{0.997} &36.96 &0.991 &- &- &- &- &3.40 &24.03 \\
  PGH2Net~\cite{su2025prior} &AAAI'25 &\textbf{41.70} &\underline{0.996} &37.52 &0.989 &17.02 &0.61 &- &- &1.76 &16.05 \\
  \midrule
  Fourier-RWKV &-  &\underline{41.60} &\underline{0.996} &\textbf{39.76} &\textbf{0.996} &\textbf{17.34} &\textbf{0.66} &\textbf{21.01} &\textbf{0.83} &5.31 &15.69 \\
  \bottomrule
  \end{tabular}}
  \label{tab: table1}
\end{table}

\section{Experiments}
\label{sec: exp}

\subsection{Experimental settings}
\label{subsec: set}

\textbf{Datasets.}
To thoroughly evaluate the dehazing performance and generalization ability of the proposed Fourier-RWKV, we conduct experiments on both widely-used synthetic datasets and challenging real-world datasets.
For synthetic data, we utilize the RESIDE~\cite{li2018benchmarking} benchmark dataset, which covers indoor and outdoor scenes with various haze conditions. Specifically, the model is independently trained on the indoor training subset (ITS, containing 13,990 image pairs) and the outdoor training subset (OTS, containing 313,950 image pairs), with performance evaluated on 500 indoor and 500 outdoor scenes from the SOTS test set, respectively.
For real-world data, we select two representative haze datasets, Dense-Haze~\cite{ancuti2019dense} and NH-HAZE~\cite{9150807}. Dense-Haze focuses on dense haze conditions, while NH-HAZE deals with non-uniform haze distributions. Both datasets consist of 55 pairs of hazy and clear images. The first 50 pairs from each dataset are used for training, and the remaining 5 pairs are reserved for testing, ensuring consistent and comparable results.

\textbf{Implementation details.} For the model architecture, we use the configuration \{2, 3, 3, 4, 3, 3, 2\} for the number of FRWKV blocks at each stage. The base channel size is set to 24, striking a balance between performance and parameter efficiency. We use the Adam optimizer with $\beta_1 = 0.9$ and $\beta_2 = 0.999$ during training. The initial learning rate is adjusted according to the specific dataset, and we employ a cosine annealing strategy to gradually decay it to $1 \times 10^{-6}$. The computational complexity (FLOPs) is calculated using $256 \times 256$ image patches as input. To improve generalization, we apply data augmentation techniques such as random cropping and flipping. For synthetic datasets, we train on $256 \times 256$ patches with a batch size of 8, while for real-world datasets, we use $512 \times 512$ patches with a smaller batch size of 2. All experiments are conducted on NVIDIA GeForce RTX 3090 GPUs using the PyTorch framework.

\textbf{Evaluation Metrics.}
To quantitatively evaluate the dehazing performance of Fourier-RWKV, we use Peak Signal-to-Noise Ratio (PSNR)~\cite{1284395} and Structural Similarity Index (SSIM)~\cite{1284395} as evaluation metrics. PSNR primarily reflects the fidelity of the dehazed result at the pixel level, while SSIM focuses on assessing the structural similarity between the dehazed image and the reference clear image in terms of visual perception. The combination of these two metrics provides a comprehensive assessment of restoration quality.

\subsection{Comparison with State-of-the-art Methods}
\label{subsec: comp}
In this section, we compare Fourier-RWKV with a wide range of state-of-the-art dehazing approaches, including CNN-based methods (GridDehazeNet~\cite{Liu_2019_ICCV}, MSBDN~\cite{dong2020multi}, FFA-Net~\cite{qin2020ffa}, PMNet~\cite{ye2022perceiving}, MAXIM-2S~\cite{tu2022maxim}, DEA-Net~\cite{10411857}, FocalNet~\cite{10377428}, OKNet~\cite{cui2024omni}, CSNet~\cite{cui2024hybrid}, PGH2Net~\cite{su2025prior}), recent Transformer-based models (DeHamer~\cite{9879191}, DehazeFormer~\cite{10076399}, Fourmer~\cite{zhou2023fourmer}, MB-TaylorFormer-B~\cite{10378631}), and a Mamba-based method (MAIR~\cite{Li_2025_CVPR}). Both quantitative and qualitative results are discussed below.

\begin{figure}[t]
    \scriptsize
    \centering
    \renewcommand{\tabcolsep}{1pt} 
    \renewcommand{\arraystretch}{1}
    
    \begin{tabular}{cccc}
        \includegraphics[width=0.195\linewidth]{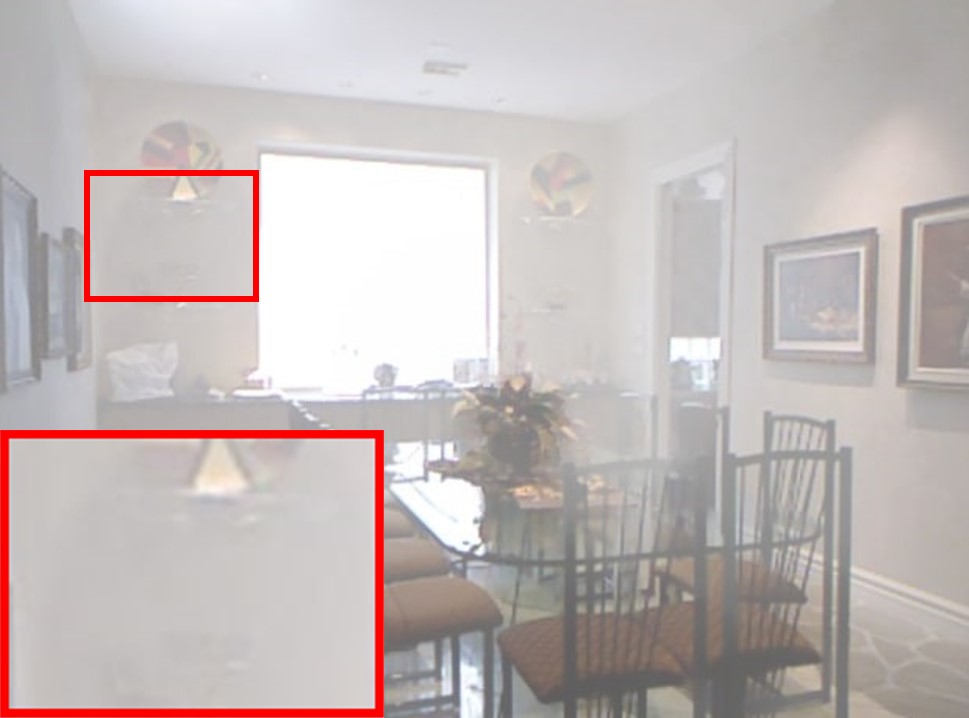} &
        \includegraphics[width=0.195\linewidth]{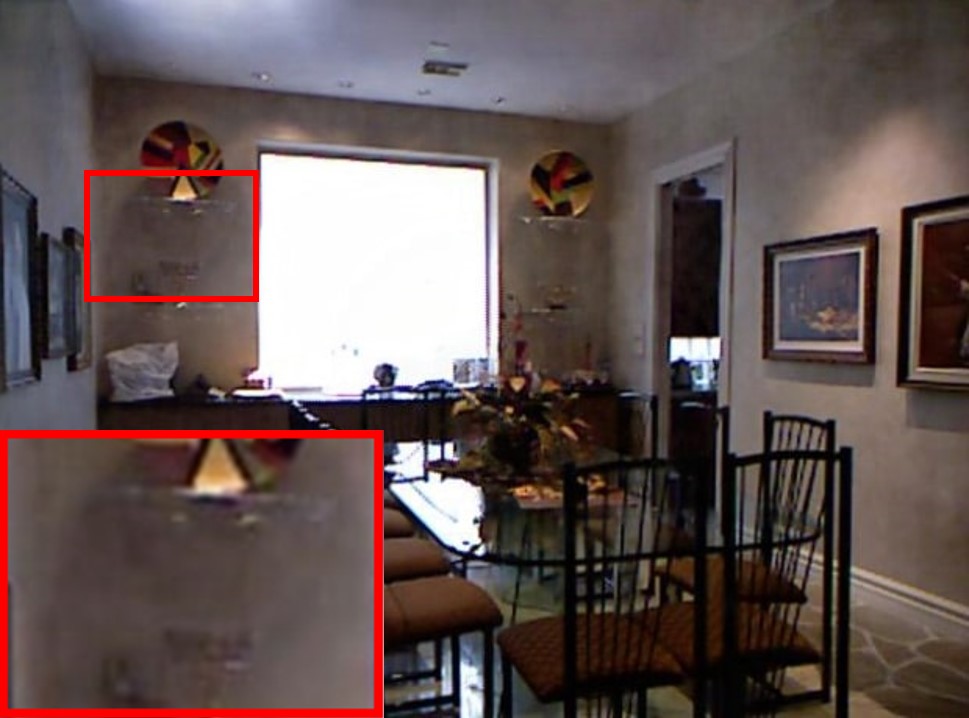} &
        \includegraphics[width=0.195\linewidth]{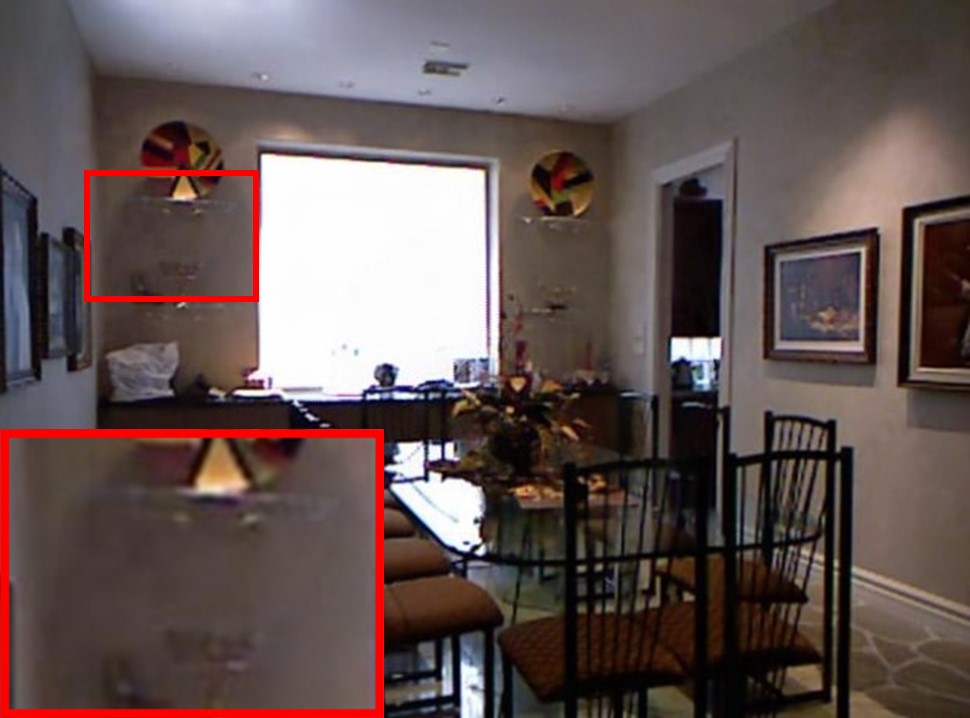} &
        \includegraphics[width=0.195\linewidth]{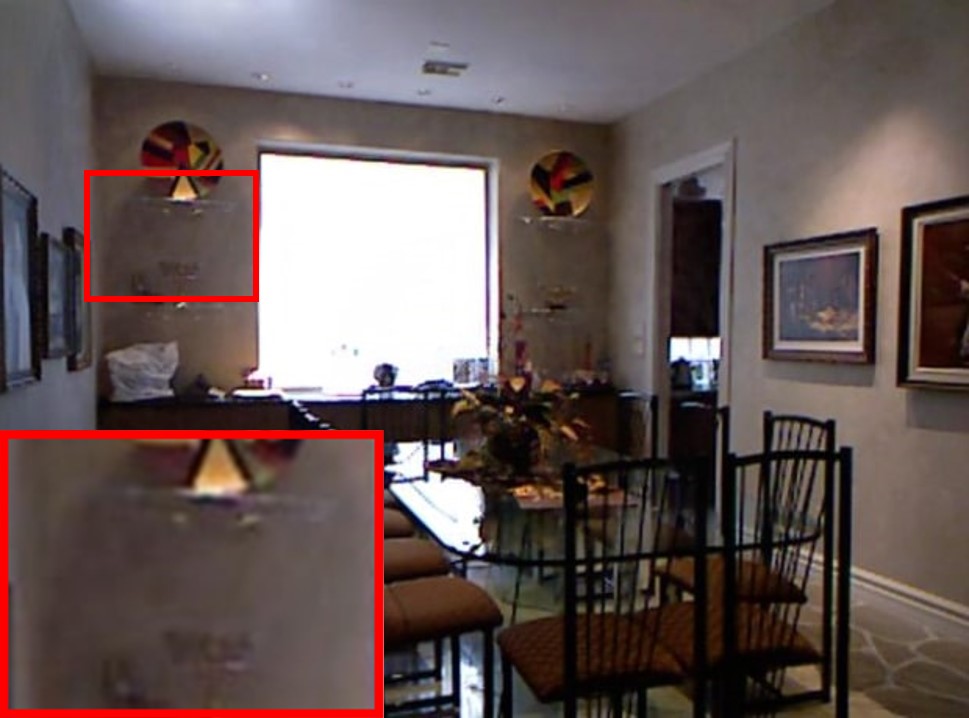} \\
        
        7.62 dB & 29.98 dB & 33.95 dB & 32.91 dB \\
        
        \includegraphics[width=0.195\linewidth]{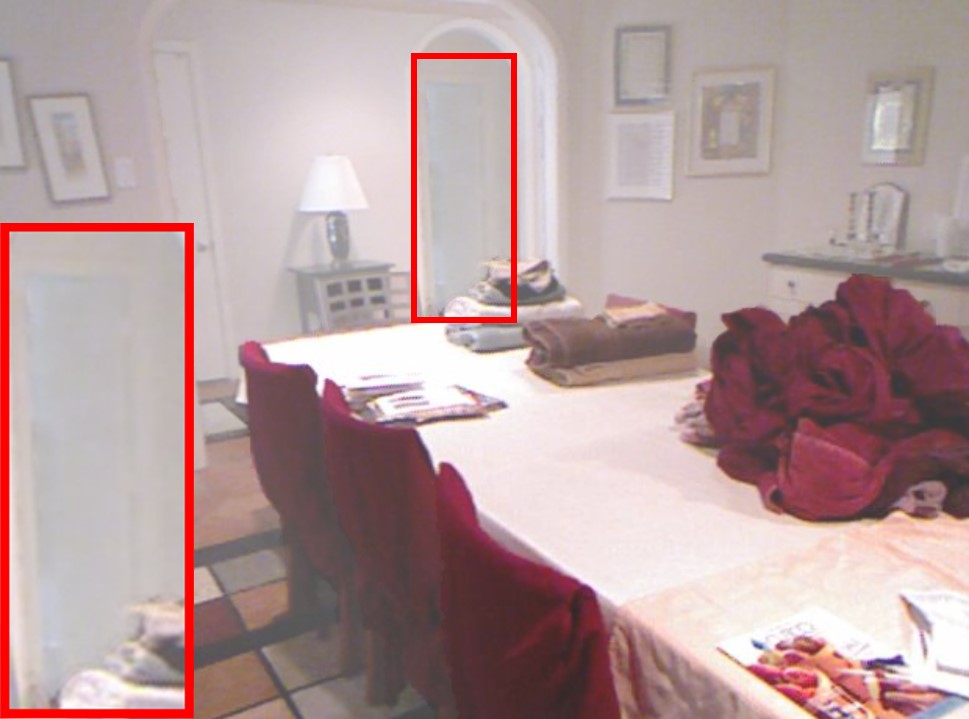} &
        \includegraphics[width=0.195\linewidth]{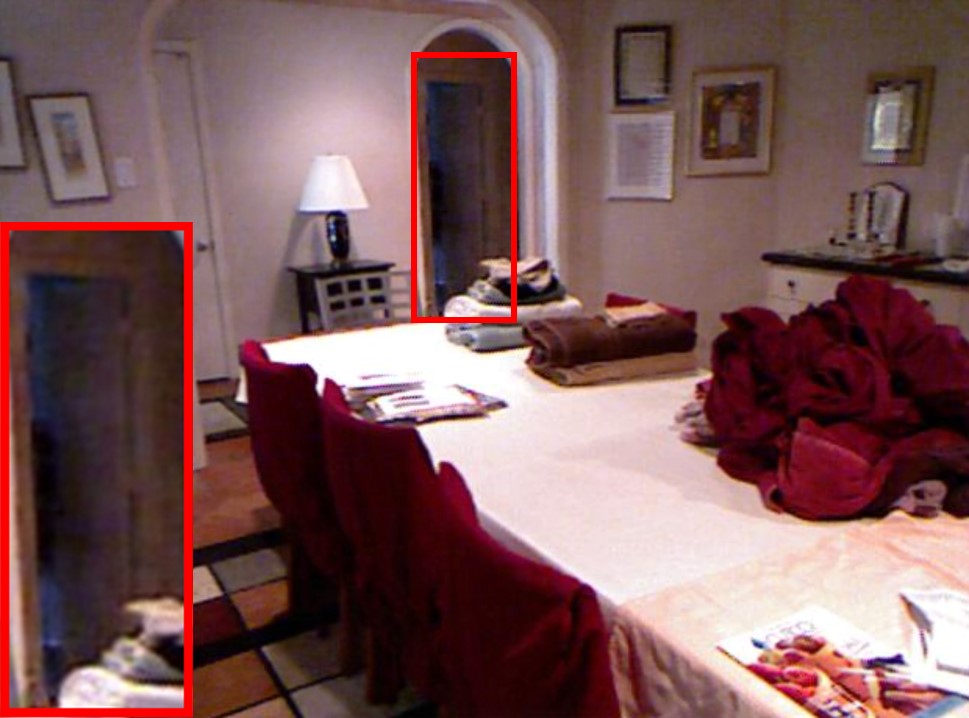} &
        \includegraphics[width=0.195\linewidth]{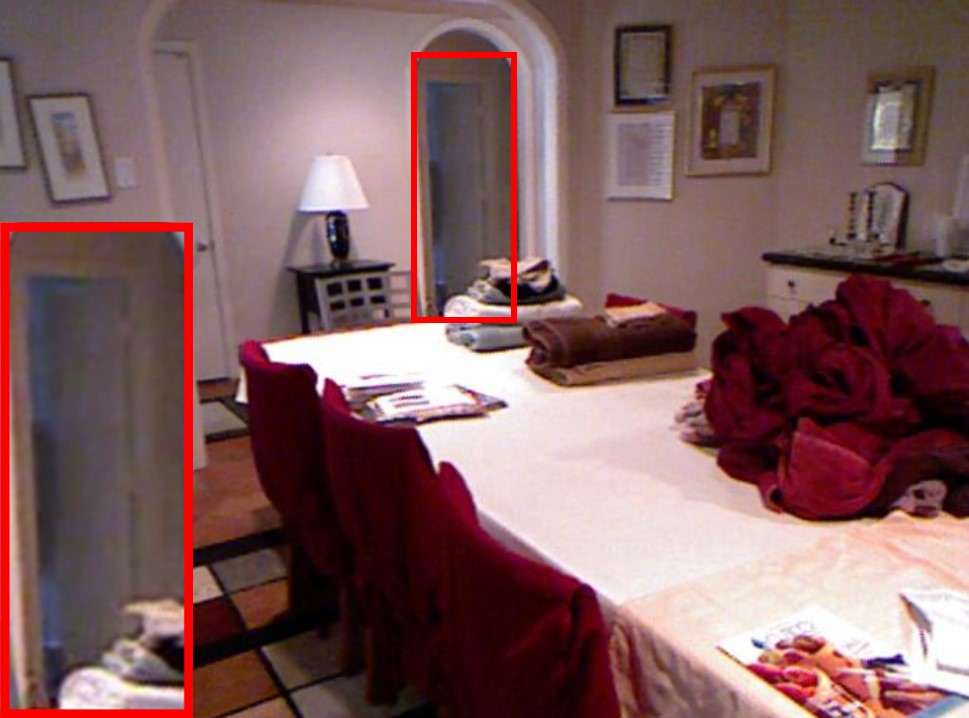} &
        \includegraphics[width=0.195\linewidth]{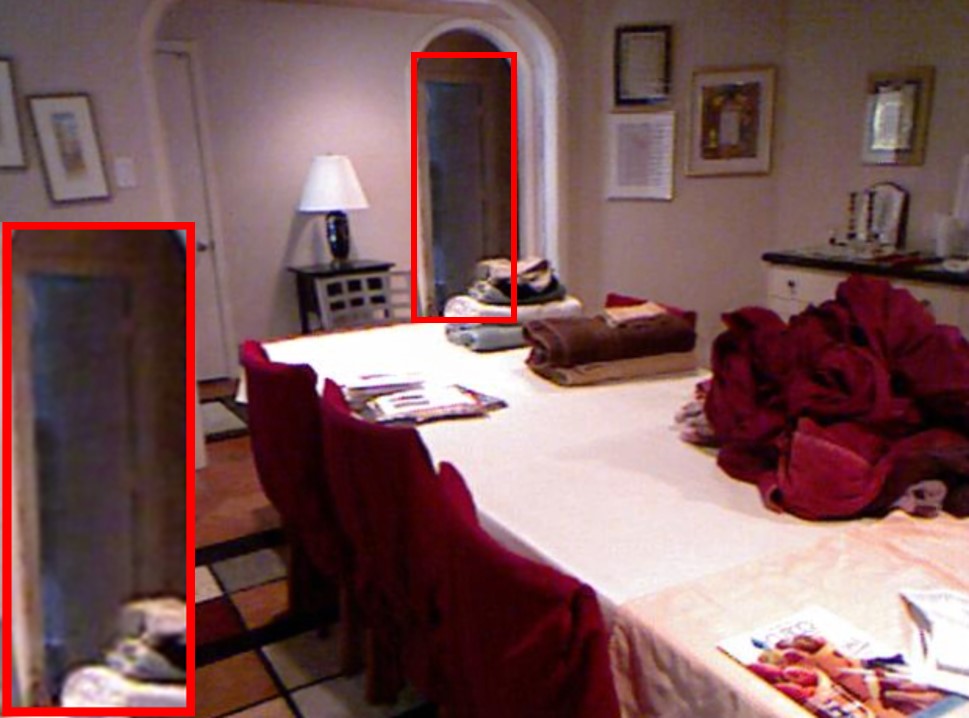} \\
        
        11.54 dB & 34.49 dB & 30.01 dB & 34.38 dB \\
        (a) Hazy image & (b) FFA-Net~\cite{qin2020ffa} & (c) MAXIM-2S~\cite{tu2022maxim} & (d) Dehamer~\cite{9879191}
    \end{tabular}
    \hspace{1em}
    \begin{tabular}{cccc}
        \includegraphics[width=0.195\linewidth]{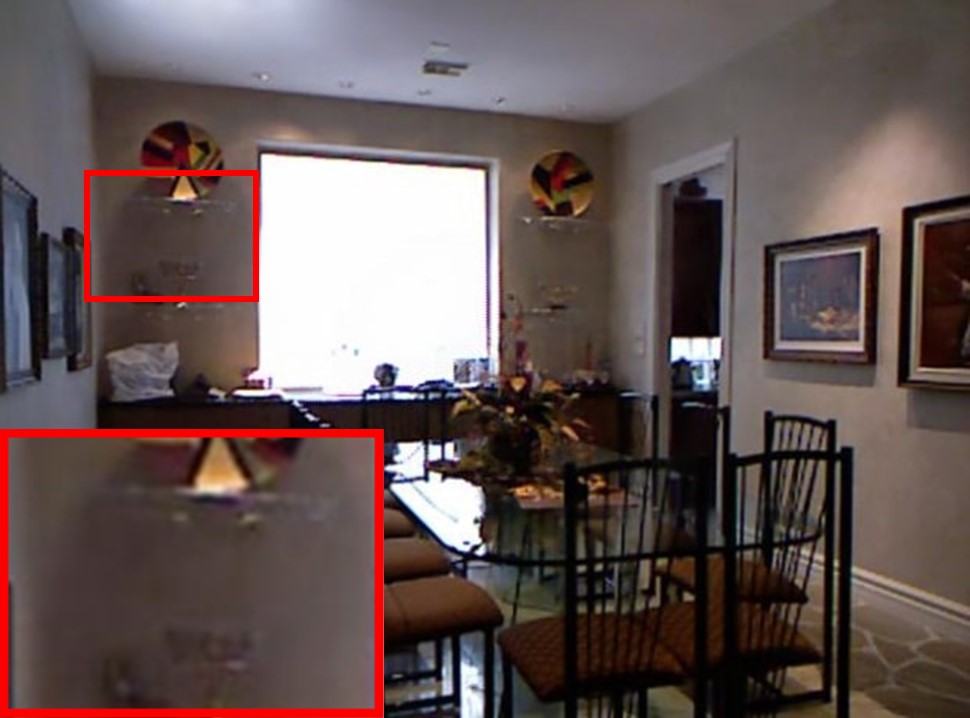} &
        \includegraphics[width=0.195\linewidth]{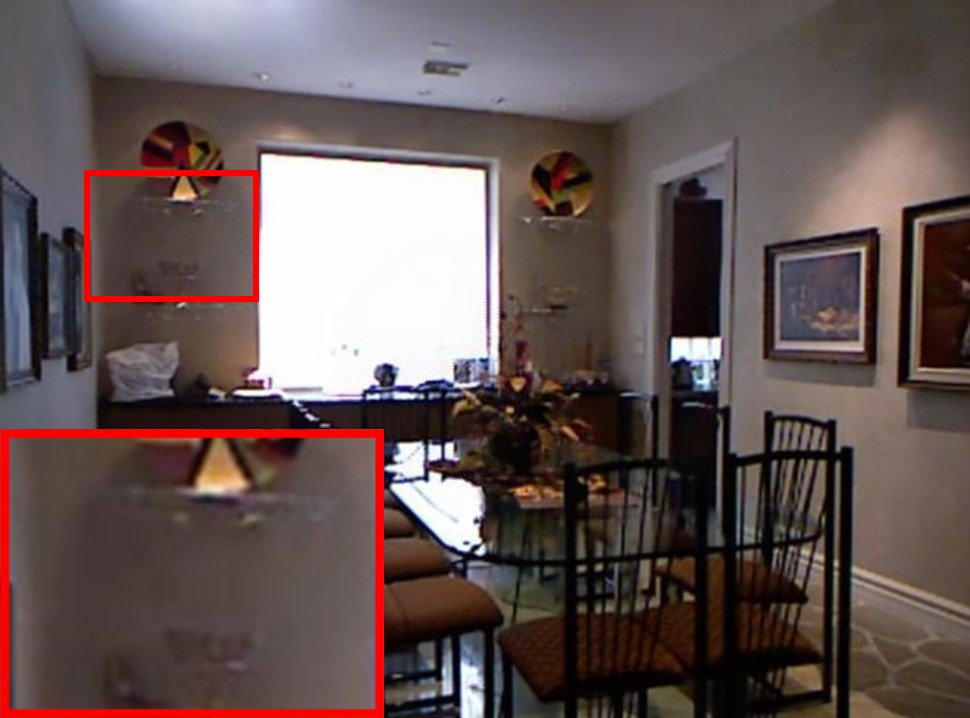} &
        \includegraphics[width=0.195\linewidth]{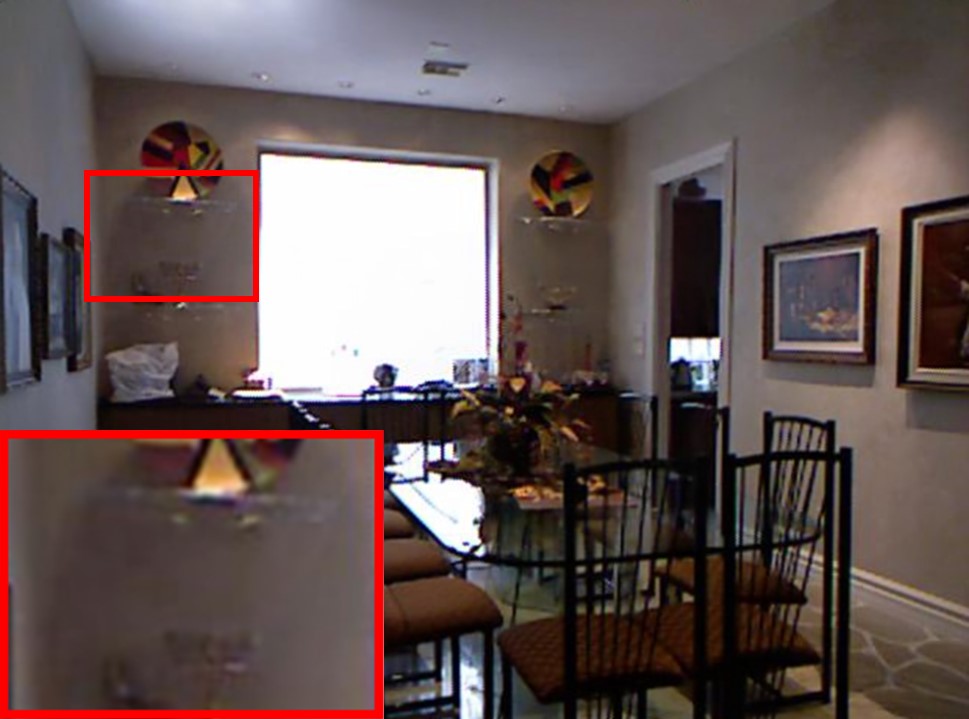} &
        \includegraphics[width=0.195\linewidth]{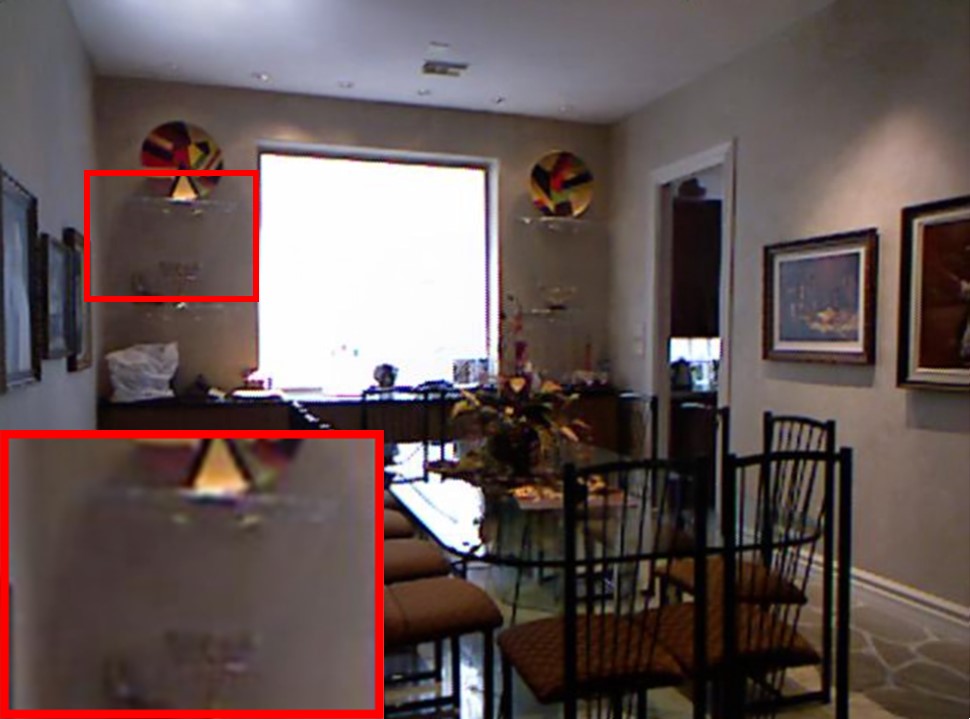} \\
        
        36.01 dB & 35.74 dB & 37.19 dB & PSNR \\
        
        \includegraphics[width=0.195\linewidth]{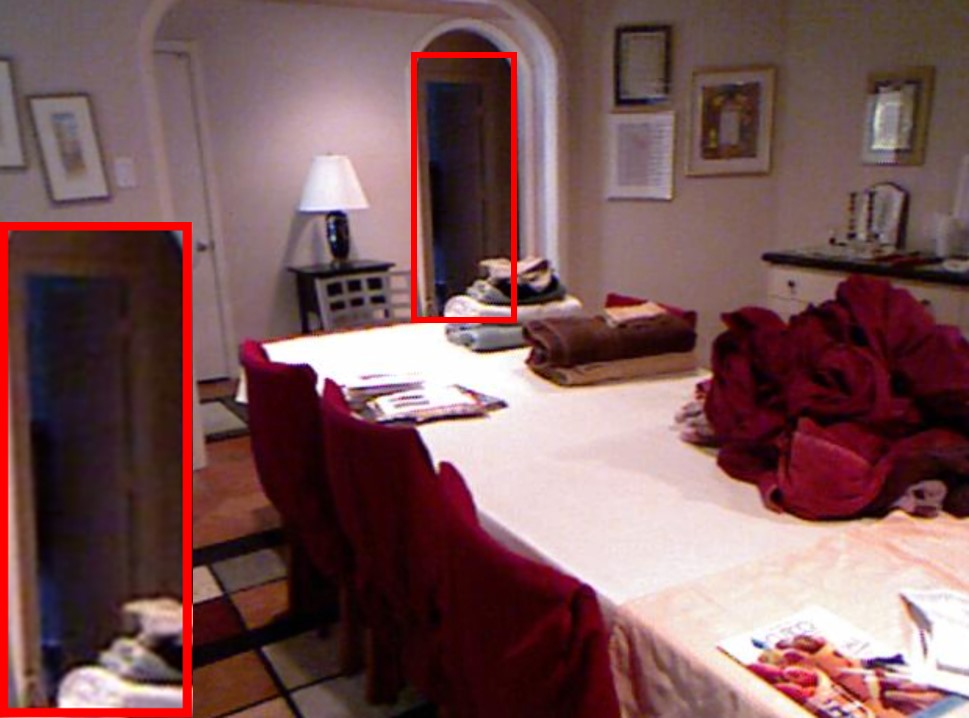} &
        \includegraphics[width=0.195\linewidth]{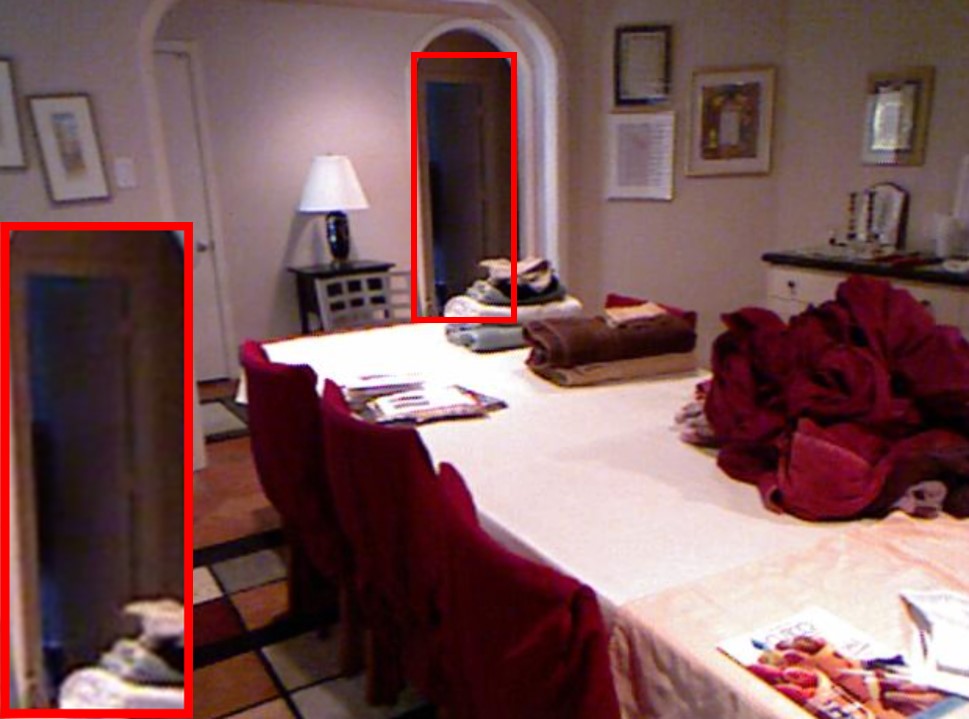} &
        \includegraphics[width=0.195\linewidth]{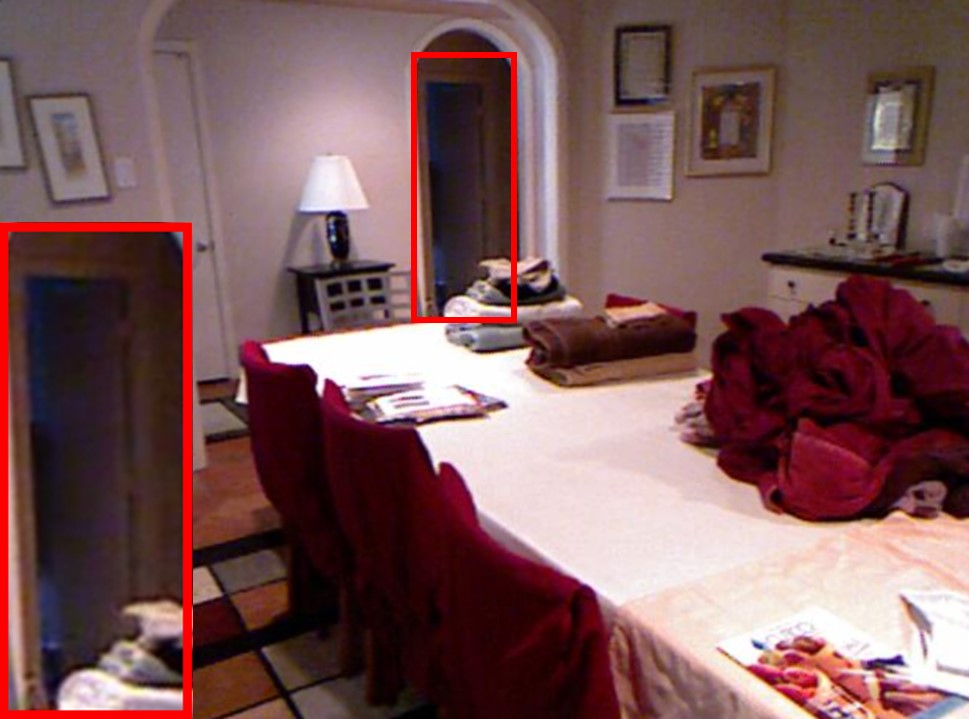} &
        \includegraphics[width=0.195\linewidth]{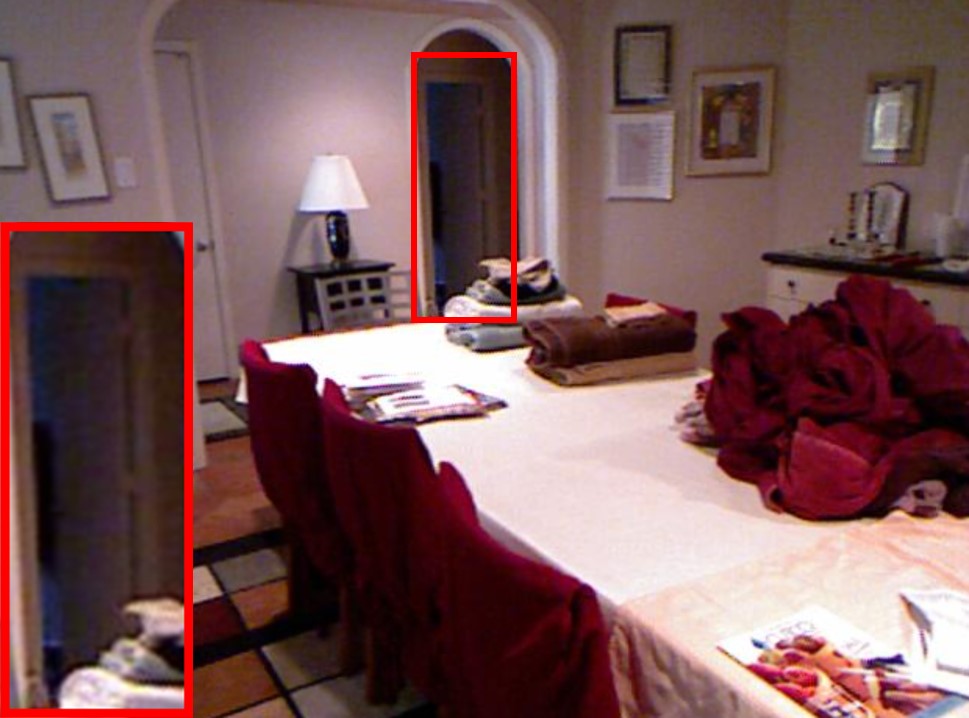} \\
        
        40.42 dB & 40.29 dB & 40.96 dB & PSNR \\
        
        (e) FocalNet~\cite{10377428} & (f) OKNet~\cite{cui2024omni} & (g) Ours & (h) GT
    \end{tabular}
    
    \caption{Visual comparisons on synthetic hazy images from the SOTS-Indoor dataset. Key regions highlighted by red boxes are enlarged in the lower-left corner for clearer comparison.}
    \label{fig: Indoor}
\end{figure}

\subsubsection{Quantitative Comparisons}
\label{subsubsec: quan}
The quantitative comparison of Fourier-RWKV against SOTA models is summarized in Table~\ref{tab: table1}, where the best and second-best results are indicated in bold and underlined, respectively. Fourier-RWKV demonstrates outstanding performance across all test sets, with significant improvements in both PSNR and SSIM metrics.

\begin{figure}[t!]
    \scriptsize
    \centering
    \renewcommand{\tabcolsep}{1pt} 
    \renewcommand{\arraystretch}{1}
    
    \begin{tabular}{cccc}
        \includegraphics[width=0.195\linewidth]{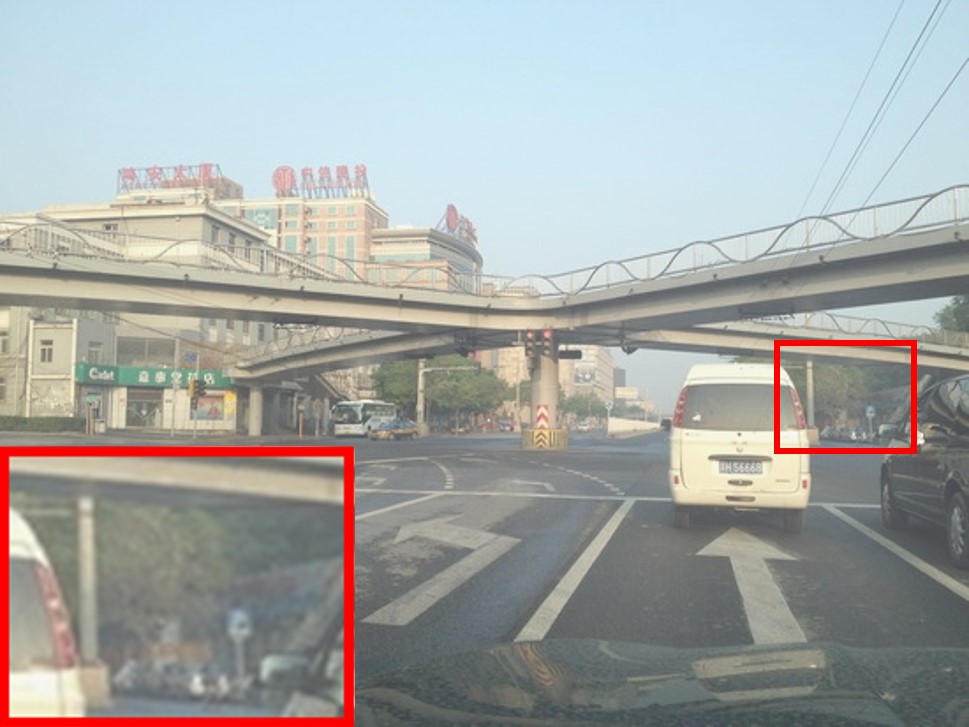} &
        \includegraphics[width=0.195\linewidth]{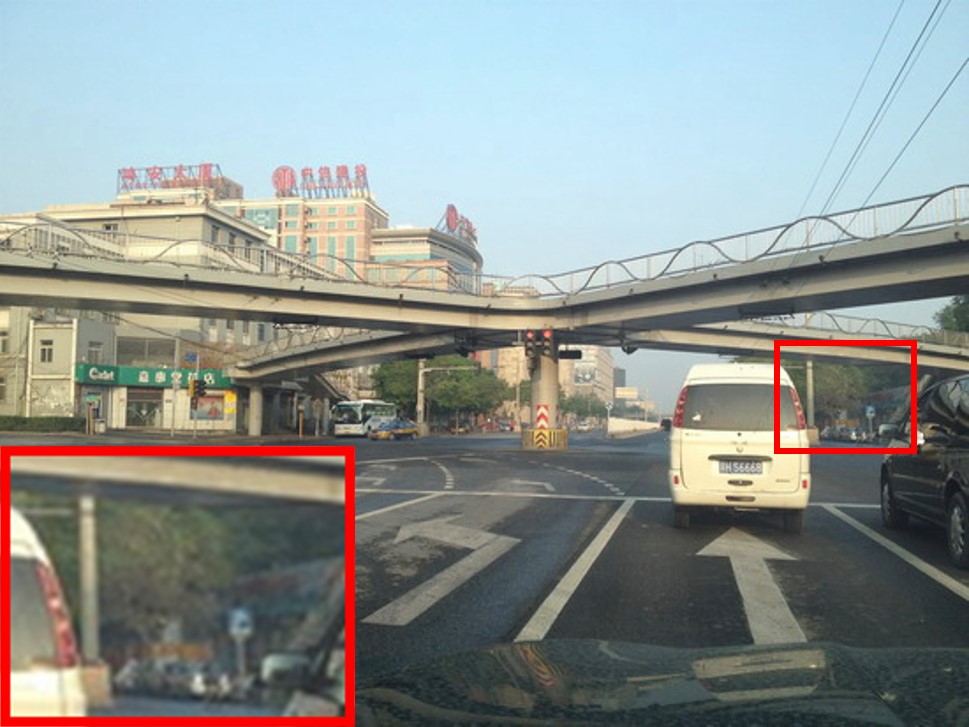} &
        \includegraphics[width=0.195\linewidth]{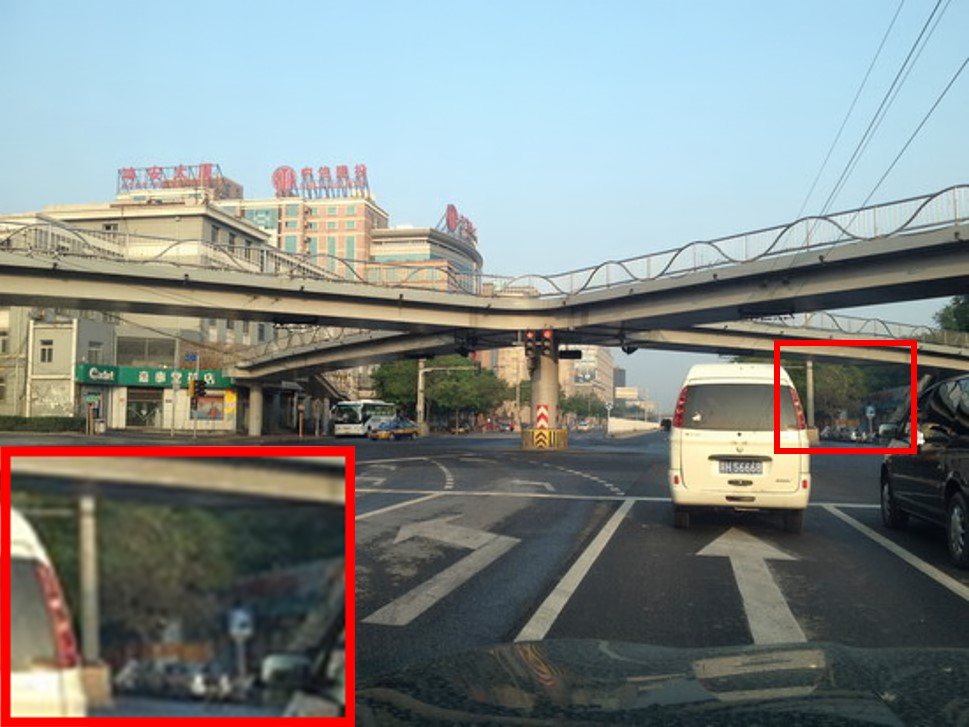} &
        \includegraphics[width=0.195\linewidth]{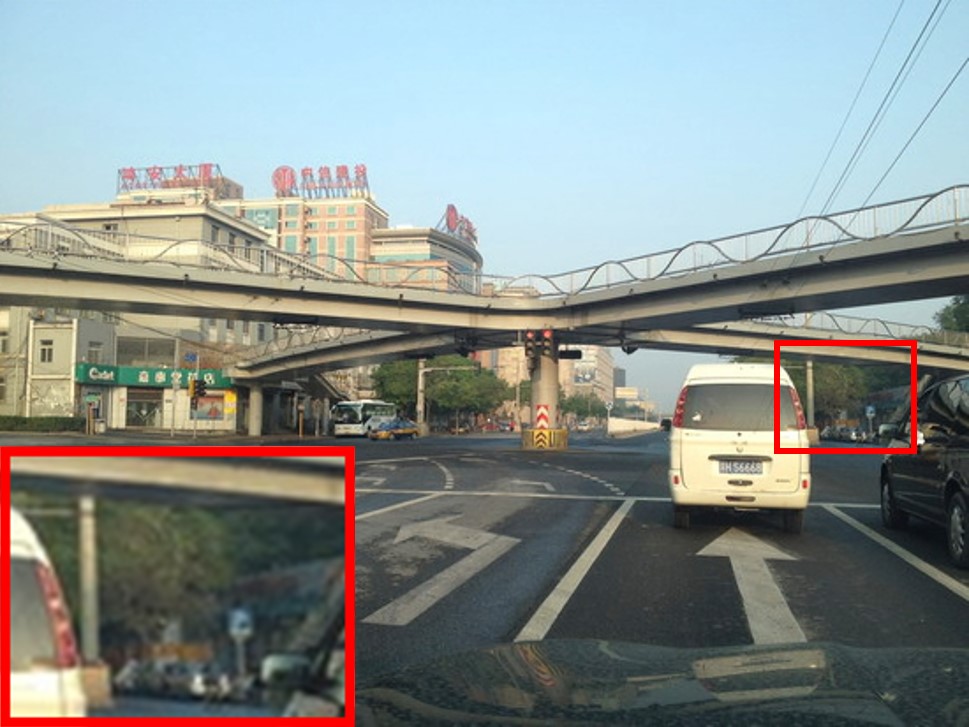} \\
        
        16.49 dB & 28.10 dB & 34.53 dB & 33.31 dB \\
        
        \includegraphics[width=0.195\linewidth]{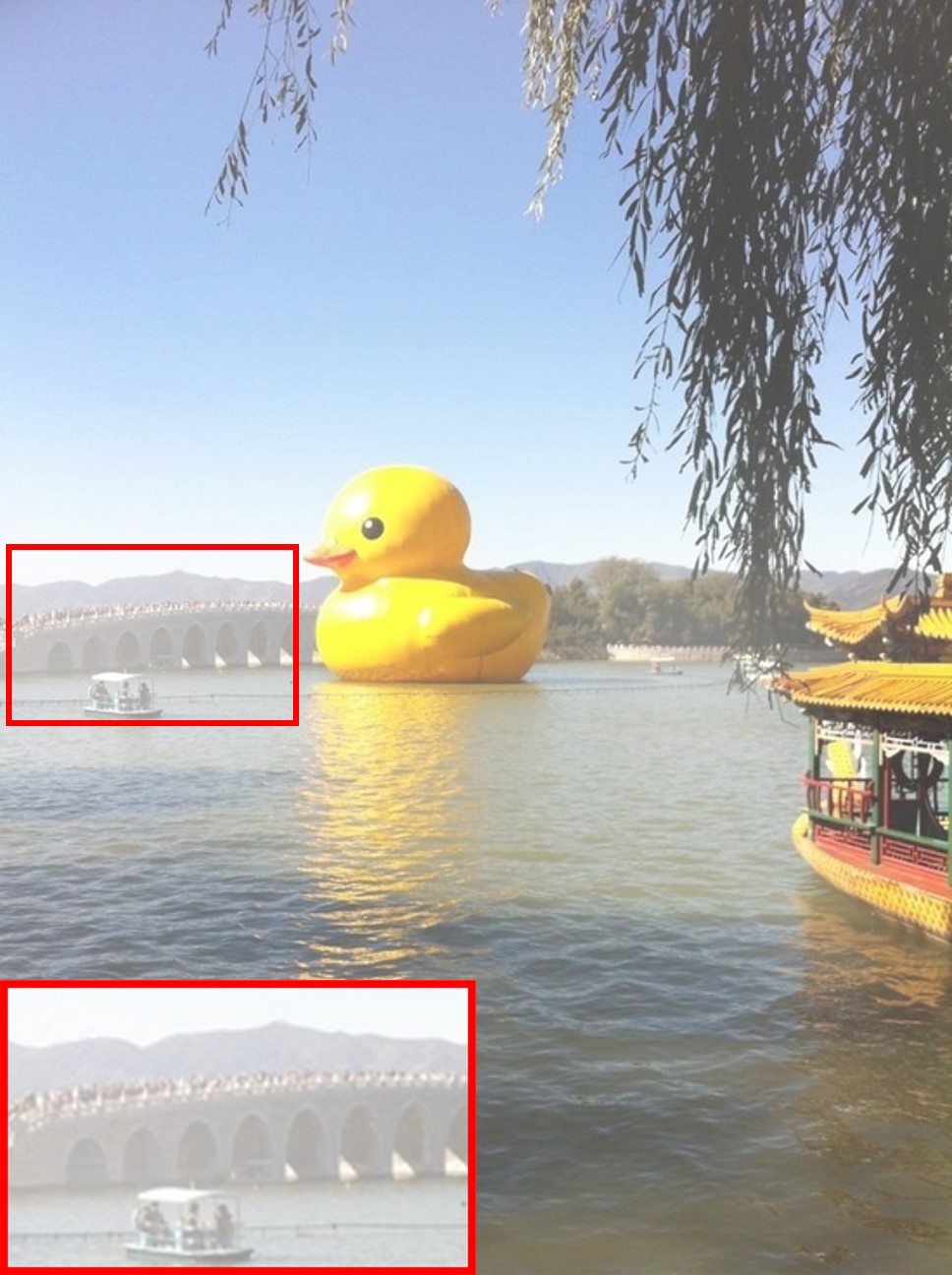} &
        \includegraphics[width=0.195\linewidth]{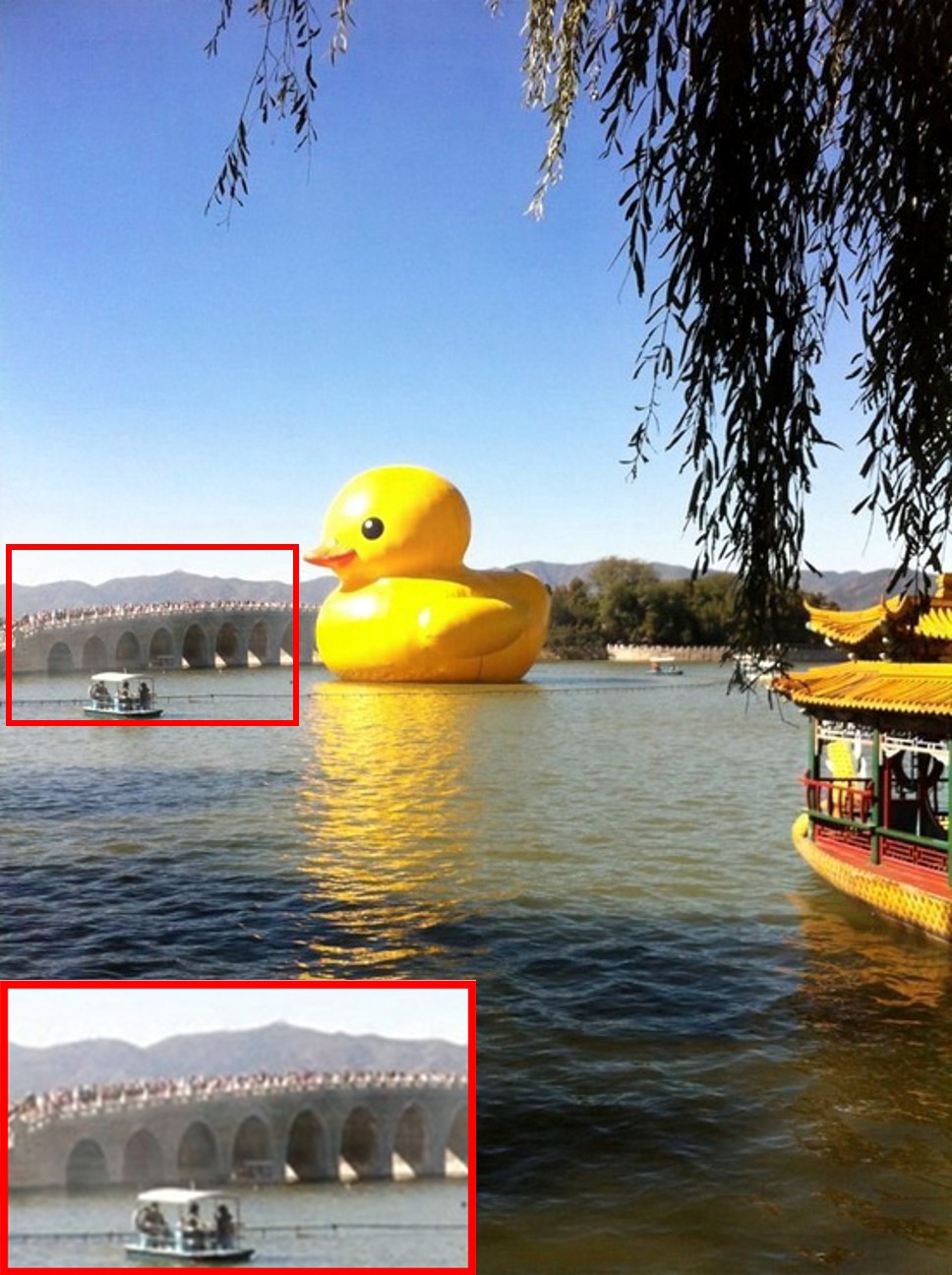} &
        \includegraphics[width=0.195\linewidth]{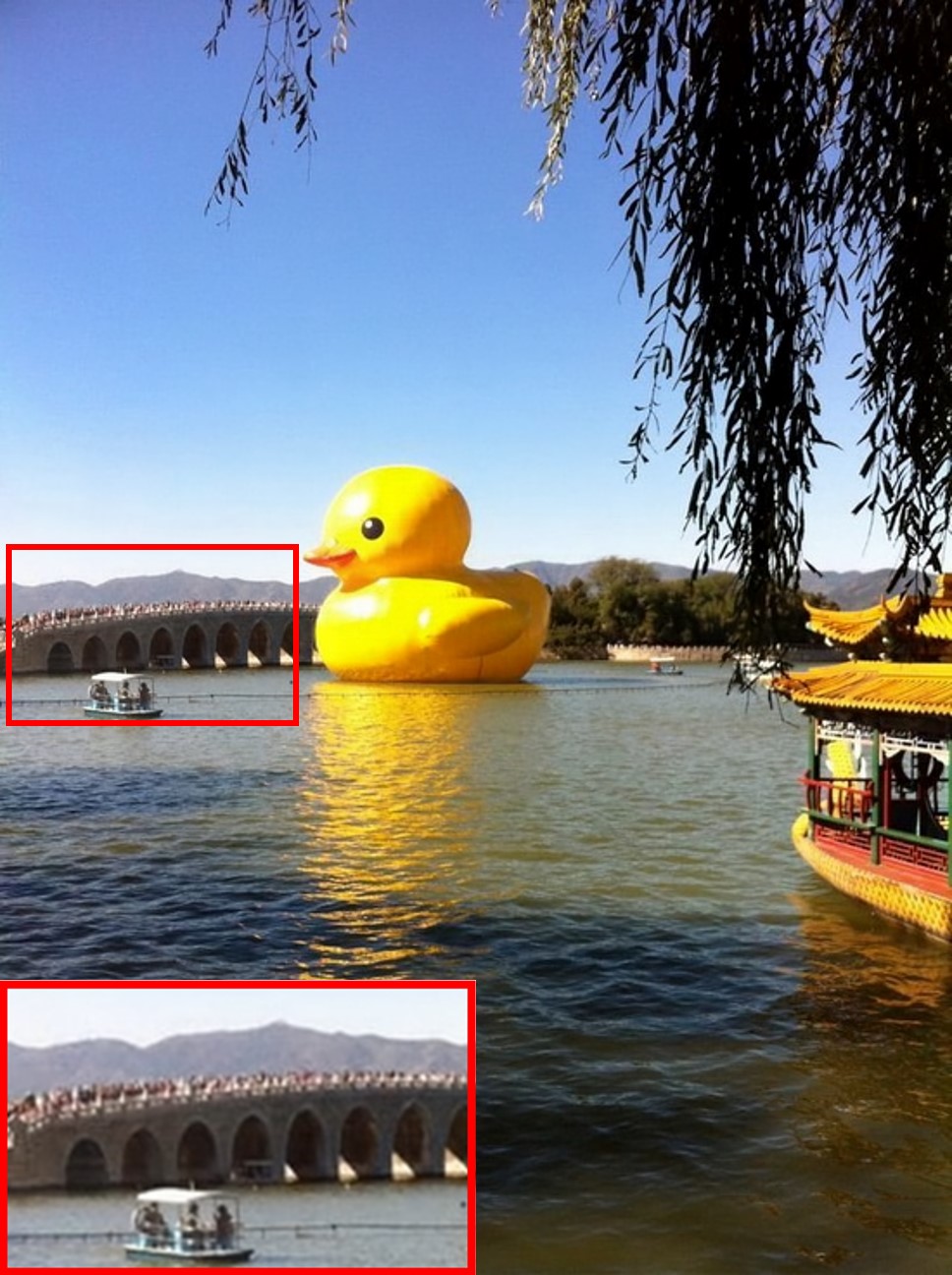} &
        \includegraphics[width=0.195\linewidth]{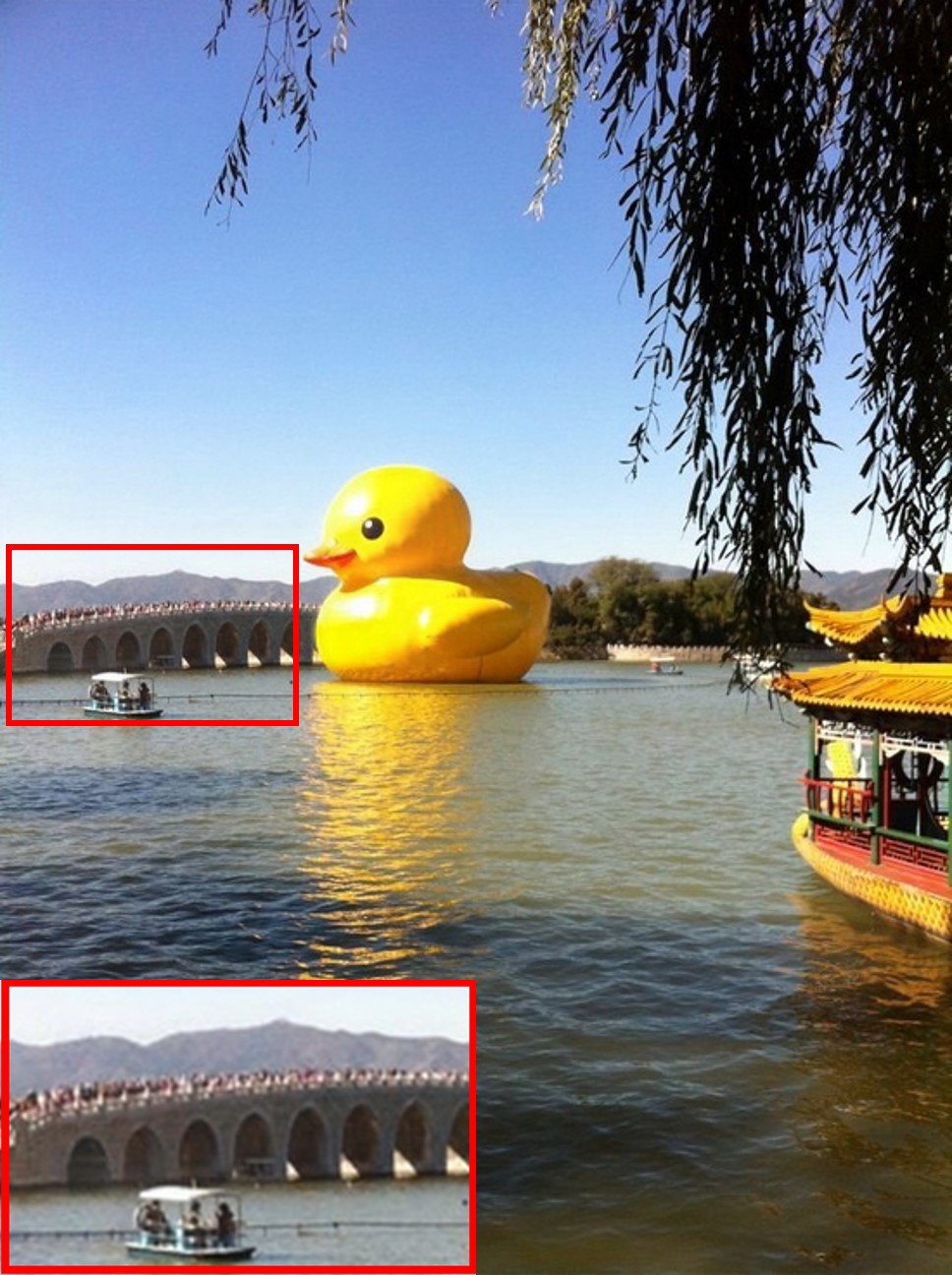} \\
        
        13.36 dB & 26.95 dB & 28.42 dB & 31.65 dB \\
        
        (a) Hazy image & (b) FFA-Net~\cite{qin2020ffa} & (c) MAXIM-2S~\cite{tu2022maxim} & (d) Dehamer~\cite{9879191}
    \end{tabular}
    \hspace{1em}
    \begin{tabular}{cccc}
        
        \includegraphics[width=0.195\linewidth]{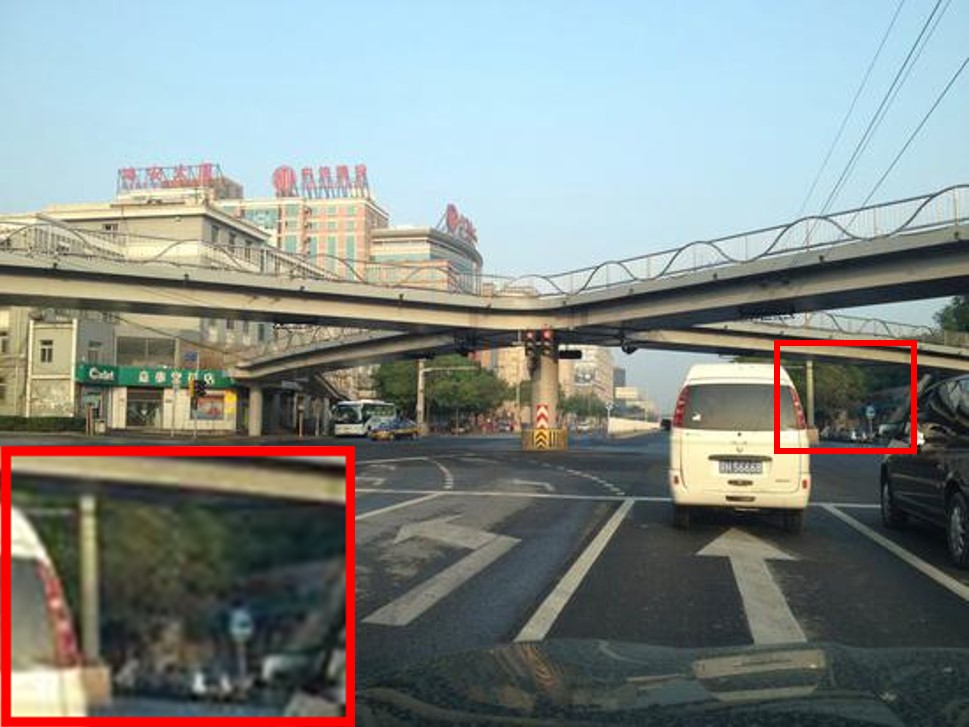} &
        \includegraphics[width=0.195\linewidth]{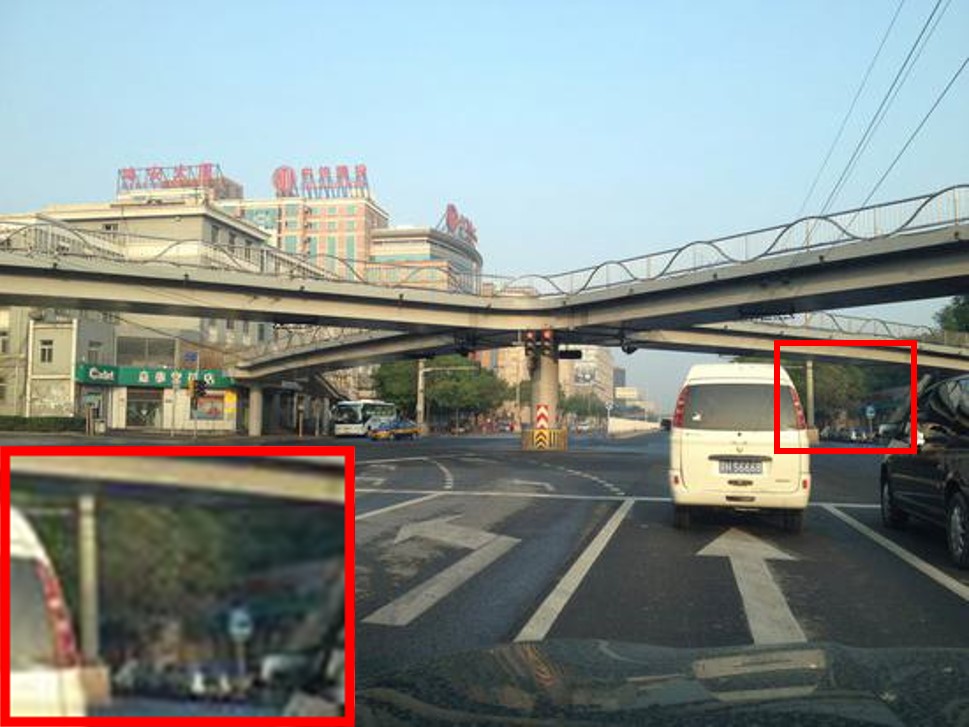} &
        \includegraphics[width=0.195\linewidth]{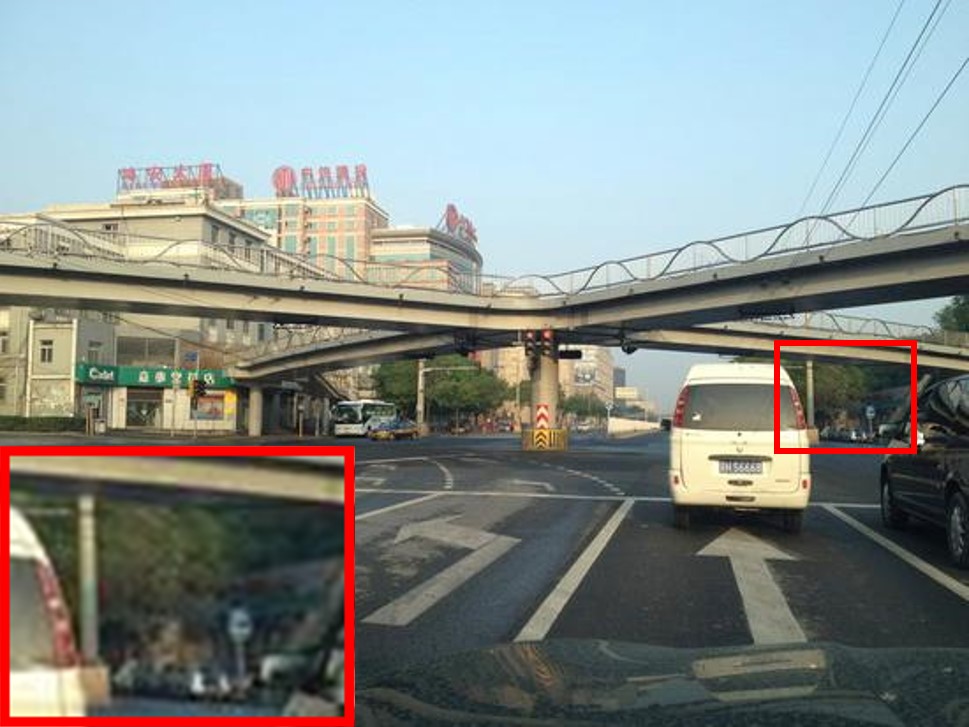} &
        \includegraphics[width=0.195\linewidth]{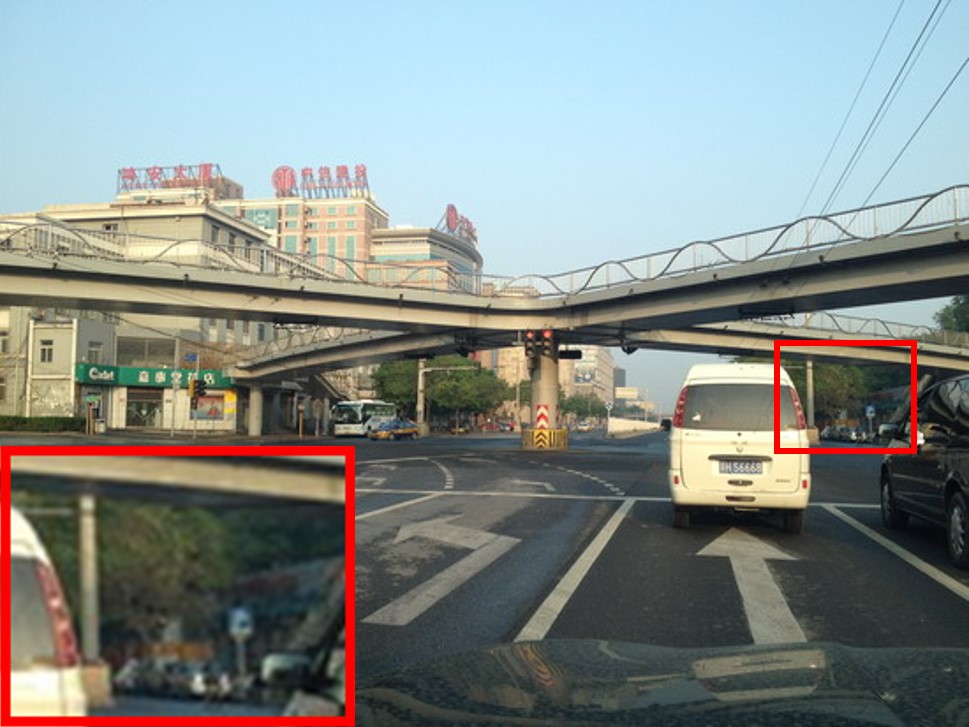} \\
        
        38.96 dB & 38.31 dB & 40.32 dB & PSNR \\
        
        \includegraphics[width=0.195\linewidth]{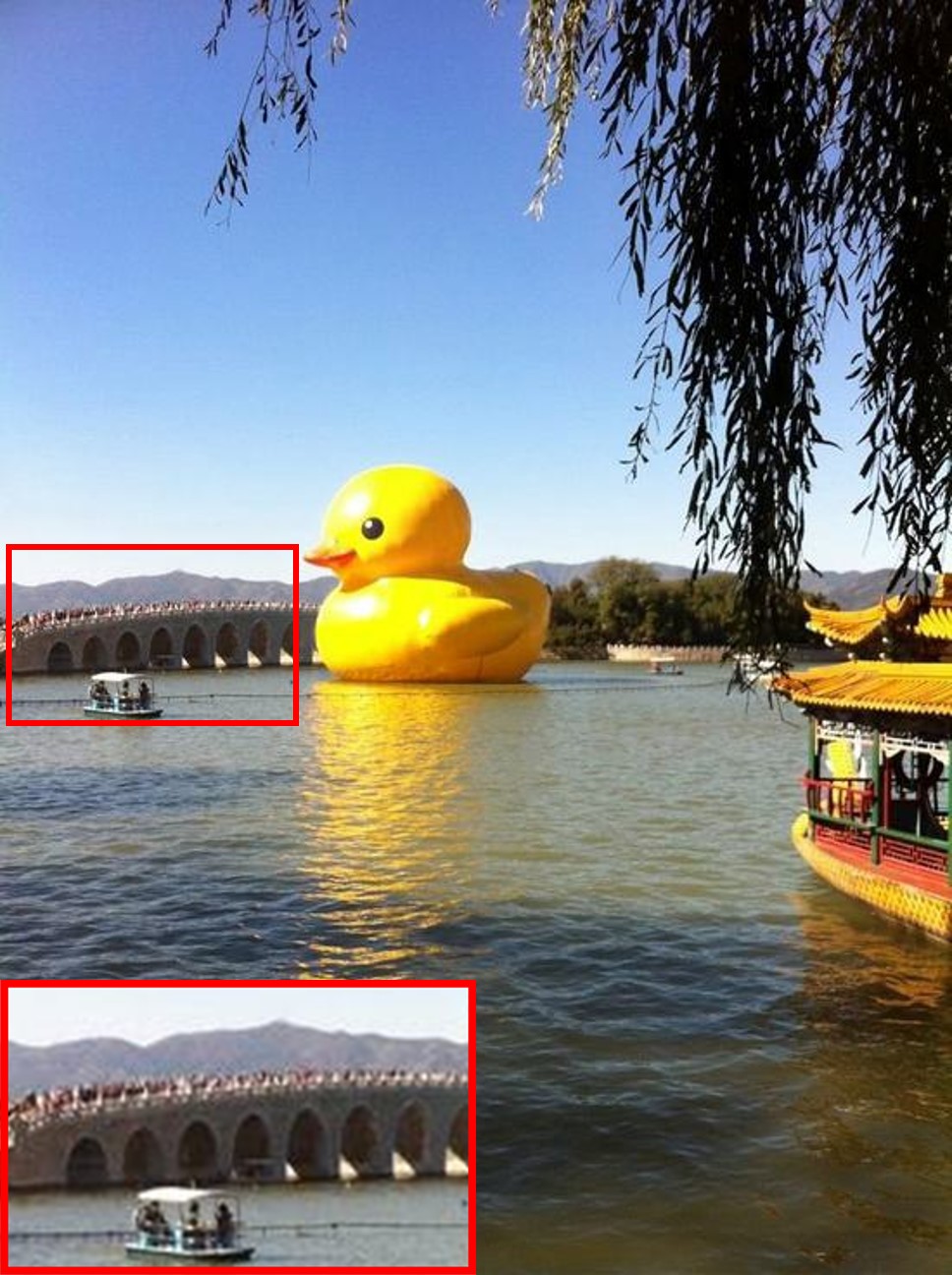} &
        \includegraphics[width=0.195\linewidth]{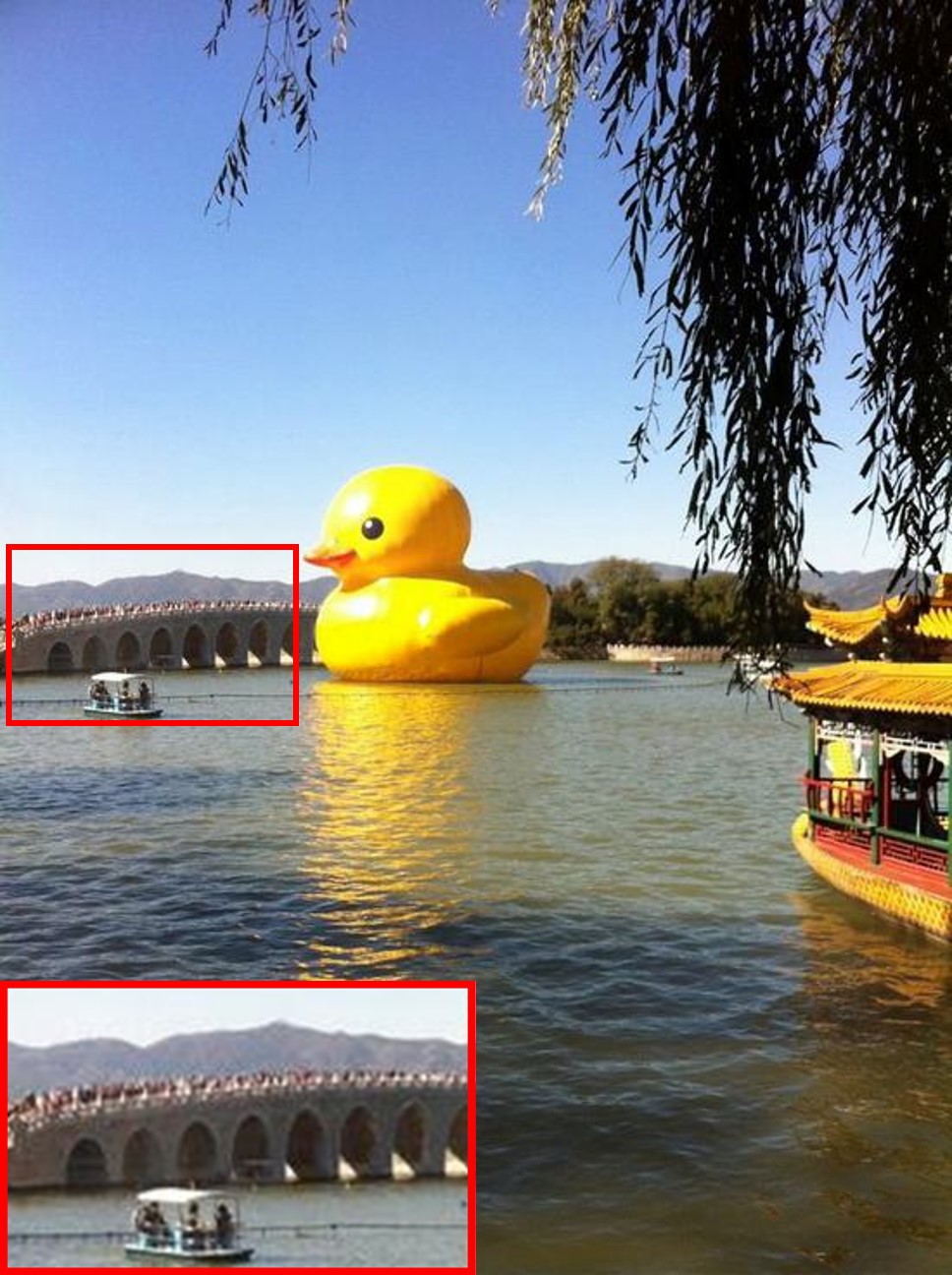} &
        \includegraphics[width=0.195\linewidth]{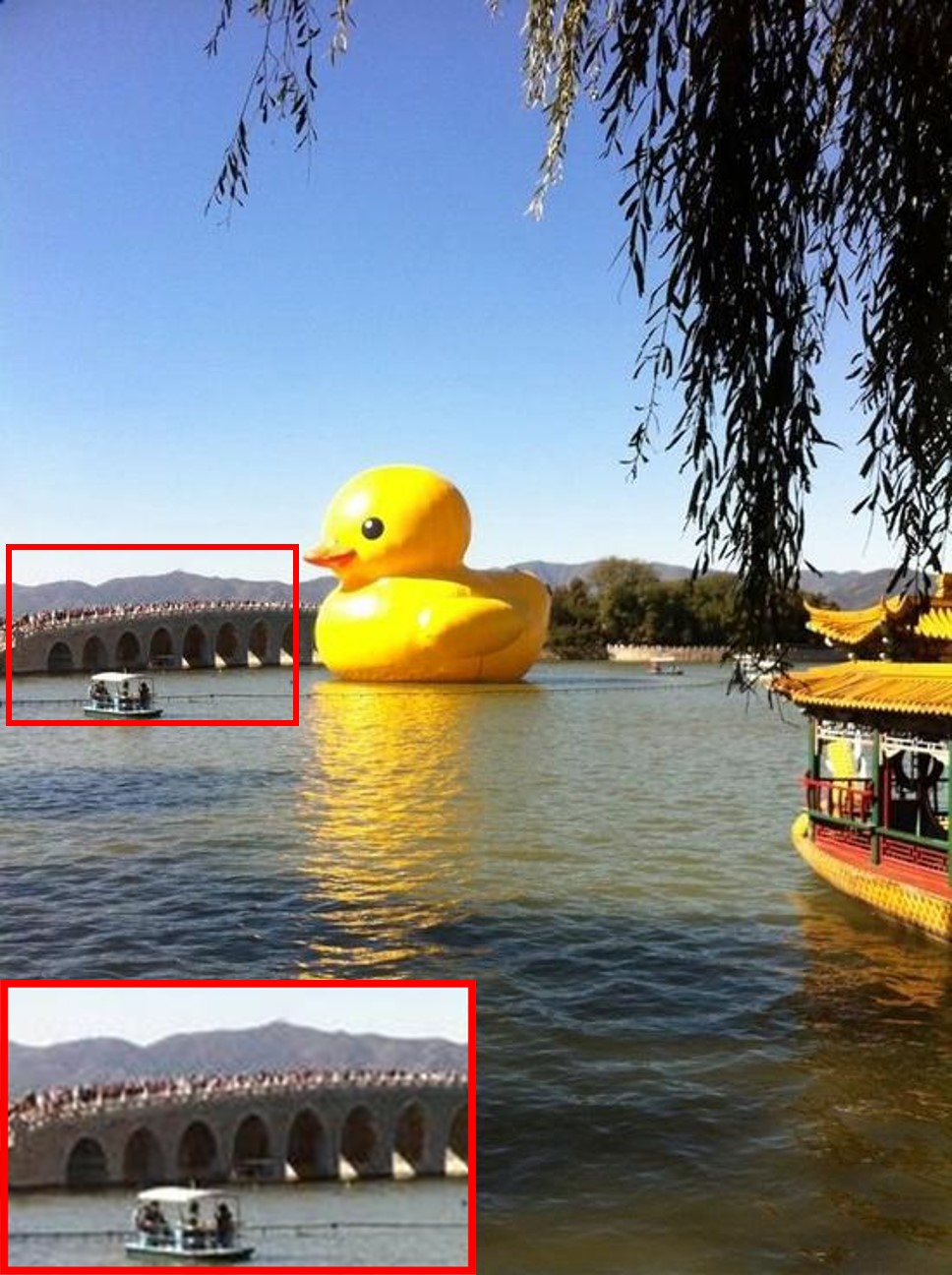} &
        \includegraphics[width=0.195\linewidth]{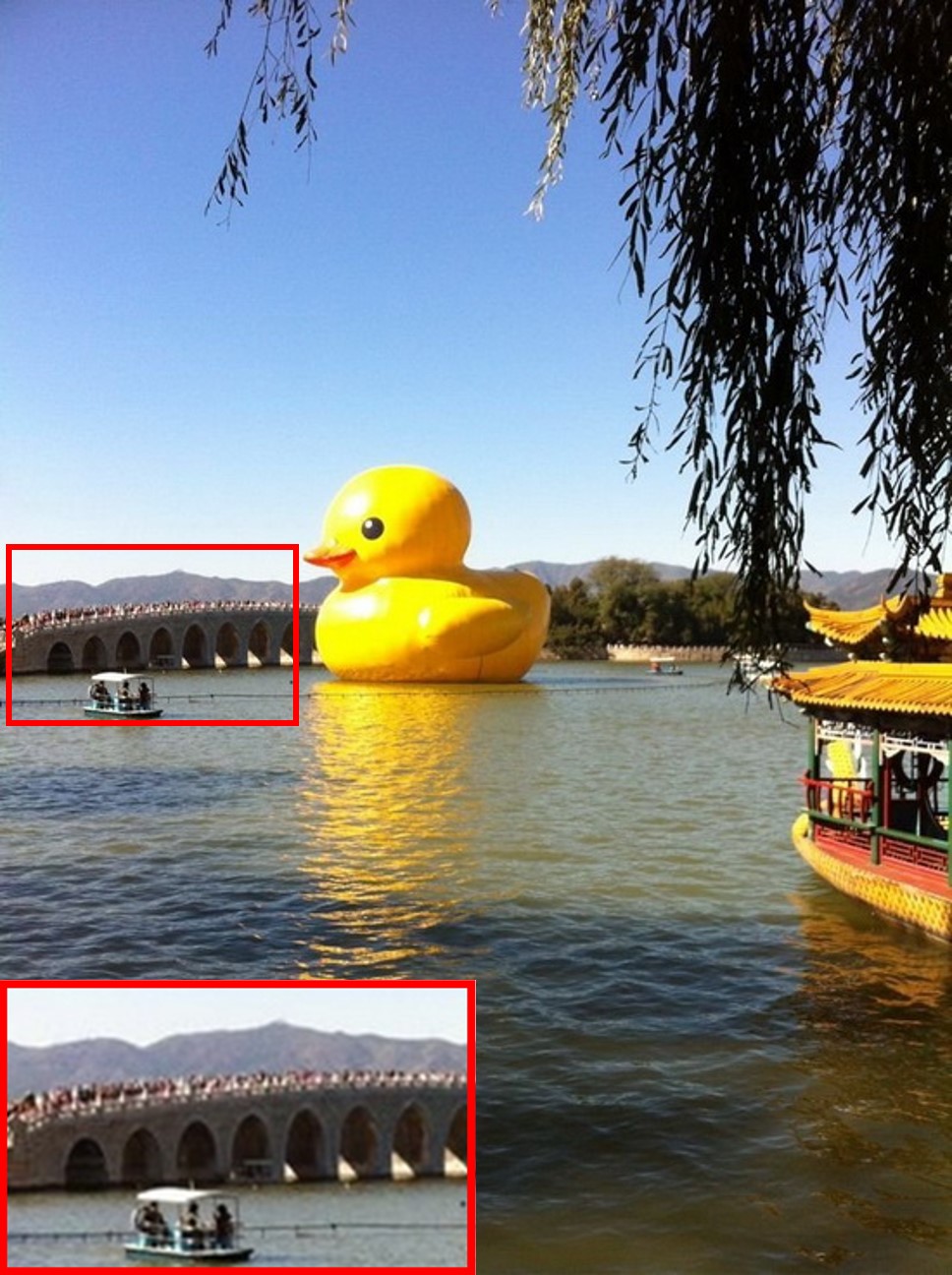} \\
        
        34.99 dB & 33.41 dB & 38.21 dB & PSNR \\
        
        (e) FocalNet~\cite{10377428} & (f) OKNet~\cite{cui2024omni} & (g) Ours & (h) GT
    \end{tabular}
    
    \caption{Visual comparisons on synthetic hazy images from the SOTS-Outdoor dataset. Key regions highlighted by red boxes are enlarged in the lower-left corner for clearer comparison.}
    \label{fig: Outdoor}
\end{figure}

For the synthetic datasets, Fourier-RWKV achieves the second-best overall performance on the SOTS-Indoor subset. Its PSNR is 0.1 dB lower than PGH2Net, which is a similarly lightweight model. This minor gap primarily stems from PGH2Net’s integration of three distinct haze priors, giving it an advantage in relatively uniform synthetic indoor scenes. In comparison with another linear attention model MAIR~\cite{Li_2025_CVPR}, Fourier-RWKV adds only 1.91 M additional parameters, while reducing FLOPs to 65.29\% of MAIR, highlighting its exceptional computational efficiency. Despite a slight SSIM decrease of 0.001, Fourier-RWKV improves PSNR by 2.15 dB, underscoring its ability to recover fine image details and preserve structural integrity. On the SOTS-Outdoor subset, Fourier-RWKV outperforms all other models with a 2.05 dB PSNR gain and a 0.001 SSIM increase, reinforcing its robustness in handling more challenging outdoor haze scenarios.

For the real-world datasets, Fourier-RWKV achieves the best performance across all evaluation metrics on both Dense-Haze~\cite{ancuti2019dense} and NH-HAZE~\cite{9150807}. Notably, on the NH-HAZE dataset, which focuses on non-uniform haze, our method surpasses the previous best results, improving PSNR by $0.35$ dB and SSIM by $0.03$, showcasing its strong generalization capability in complex real-world haze scenarios.

\begin{figure}[t]
	\scriptsize
	\centering
	\renewcommand{\tabcolsep}{1pt} 
	\renewcommand{\arraystretch}{1}
	\begin{center}
    \begin{tabular}{ccccc}
      \includegraphics[width=0.190\linewidth]{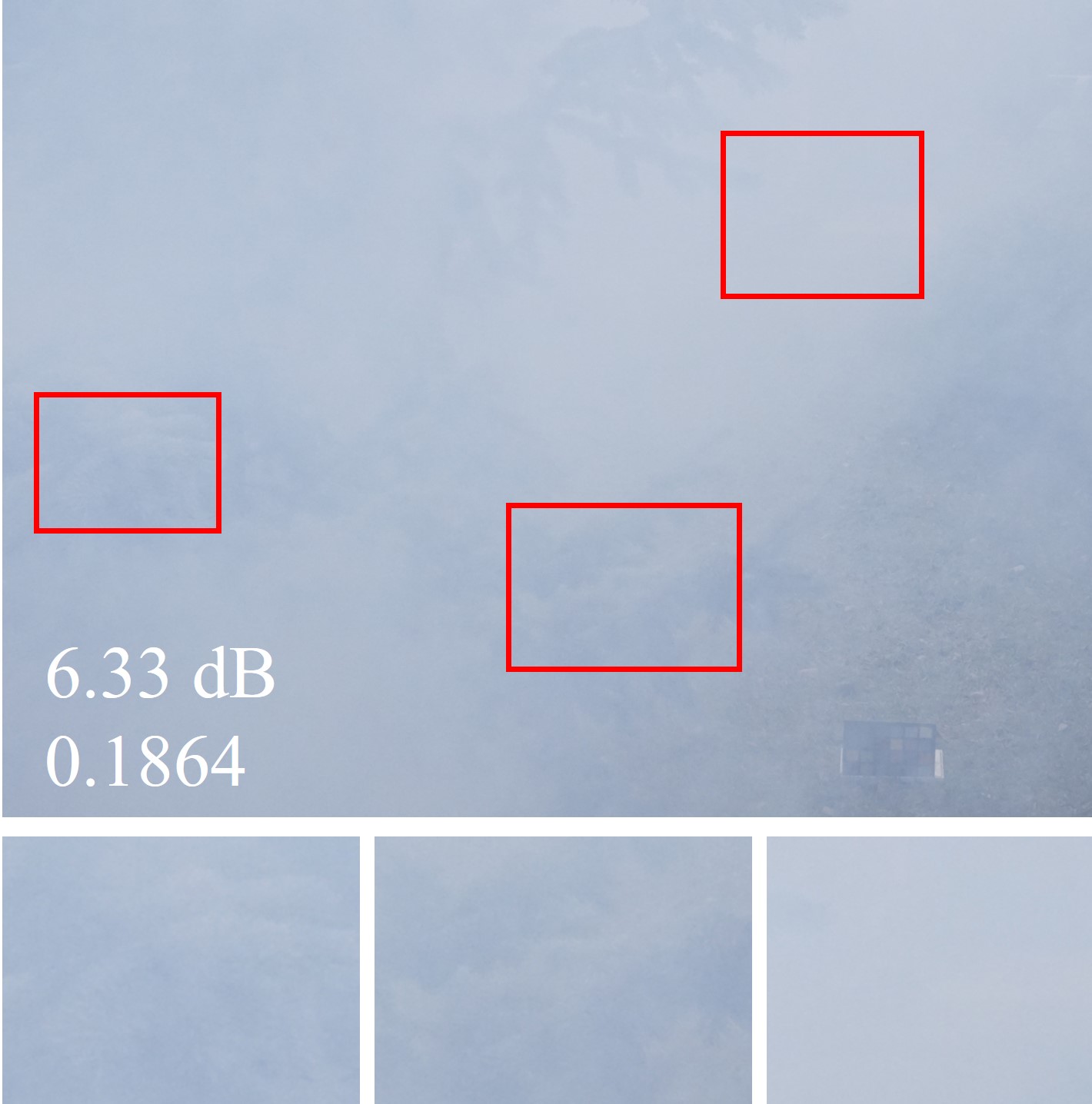} &
      \includegraphics[width=0.190\linewidth]{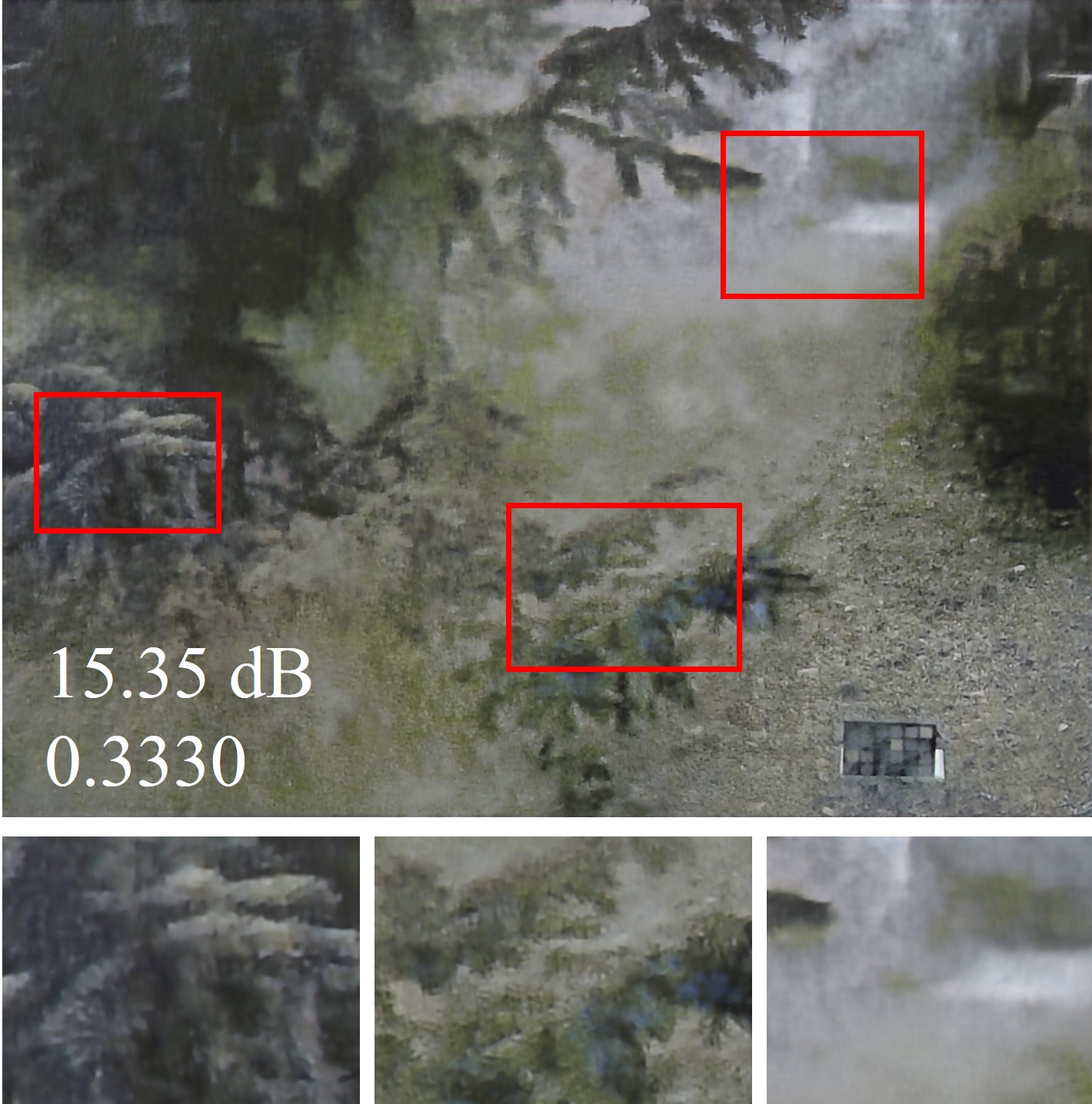} &
      \includegraphics[width=0.190\linewidth]{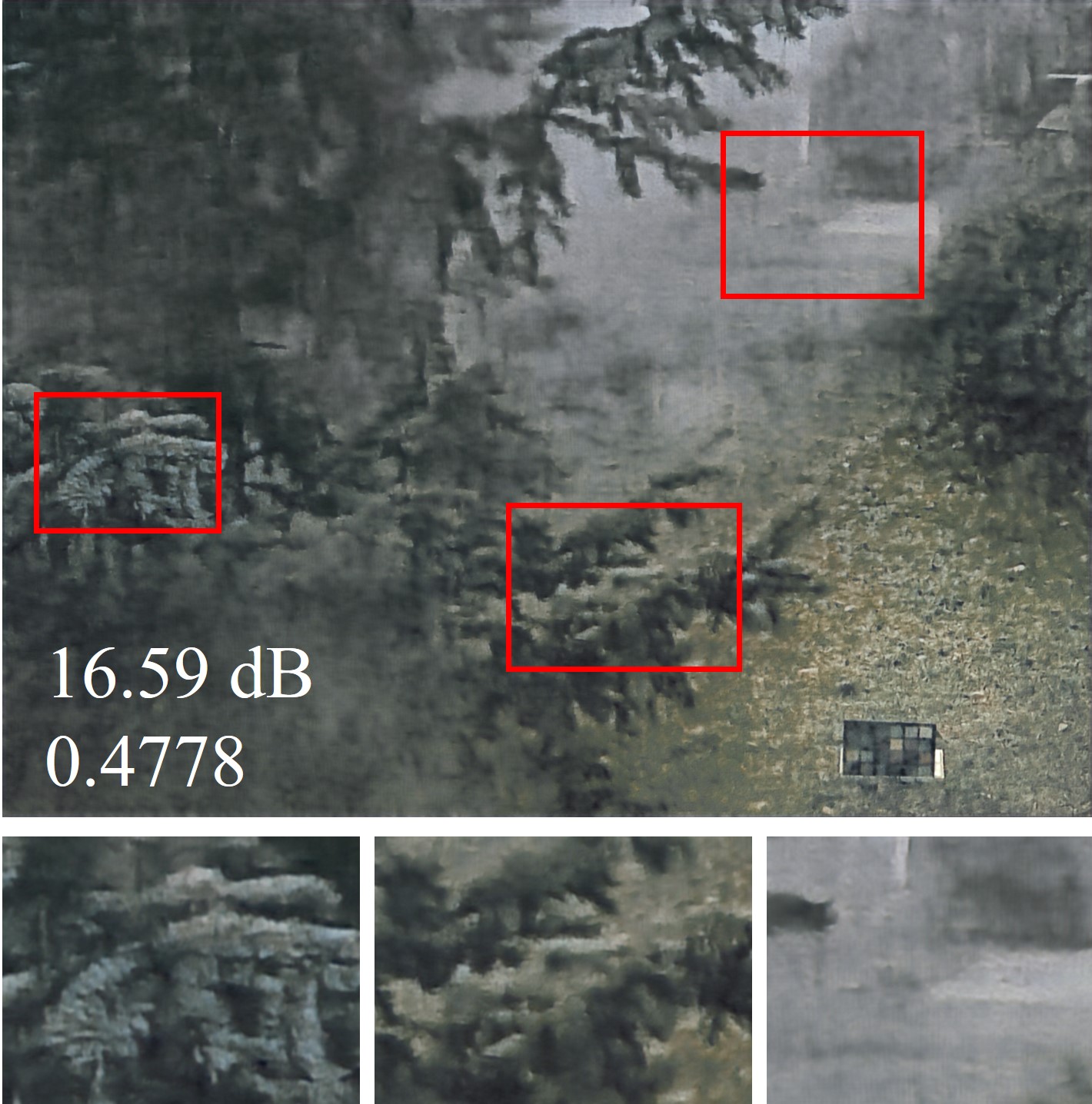} &
      \includegraphics[width=0.190\linewidth]{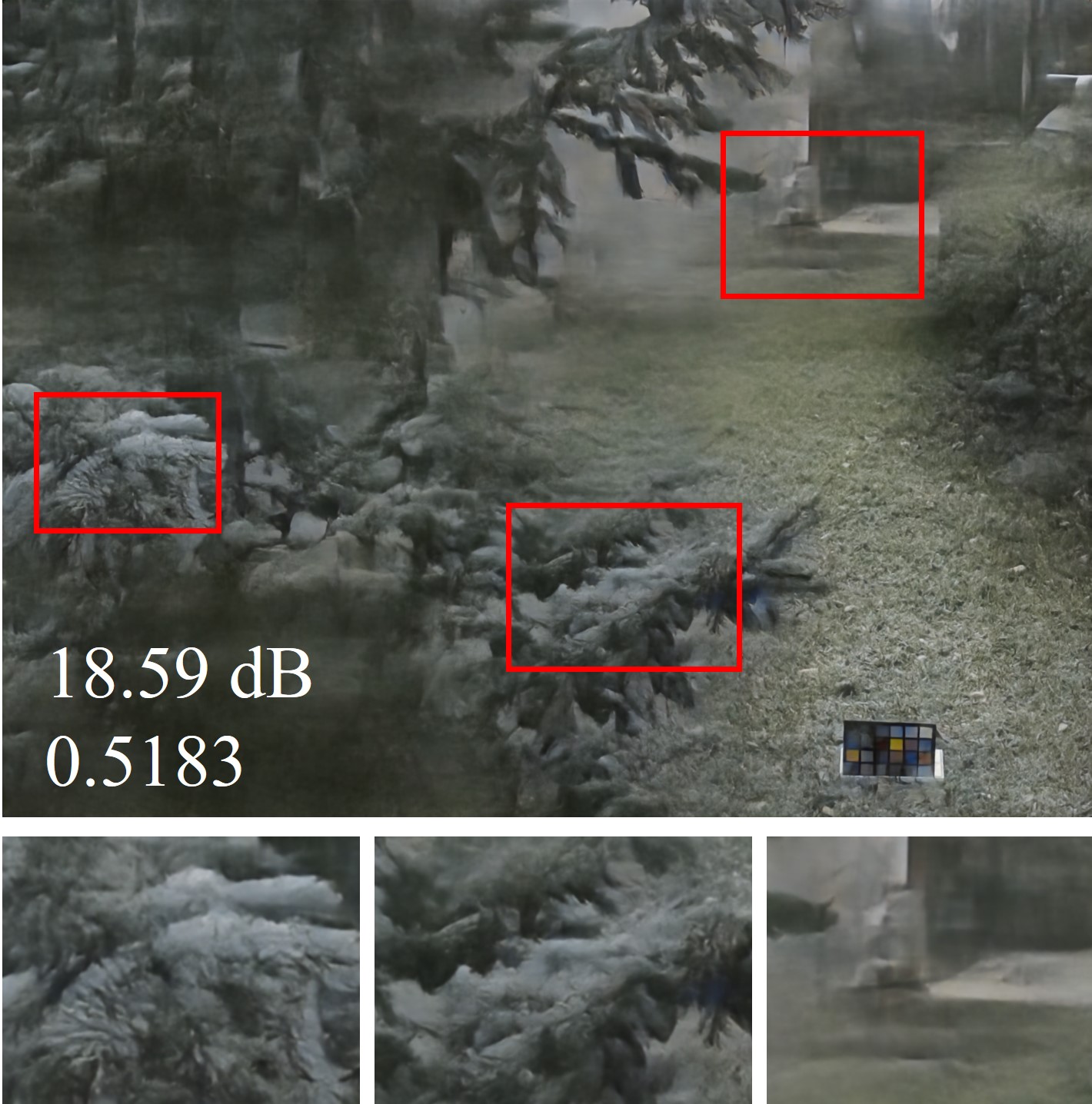} &
      \includegraphics[width=0.190\linewidth]{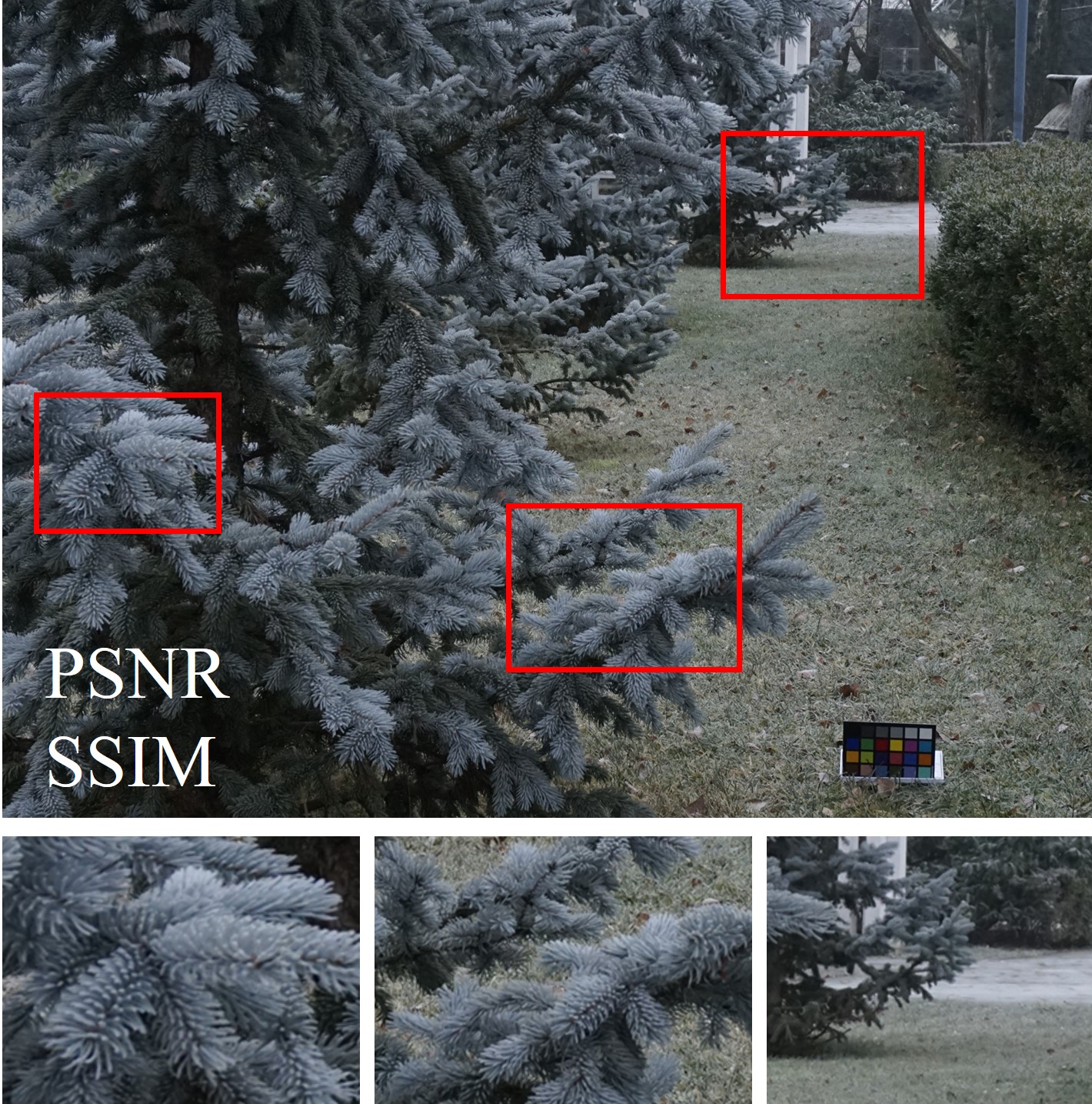} \\

      \includegraphics[width=0.190\linewidth]{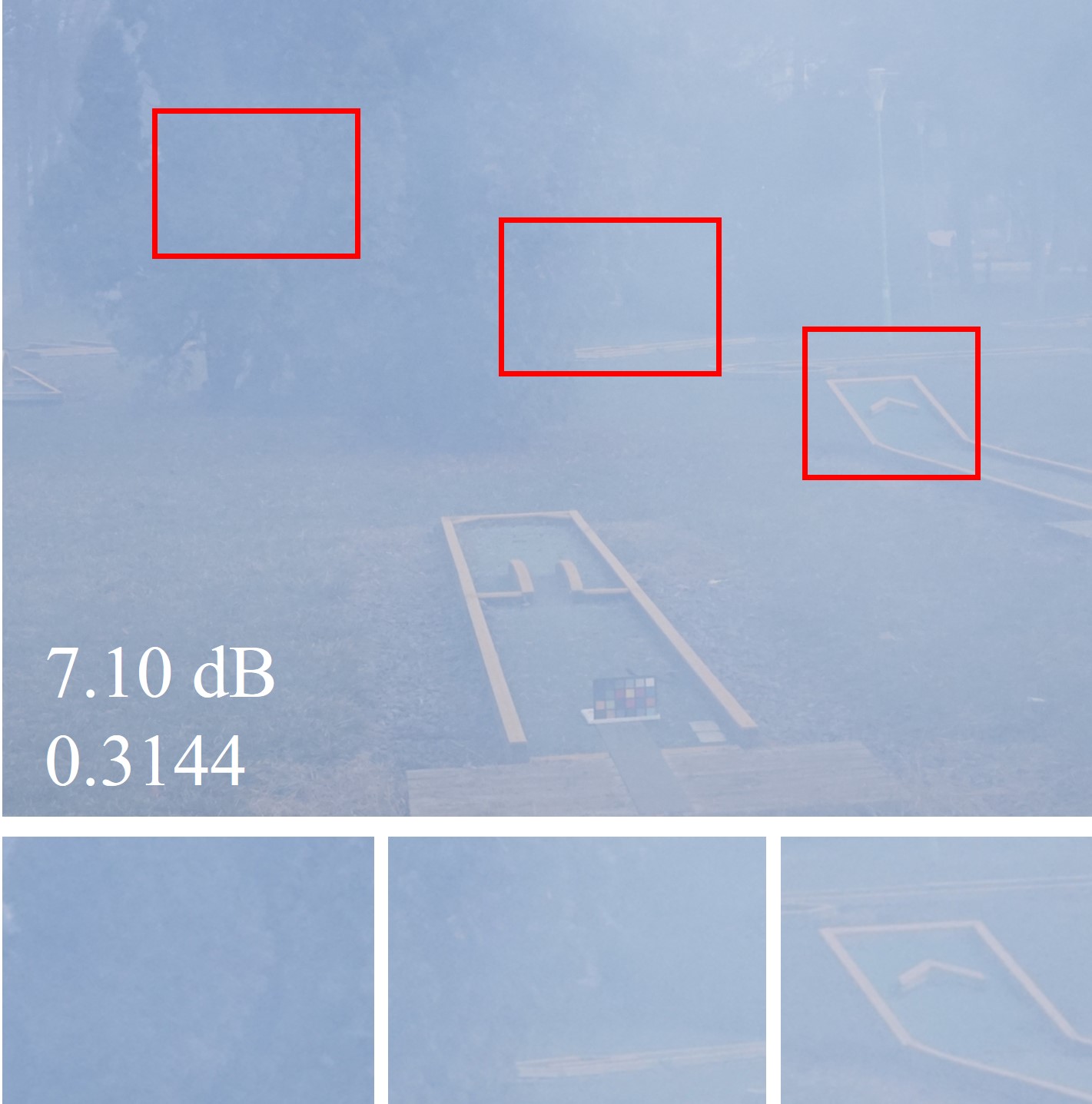} &
      \includegraphics[width=0.190\linewidth]{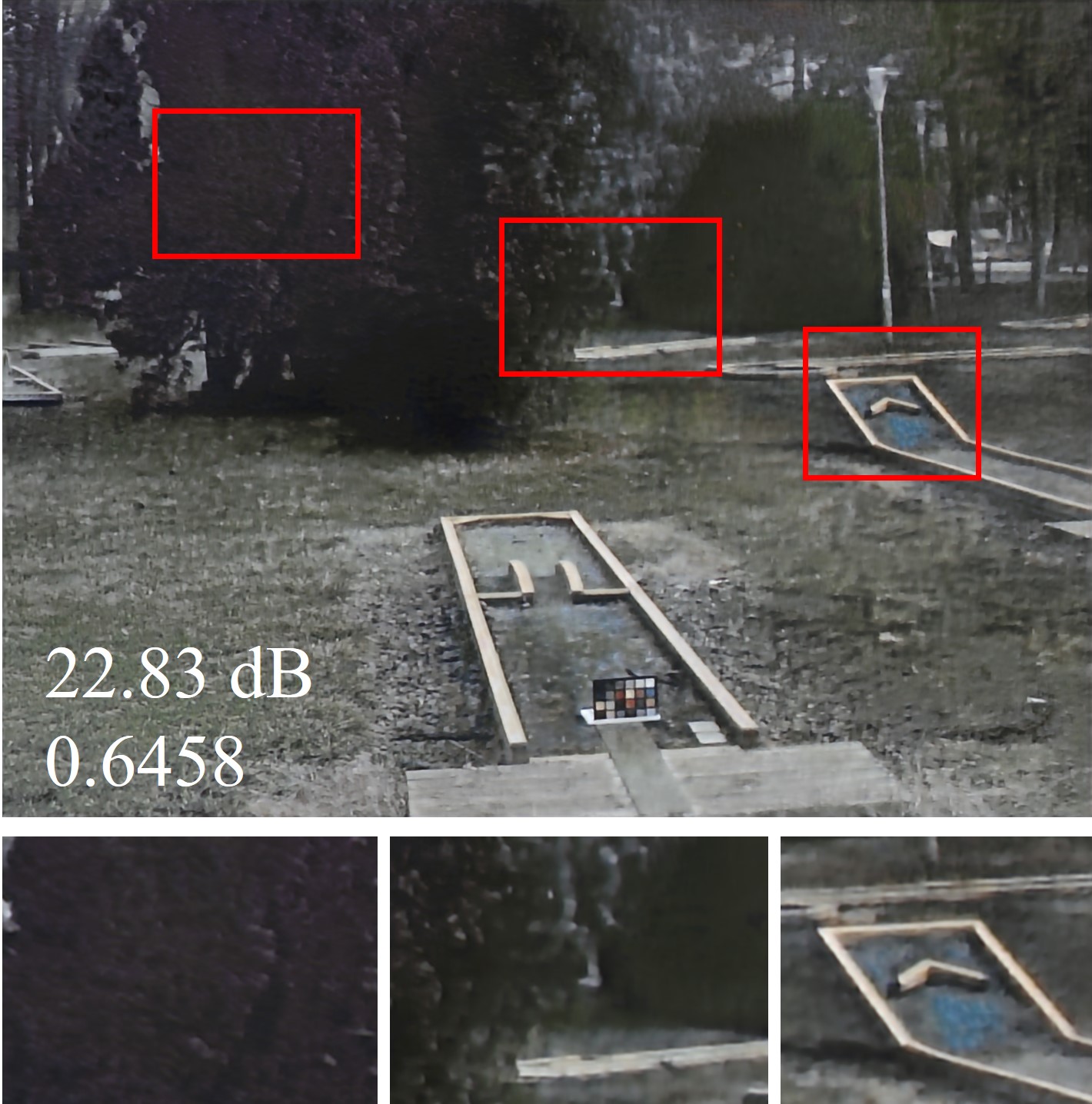} &
      \includegraphics[width=0.190\linewidth]{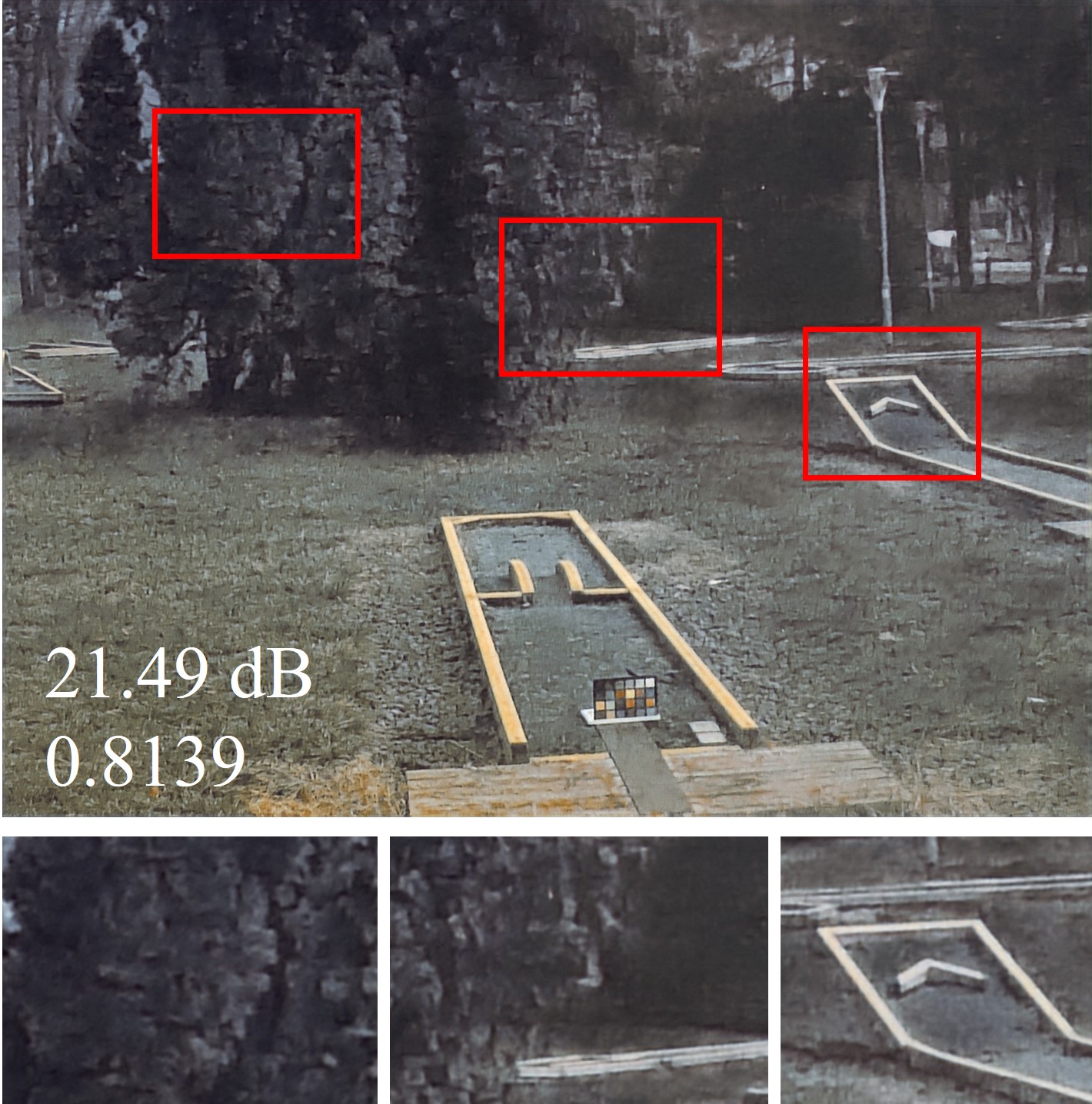} &
      \includegraphics[width=0.190\linewidth]{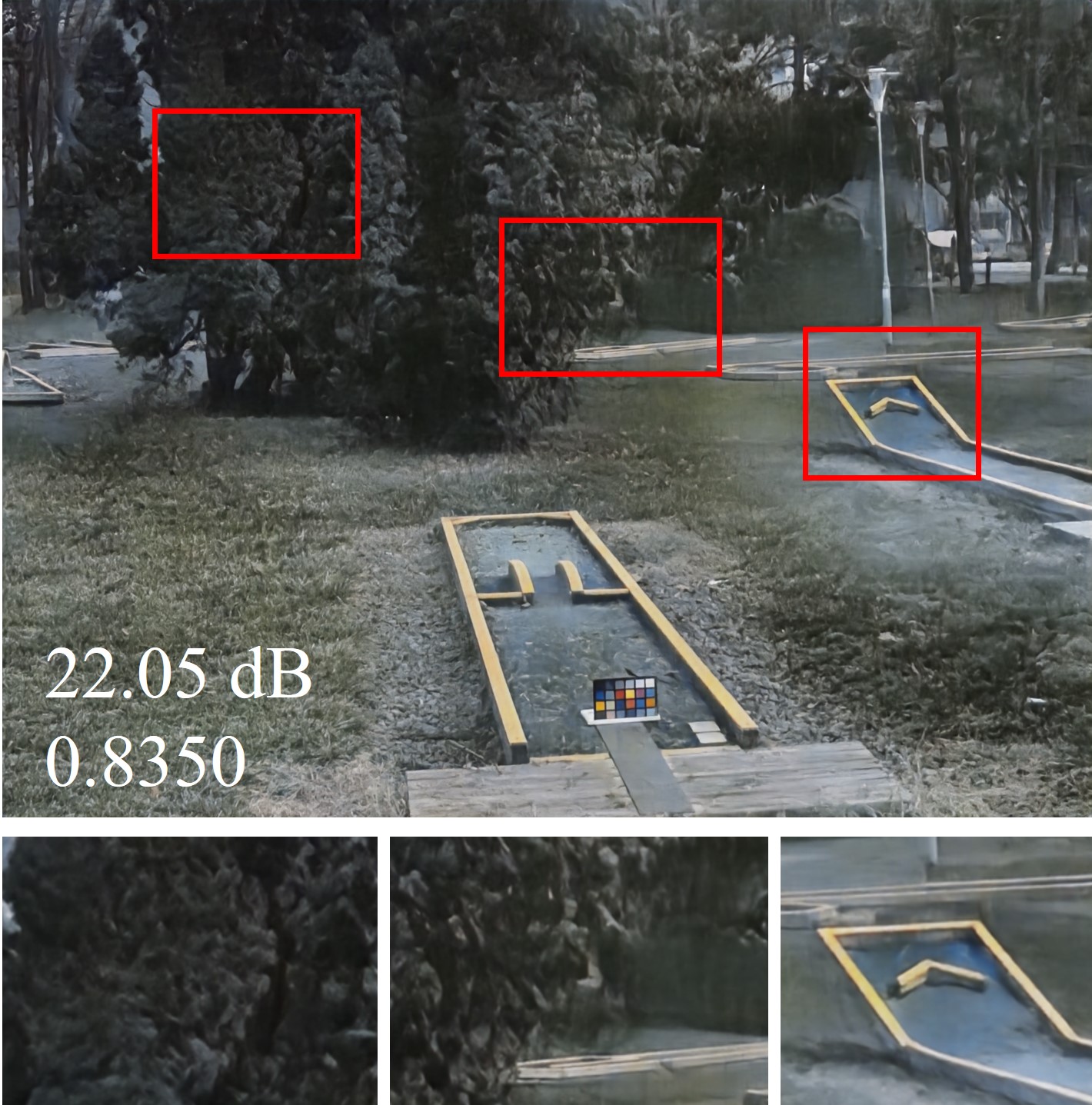} &
      \includegraphics[width=0.190\linewidth]{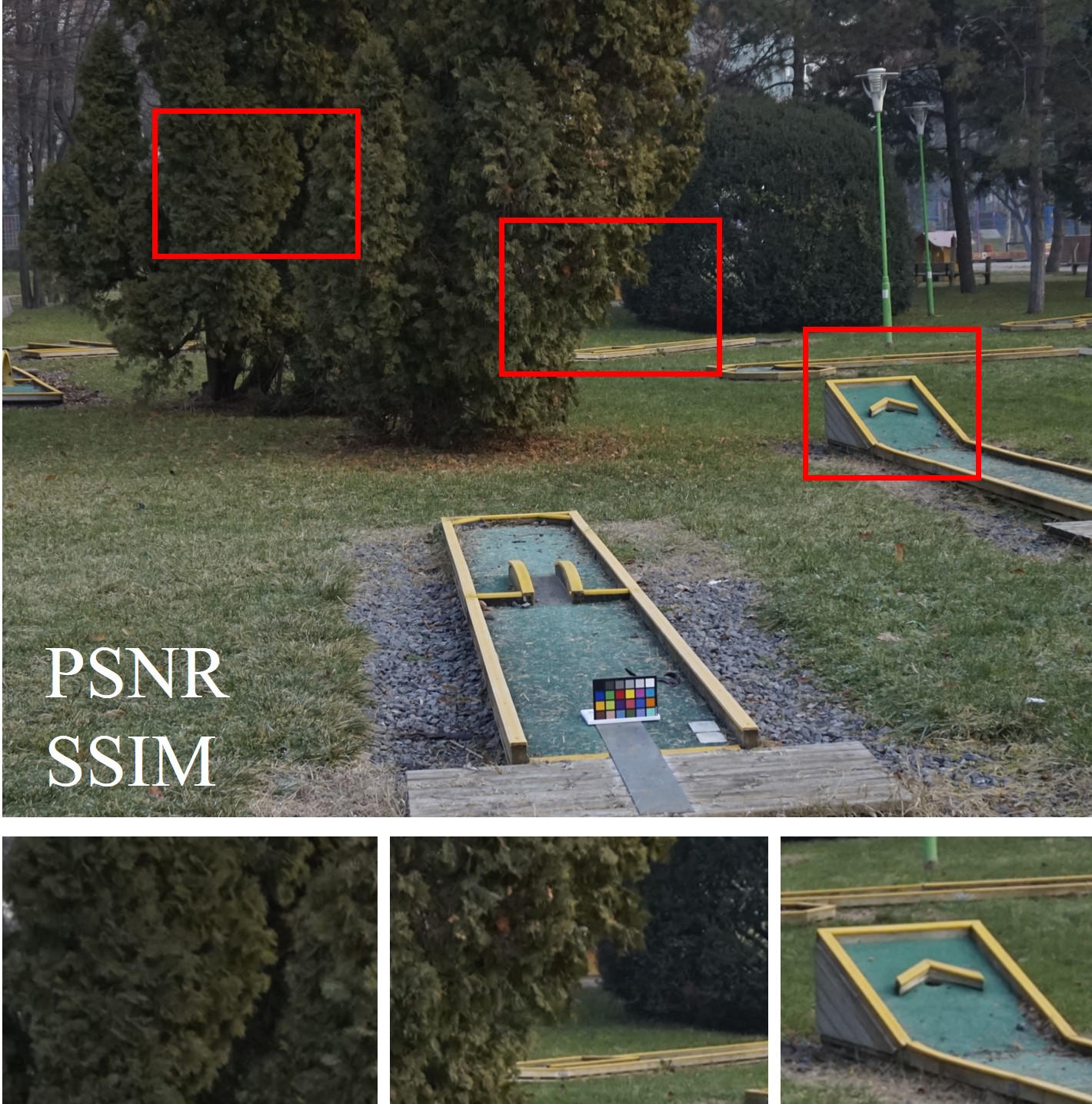} \\

	(a) Hazy image & 
        (b) Dehamer~\cite{9879191} & 
        (c) MB~\cite{10378631} & 
        (d) Ours & 
        (e) GT 
		\end{tabular}
	\end{center}
	\caption{Visual comparisons on real-world hazy images from the Dense dataset. Key regions marked with red boxes are enlarged in left-to-right order and arranged horizontally at the bottom.}
	\label{fig: dense}
\end{figure}

\begin{figure}[t]
	\scriptsize
	\centering
	\renewcommand{\tabcolsep}{1pt} 
	\renewcommand{\arraystretch}{1}
	\begin{center}
    \begin{tabular}{ccccc}
      \includegraphics[width=0.190\linewidth]{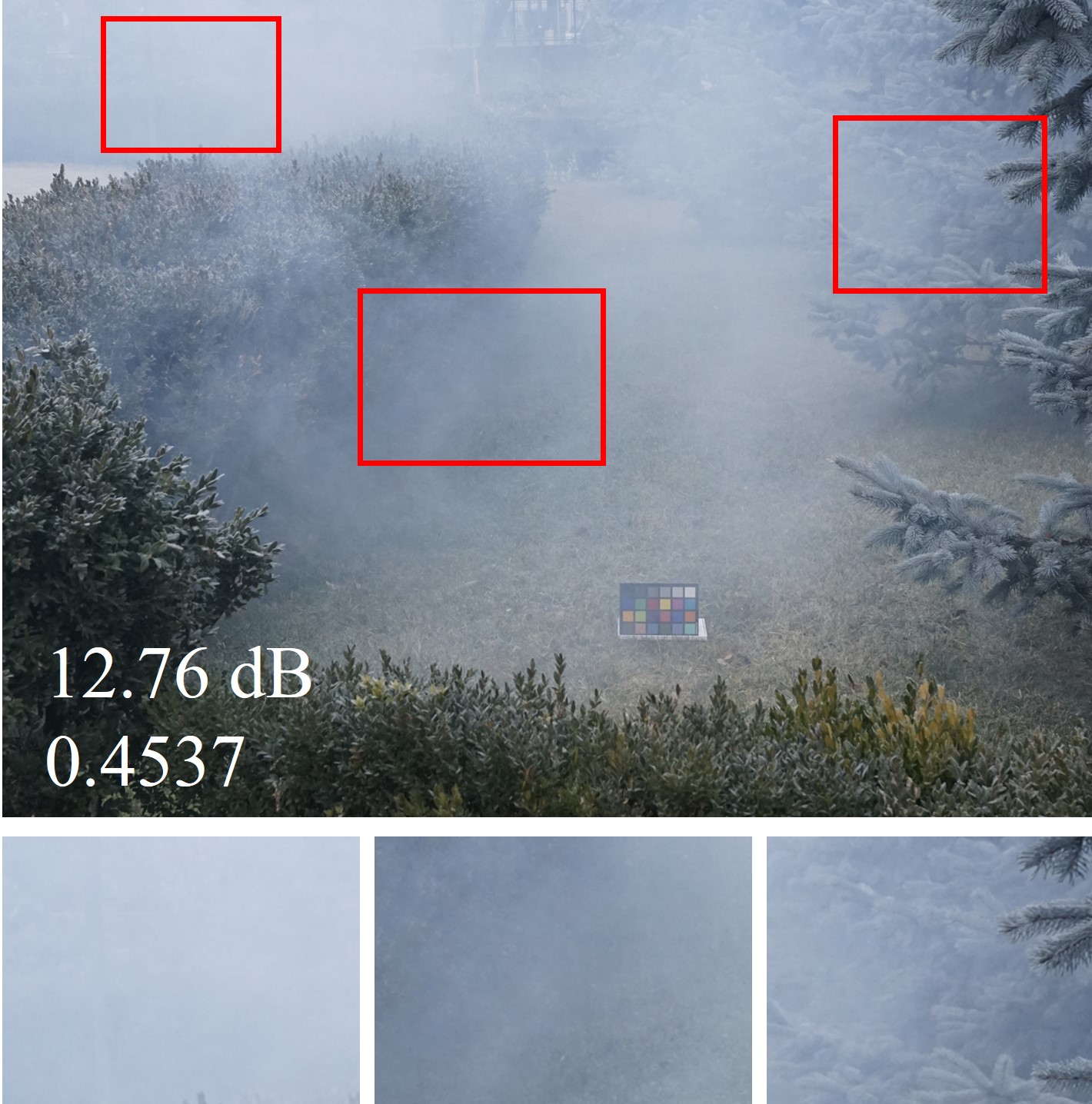} &
      \includegraphics[width=0.190\linewidth]{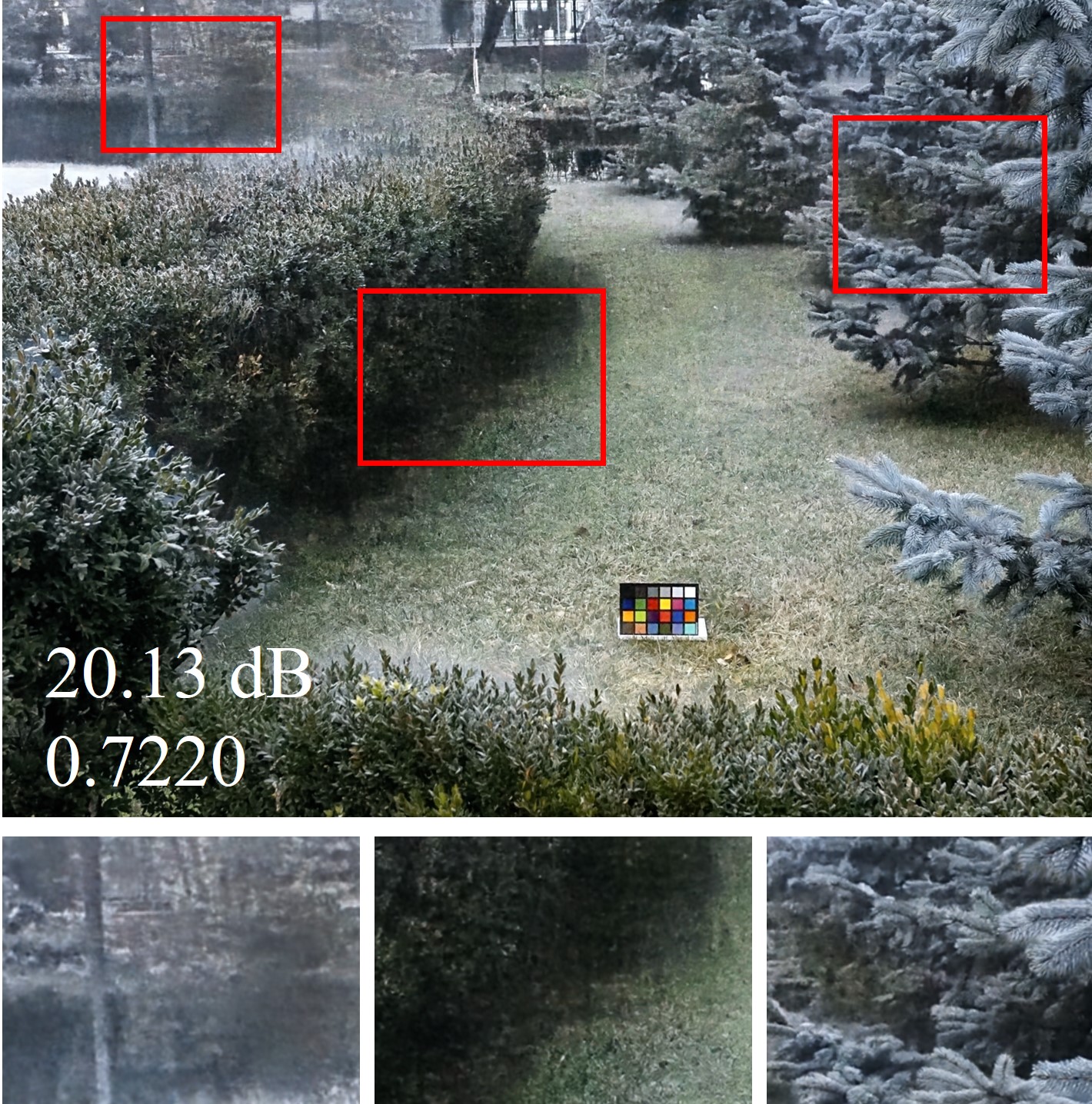} &
      \includegraphics[width=0.190\linewidth]{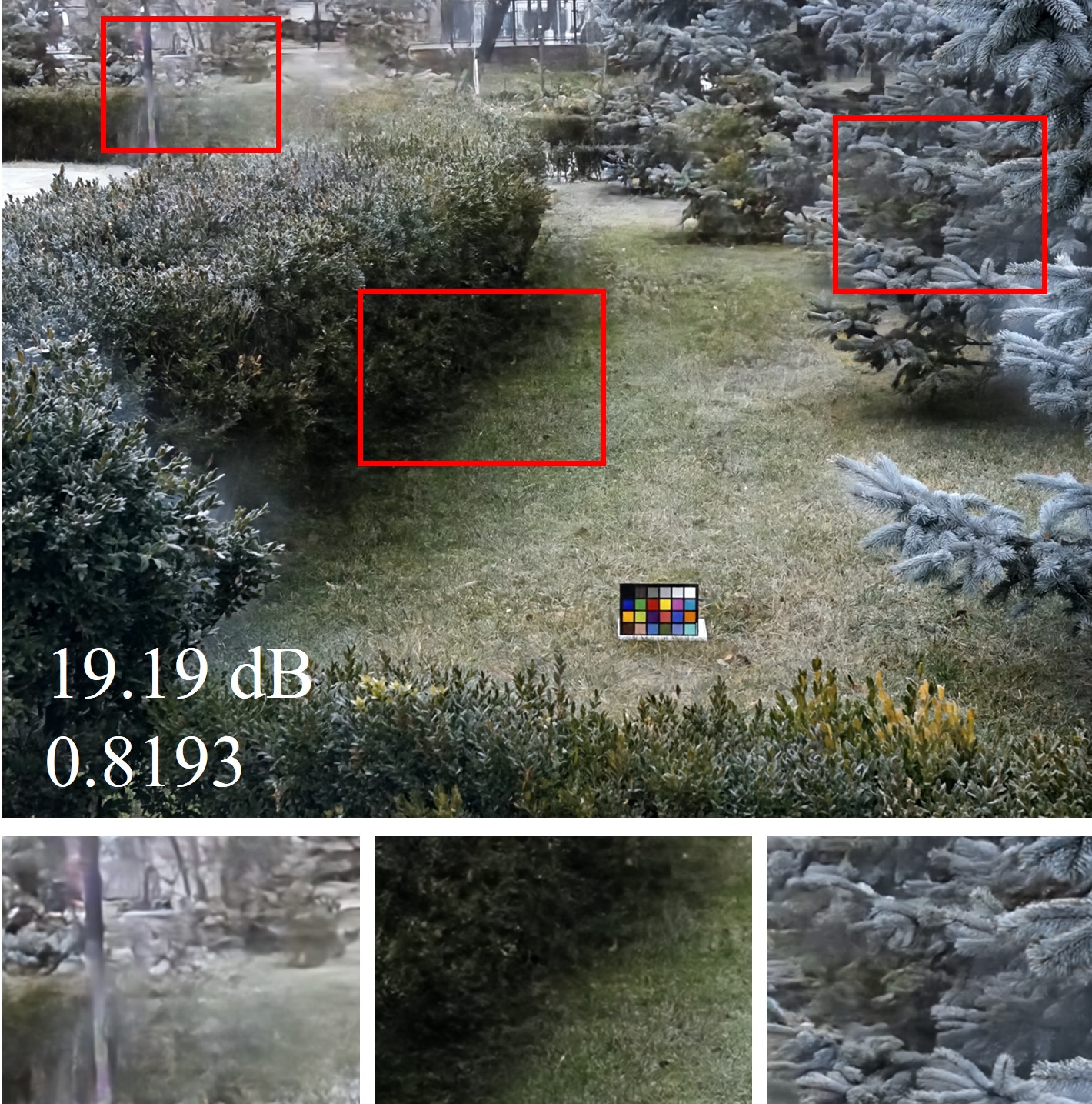} &
      \includegraphics[width=0.190\linewidth]{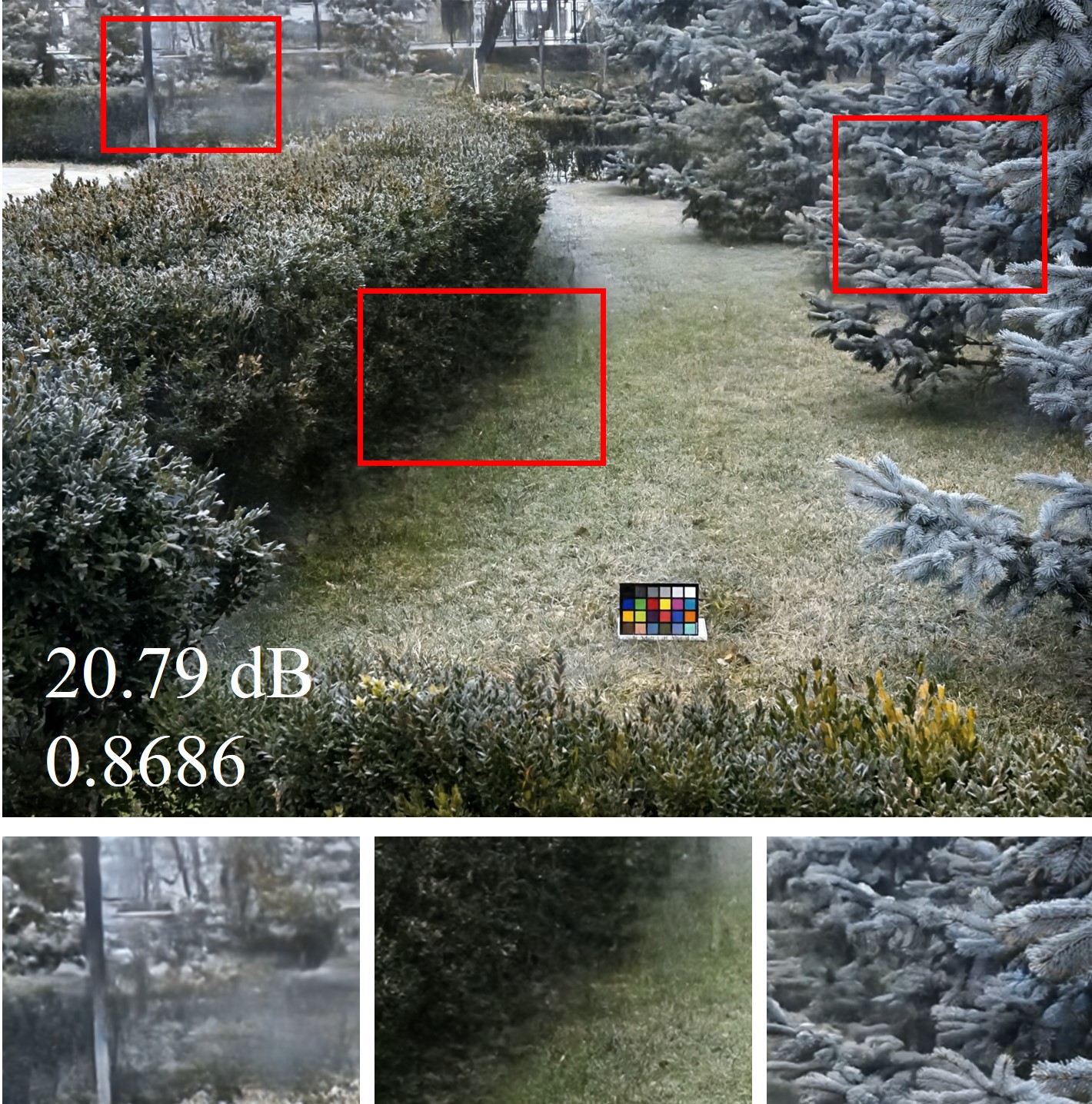} &
      \includegraphics[width=0.190\linewidth]{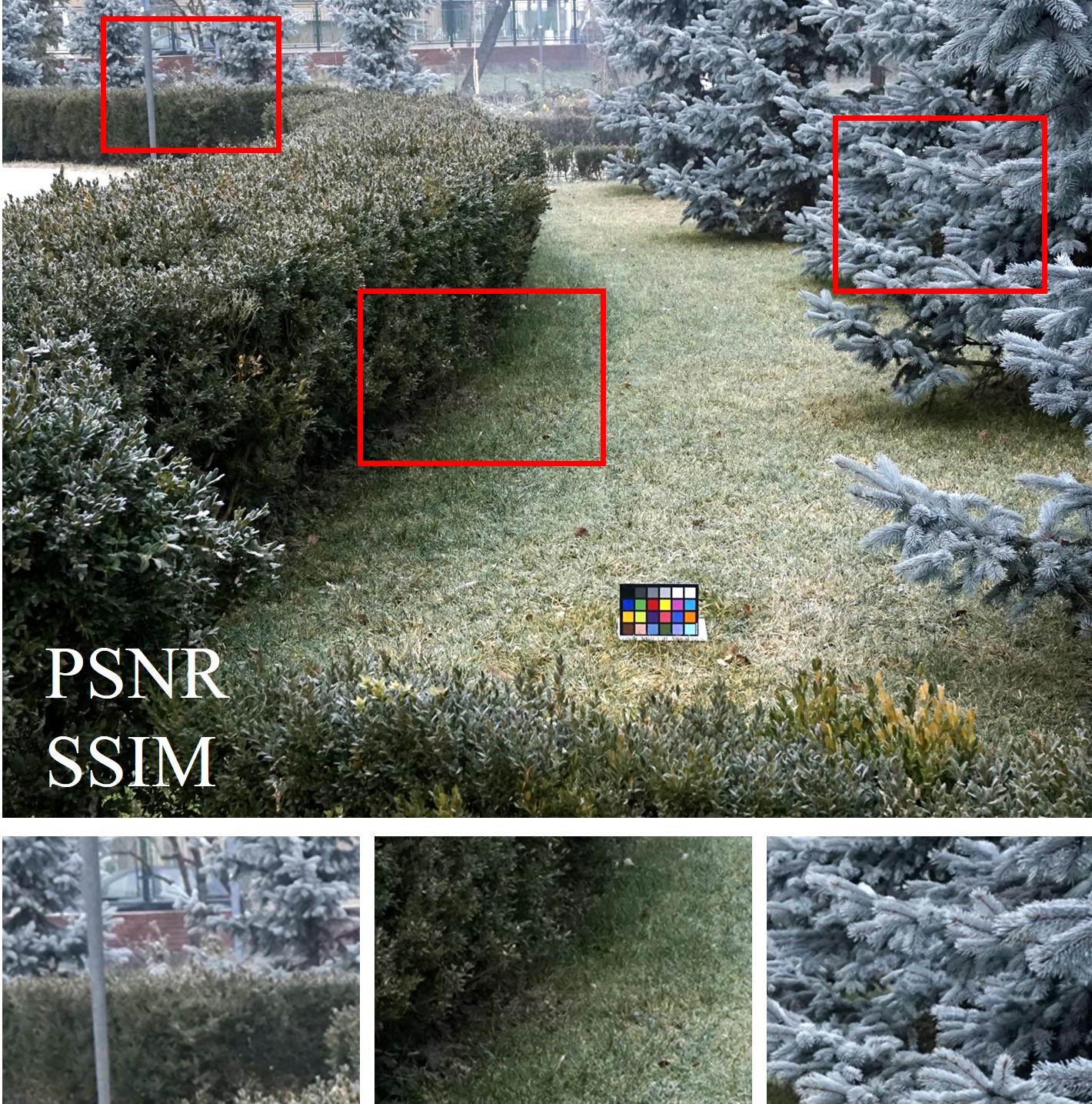} \\

      \includegraphics[width=0.190\linewidth]{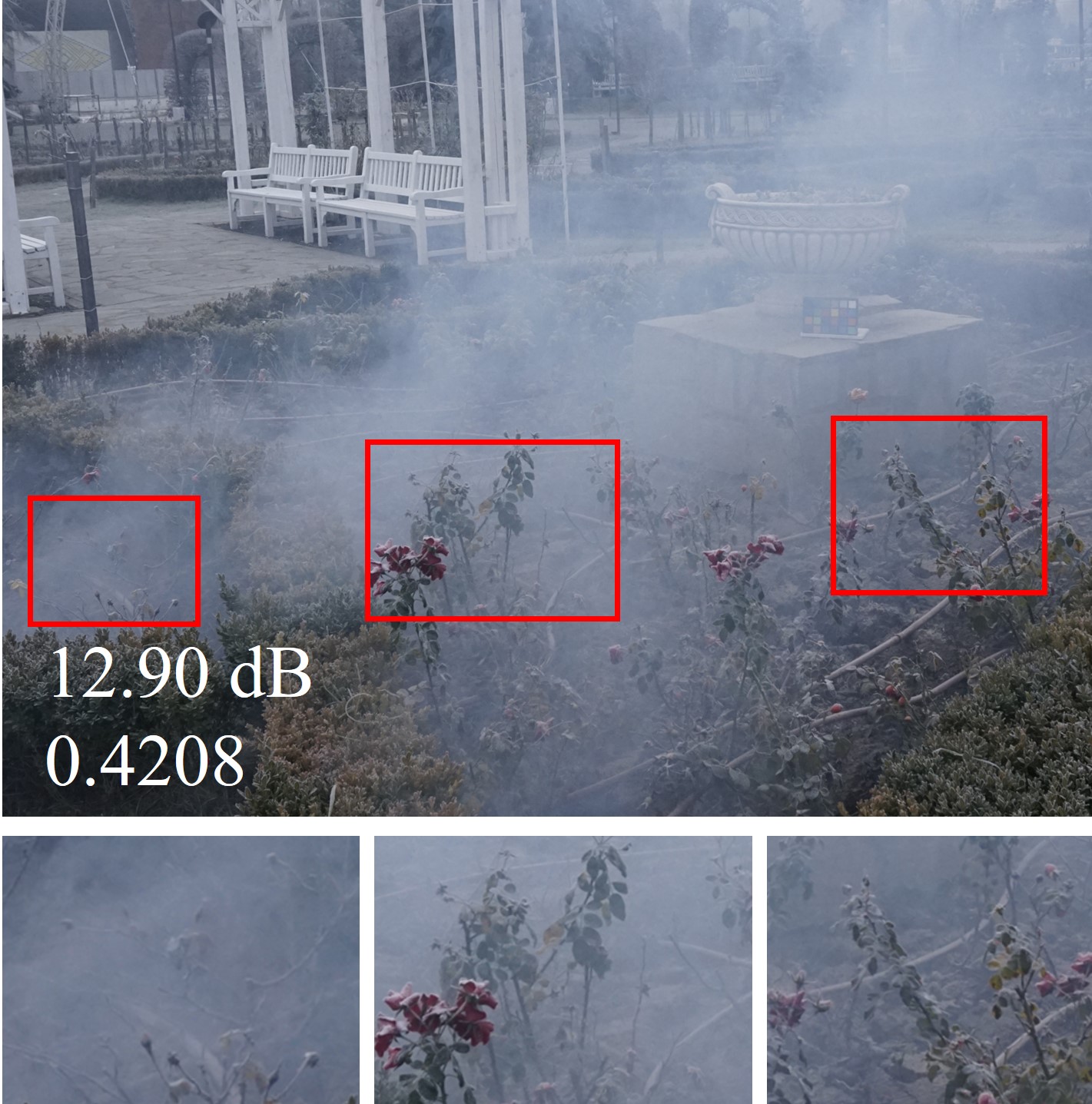} &
      \includegraphics[width=0.190\linewidth]{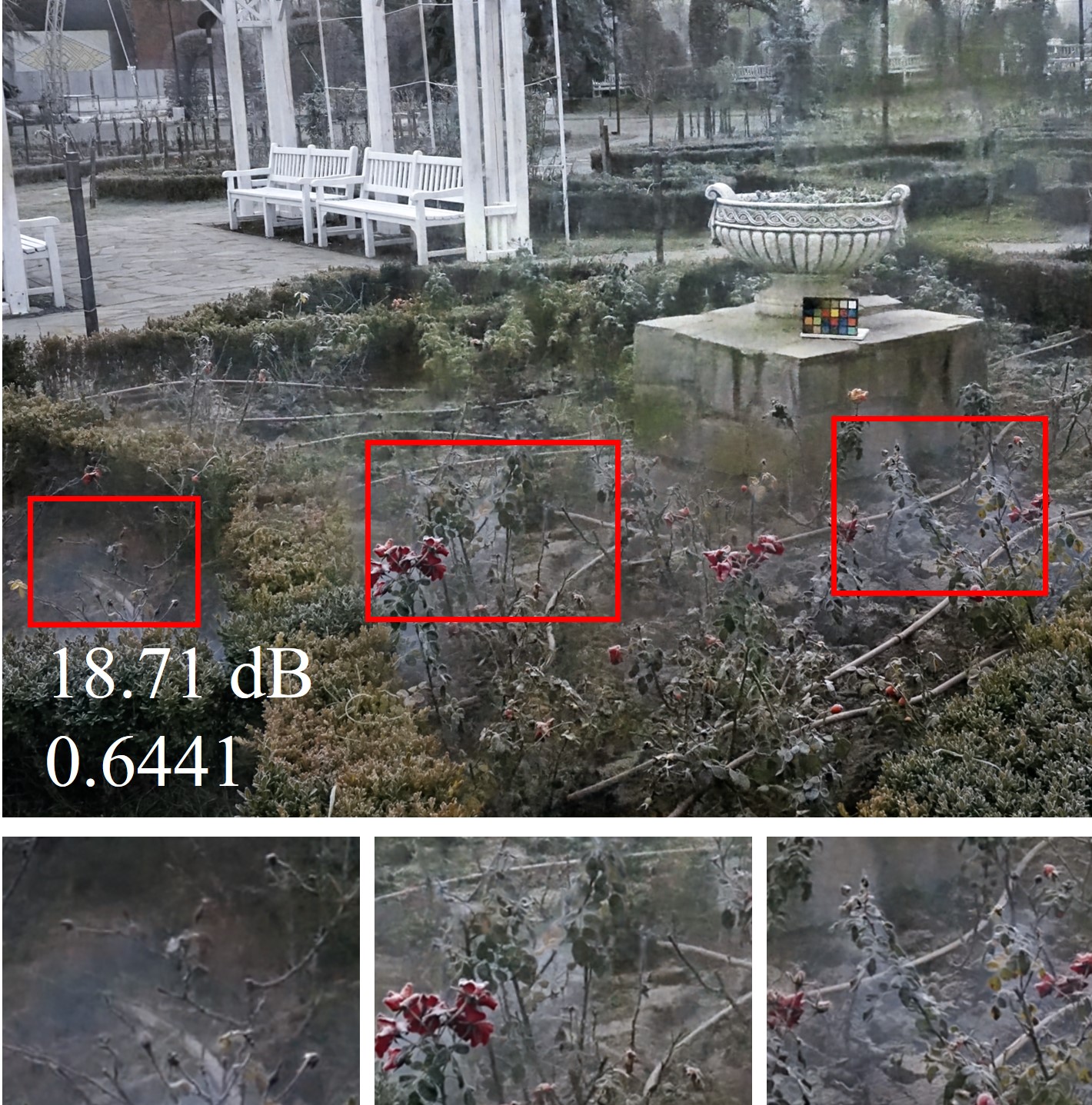} &
      \includegraphics[width=0.190\linewidth]{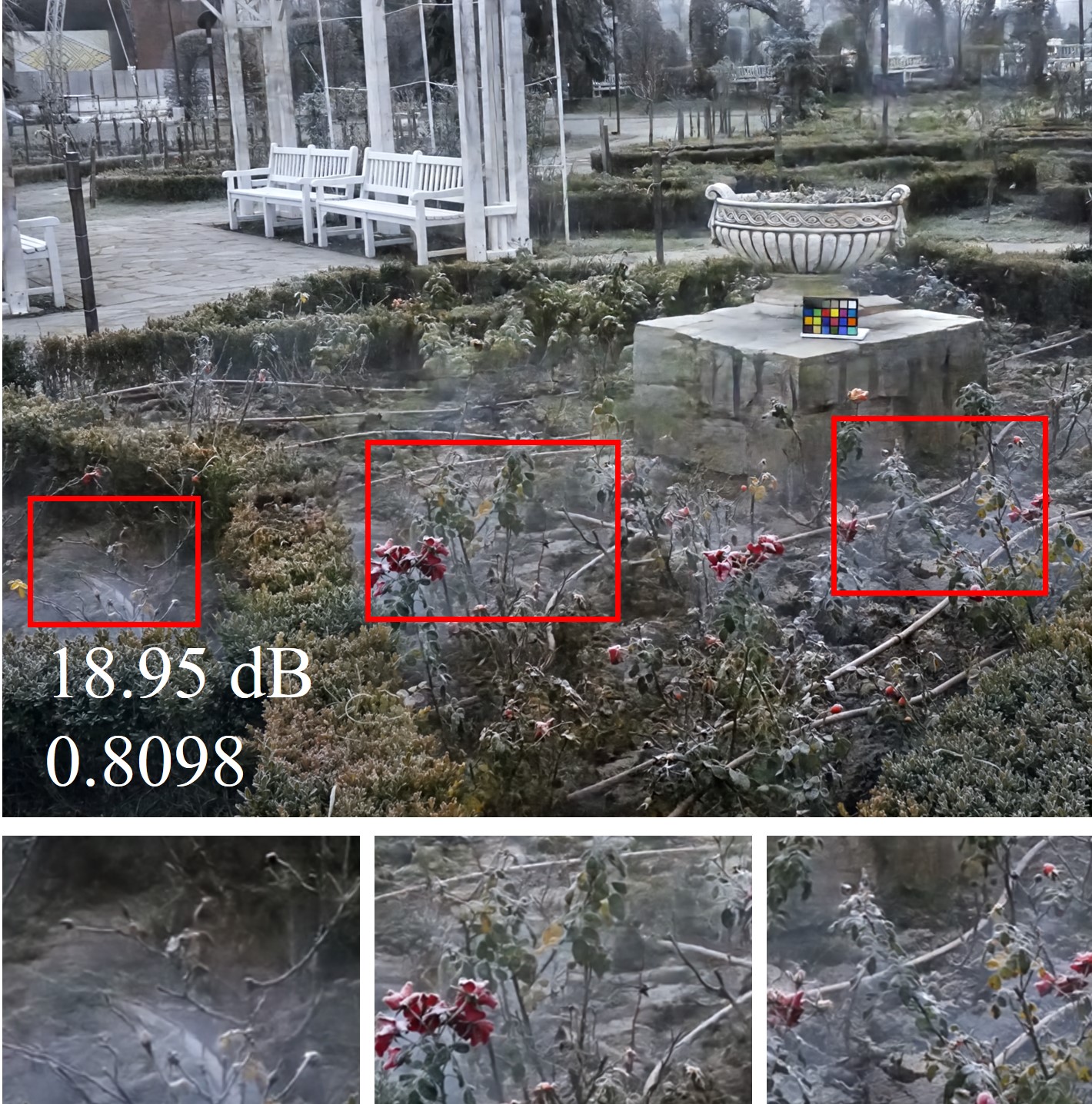} &
      \includegraphics[width=0.190\linewidth]{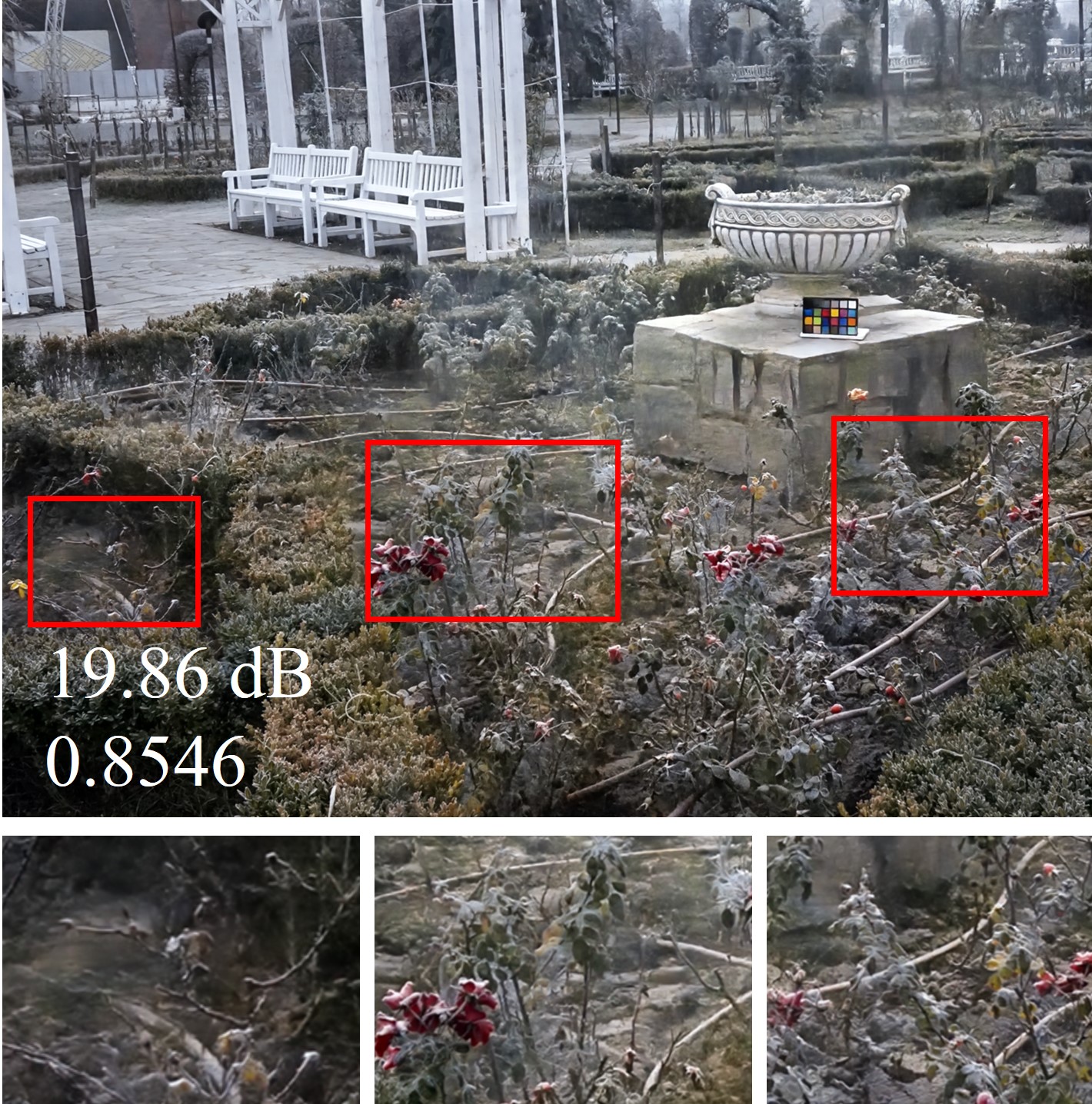} &
      \includegraphics[width=0.190\linewidth]{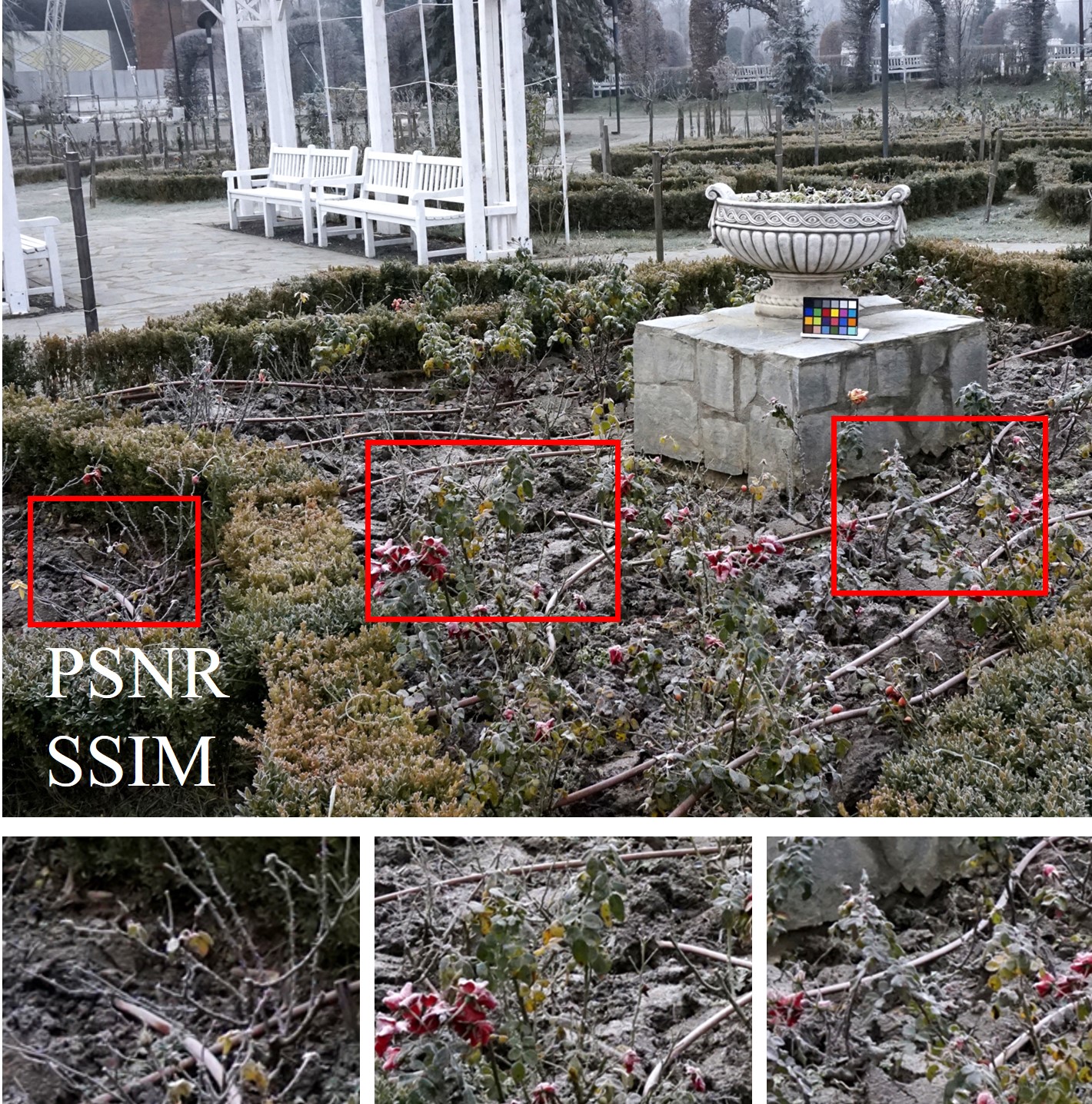} \\

	(a) Hazy image & 
        (b) Dehamer~\cite{9879191} & 
        (c) FocalNet~\cite{10377428} & 
        (d) Ours & 
        (e) GT 
		\end{tabular}
	\end{center}
	\caption{Visual comparisons on real-world hazy images from the NH-Haze dataset. Key regions marked with red boxes are enlarged in left-to-right order and arranged horizontally at the bottom.}
	\label{fig: nh}
\end{figure}

\subsubsection{Qualitative Comparisons}
\label{subsubsec: qual}
Figure~\ref{fig: Indoor} and Figure~\ref{fig: Outdoor} show the dehazing results on synthetic images from the SOTS-Indoor and SOTS-Outdoor datasets, respectively. Most comparative methods exhibit similar trends in both indoor and outdoor scenes. Earlier methods, such as FFA-Net~\cite{qin2020ffa}, MAXIM-2S~\cite{tu2022maxim}, and DeHamer~\cite{9879191}, introduce edge blurring and artifacts in regions with significant local variations (Figure~\ref{fig: Indoor} Line 1, (a)–(c)) and struggle to fully remove haze in areas with large depth of field (Figure~\ref{fig: Indoor}, Line 2, and Figure~\ref{fig: Outdoor}, Line 1, (a)–(c)). In contrast, FocalNet~\cite{10377428} and OKNet~\cite{cui2024omni} show improved performance, while our method provides the most thorough dehazing and delivers the highest fidelity, especially in detail-rich regions.

Figure~\ref{fig: dense} presents the dehazing results for two real-world scenes from the Dense-Haze~\cite{ancuti2019dense} dataset. In the first scene with extremely dense haze, the zoomed-in regions clearly show that Fourier-RWKV outperforms DeHamer~\cite{9879191} and MB-TaylorFormer-B~\cite{10378631} in detail restoration, while better preserving depth structure and spatial hierarchy. In the second scene, although our PSNR is slightly lower than that of DeHamer, our method excels in color fidelity and structural detail reconstruction, resulting in a higher SSIM score and better overall visual quality.

Figure~\ref{fig: nh} compares two non-uniform real-world haze scenes from the NH-HAZE~\cite{9150807} dataset. The zoomed-in regions from both scenes reveal that Fourier-RWKV achieves the most comprehensive dehazing, effectively handling regions with large depth and complex textures, while recovering texture details more effectively than other methods.

The quantitative and qualitative comparison results collectively validate the effectiveness of the proposed Multi-State Perception paradigm, emphasizing the strong adaptability of Fourier-RWKV to various haze degradation patterns. In particular, it demonstrates exceptional dehazing performance and robustness in real-world scenes, while maintaining high fidelity and low computational overhead.

\subsection{Ablation Studies}
\label{subsec: abla}
To evaluate the contributions and interactions of the core modules, we conduct systematic ablation studies. First, we assess the individual and combined effects of DQ-Shift, Fourier Mix, and SBM. Then, we perform component-level analysis by modifying each module independently while keeping the rest of the architecture fixed. All models are trained on the ITS dataset and evaluated on the SOTS-Indoor test set under consistent experimental conditions.

\begin{table}[tbp]
  \centering
  \caption{Ablation study for three core modules.}
  \renewcommand\arraystretch{1.2}
  \scalebox{0.8}{
  \begin{tabular}{l|cc|cc}
      \toprule
      \multicolumn{1}{c|}{\multirow{1}[4]{*}{Variant}}  & \multicolumn{2}{c|}{SOTS-Indoor}  &\multicolumn{1}{c}{Params} &\multicolumn{1}{c}{FLOPs} \\
      & \multicolumn{1}{c}{PSNR$\uparrow$} & \multicolumn{1}{c|}{SSIM$\uparrow$} &\multicolumn{1}{c}{(M)} &\multicolumn{1}{c}{(FLOPs)} \\
      \midrule
      Baseline &38.39 &0.994 &4.31 &13.25 \\ 
      $\rightarrow$ DQ-Shift &39.86 &0.995 &4.40 &14.03 \\ 
      $\rightarrow$ Fourier Mix &40.29 &0.996 &5.20 &14.81  \\
      $\rightarrow$ SBM &38.71 &0.995 &4.77 &13.36  \\  
      \makecell[l]{$\rightarrow$ DQ-Shift \&\\ \;\;\;\; Fourier Mix} &40.82 &0.996 &5.29 &15.58  \\
      \midrule
      Full Model &41.60 &0.996 &5.31 &15.69 \\
      \bottomrule
  \end{tabular}}
  \label{tab: table2}
\end{table}

\subsubsection{Effectiveness of Core Modules}
\label{subsubsec: core}
To isolate the individual effects and interactions among modules, we design the following configurations. Starting with a baseline that includes Q-Shift, Spatial Mix, and $1\times1$ convolutional skip connections, we construct six variants: (1) Baseline; (2) DQ-Shift replacing Q-Shift; (3) Fourier Mix replacing Spatial Mix; (4) SBM replacing $1\times1$ convolutions; (5) Combined DQ-Shift and Fourier Mix replacing their respective counterparts; (6) Full model incorporating all three proposed modules. The results are presented in Table~\ref{tab: table2}.

As shown in Table~\ref{tab: table2}, the baseline achieves a PSNR of 38.39 dB on the SOTS-Indoor dataset. Replacing the corresponding components with DQ-Shift and Fourier Mix increases the PSNR by 1.47 dB and 1.90 dB, respectively, validating the effectiveness of spatial deformable perception and frequency-domain global perception. Combining these two modules raises the PSNR to 40.82 dB, underscoring the complementarity of the two perceptual states, which synergistically enhance the model's degradation modeling capability.

Notably, introducing SBM alone into the baseline yields a modest 0.32 dB PSNR gain. However, when combined with DQ-Shift and Fourier Mix, SBM boosts PSNR by an additional 0.78 dB. The core function of SBM is to achieve semantic alignment between the encoder and decoder, its effectiveness depends on the representational power of the baseline model. In models with weaker representation, SBM has limited effect. Once the model learns robust and discriminative features, SBM can more effectively perform cross-stage semantic fusion, improving restoration consistency.

The full model, integrating all three modules, achieves the best performance. Compared to the baseline, the parameter count and FLOPs increase by only 1.0 M and 2.44 G, respectively. This demonstrates that our proposed Multi-State Perception paradigm significantly boosts dehazing performance with relatively low overhead through the collaborative modeling of spatial, frequency-domain, and semantic relationships. 

\begin{table}[tbp]
  \centering
  \caption{Ablation study for key component in DQ-Shift.}
  \renewcommand\arraystretch{1.2}
  \scalebox{0.8}{
  \begin{tabular}{l|cc|cc}
      \toprule
      \multicolumn{1}{c|}{\multirow{1}[4]{*}{Variant}}  & \multicolumn{2}{c|}{SOTS-Indoor}  &\multicolumn{1}{c}{Params} &\multicolumn{1}{c}{FLOPs} \\
      & \multicolumn{1}{c}{PSNR$\uparrow$} & \multicolumn{1}{c|}{SSIM$\uparrow$} &\multicolumn{1}{c}{(M)} &\multicolumn{1}{c}{(G)} \\
      \midrule
      Only fixed offset &40.77  &0.996  &5.22 &14.92 \\
      Only dynamic offset &41.28 &0.996 &5.31 &15.69 \\
      W/o gating &41.36 &0.996 &5.28 &15.46  \\
      \midrule
      Full DQ-Shift &41.60 &0.996 &5.31 &15.69 \\
      \toprule
  \end{tabular}}
  \label{tab: table3}
\end{table}

\subsubsection{Effectiveness of DQ-Shift}
To evaluate the impact of DQ-Shift, we design four variants: (1) The original fixed-pattern quad-directional shift; (2) A variant incorporating only the dynamic offset mechanism; (3) A combination of fixed and dynamic offsets, with the gating mechanism omitted; (4) Full DQ-Shift, which integrates both fixed and gated dynamic offsets. The results are reported in Table~\ref{tab: table3}.

As shown in Table~\ref{tab: table3}, compared to the full DQ-Shift, using only fixed or dynamic offset leads to a PSNR decrease of 0.83 dB and 0.32 dB, respectively. This highlights the importance of combining both offsets for optimal performance, with dynamic offset making a more significant contribution. When both offsets are used without the gating mechanism,  PSNR drops by 0.24 dB, indicating that the gating mechanism effectively adjusts the sampling positions of dynamic offset, improving local modeling accuracy. Additionally, while this variant has slightly fewer parameters and FLOPs than the one using only dynamic offset, it achieves a PSNR improvement of 0.08 dB, showing that the fixed offset provides valuable directional information for spatial perception.

\begin{table}[t]
  \centering
  \caption{Ablation study for key component in Fourier Mix.}
  \renewcommand\arraystretch{1.2}
  \scalebox{0.8}{
  \begin{tabular}{l|cc|cc}
      \toprule
      \multicolumn{1}{c|}{\multirow{1}[4]{*}{Variant}}  & \multicolumn{2}{c|}{SOTS-Indoor}  &\multicolumn{1}{c}{Params} &\multicolumn{1}{c}{FLOPs} \\
      & \multicolumn{1}{c}{PSNR$\uparrow$} & \multicolumn{1}{c|}{SSIM$\uparrow$} &\multicolumn{1}{c}{(M)} &\multicolumn{1}{c}{(G)} \\
      \midrule
      Classic scan &41.01 &0.996 &5.31 &15.69 \\
      Only spatial-domain gating &40.52 &0.996 &4.43 &14.14 \\
      Only Fourier-domain gating &41.14 &0.996 &5.09 &14.93  \\
      \midrule
      Full Fourier Mix &41.60 &0.996 &5.31 &15.69 \\
      \toprule
  \end{tabular}}
  \label{tab: table4}
\end{table}

\subsubsection{Effectiveness of Fourier Mix}
To analyze the role of each component in Fourier Mix, we conduct ablation studies with four configurations: (1) Classic row scanning applied to spectral features; (2) Spectral sequencing/inversion with gating applied only in the spatial domain; (3) Spectral sequencing/inversion with gating applied only in the Fourier domain; (4) Full Fourier Mix, the complete implementation with dual-domain gating. The results are presented in Table~\ref{tab: table4}.

As shown in Table~\ref{tab: table4}, compared to the full Fourier Mix, using classical row-wise scanning of spectral features results in a PSNR drop of 0.49 dB, indicating that our proposed distance-based sorting strategy better preserves the relative relationships between frequency points, enabling more accurate global dependency modeling. Retaining only the spatial-domain gating causes a PSNR decrease of 1.08 dB, highlighting that the absence of Fourier-domain gating severely limits the global representation ability. Additionally, retaining only the Fourier-domain gating leads to a PSNR 0.46 dB lower than the full module, emphasizing the crucial role of spatial-domain gating in enhancing local spatial perception. These results collectively confirm the effectiveness and necessity of the proposed dual-domain gating mechanism.

\begin{table}[t]
  \centering
  \caption{Ablation study for key component in SBM.}
  \renewcommand\arraystretch{1.2}
  \scalebox{0.8}{
  \begin{tabular}{l|cc|cc}
      \toprule
      \multicolumn{1}{c|}{\multirow{1}[4]{*}{Variant}}  & \multicolumn{2}{c|}{SOTS-Indoor}  &\multicolumn{1}{c}{Params} &\multicolumn{1}{c}{FLOPs} \\
      & \multicolumn{1}{c}{PSNR$\uparrow$} & \multicolumn{1}{c|}{SSIM$\uparrow$} &\multicolumn{1}{c}{(M)} &\multicolumn{1}{c}{(G)} \\
      \midrule
      Randomly initialized kernels &41.03 &0.996 &5.31 &15.68 \\
      Single scale dynamic kernels &41.46 &0.996 &5.31 &15.69 \\
      W/o KSFU &41.21 &0.996 &5.30 &15.59  \\
      W/o semantic replacement &41.33 &0.996 &5.31 &15.69 \\
      \midrule
      Full SBM &41.60 &0.996 &5.31 &15.69 \\
      \bottomrule
  \end{tabular}}
  \label{tab: table5}
\end{table}

\subsubsection{Effectiveness of SBM}
To validate the design rationale behind SBM, we develop five variants: (1) Replacing dynamic multi-scale semantic kernels with randomly initialized kernels of the same size; (2) Unifying multi-scale dynamic kernels into single-scale $5\times 5$ kernels; (3) Substituting KSFU with direct summation; (4) Adding semantic features to encoder features during feature correction, instead of replacing the DC component; (5) Full SBM, which integrates dynamic multi-scale semantic kernels, adaptive kernel selection fusion, and semantic replacement. The results are summarized in Table~\ref{tab: table5}.

As shown in Table~\ref{tab: table5}, compared to the full SBM, replacing dynamic multi-scale semantic kernels with randomly initialized kernels causes a 0.57 dB PSNR drop, demonstrating that kernels generated by the semantic similarity matrix more effectively capture semantic content. Using single-scale kernels results in a 0.14 dB PSNR decrease, emphasizing the advantage of multi-scale design in handling features of varying granularity. Replacing the KSFU module with direct summation leads to a 0.39 dB PSNR reduction, further confirming the benefits of adaptive kernel selection and fusion in cross-scale feature integration. Finally, adding semantic features instead of replacing the DC component causes a 0.27 dB PSNR decrease, suggesting that the semantic replacement strategy better introduces high-level semantic guidance while avoiding irrelevant low-frequency interference.

\section{Conclusion}
In this paper, we propose Fourier-RWKV, a multi-state perception dehazing model built on a linear-complexity RWKV framework, designed to efficiently handle non-uniform haze distributions. Fourier-RWKV integrates three complementary perceptual states: spatially deformable perception, frequency-domain global modeling, and semantic-guided feature fusion, overcoming the limitations of existing methods in complex haze environments. Specifically, DQ-Shift dynamically adapts the receptive field to local haze variations, Fourier Mix extends the WKV attention mechanism from the spatial domain to the Fourier domain, thereby enhancing the capture of global dependencies, and SBM ensures semantic consistency by aligning encoder-decoder features. Extensive experiments demonstrate that Fourier-RWKV outperforms existing methods across multiple benchmark datasets, particularly in challenging real-world scenarios, highlighting its robustness and adaptability. This work not only advances dehazing technology but also extends the Vision-RWKV framework, providing a strong foundation for modeling both global and local information in a wide range of vision tasks.

\section*{Acknowledgment}
This work was partially supported by National Natural Science Foundation of China (No.62471317), Natural Science Foundation of Shenzhen (No. JCYJ20240813141331042), and Guangdong Provincial Key Laboratory (Grant 2023B1212060076).

\bibliographystyle{elsarticle-num} 
\bibliography{ref}

\end{document}